\documentclass{article}
\usepackage{graphicx}
\usepackage[table,xcdraw]{xcolor}
\PassOptionsToPackage{compress,numbers}{natbib}

\usepackage{microtype}
\usepackage{graphicx}
\usepackage{subcaption}
\usepackage{titletoc}
\usepackage{booktabs} 

\usepackage[hidelinks]{hyperref}

\usepackage[main, final]{neurips_2025}

\usepackage{amsmath}
\usepackage{amssymb}
\usepackage{mathtools}
\usepackage{twemojis}
\usepackage{amsthm}
\usepackage{enumitem}
\usepackage{graphicx}
\usepackage{subcaption}

\usepackage[capitalize,noabbrev]{cleveref}

\theoremstyle{plain}

\theoremstyle{definition}

\theoremstyle{remark}

\usepackage[textsize=tiny]{todonotes}
\usepackage{multirow}
\usepackage{wrapfig}

\begin{document}

\title{ABCD: All Biases Come Disguised}

%

\author{%
Mateusz Nowak$^{*}$ \quad Xavier Cadet\thanks{Equal contribution}
\quad Peter Chin \\
Dartmouth College\\
}
\maketitle

\begin{abstract}
Multiple-choice question (MCQ) benchmarks have been a standard evaluation practice for measuring LLMs' ability to reason and answer knowledge-based questions. Through a synthetic NonsenseQA benchmark, we observe that different LLMs exhibit varying degrees of label-position-few-shot-prompt bias, where the model either uses the answer position, the label in front of the answer, the distributions of correct answers present in the few-shot prompt, or a combination of all to answer each MCQ question. We propose a simple bias-reduced evaluation protocol that replaces the labels of each question with uniform, unordered labels and prompts the LLM to use the whole answer presented. With a simple sentence similarity model, we demonstrate improved robustness and lower standard deviation between different permutations of answers with a minimal drop in LLM's performance, exposing the LLM's capabilities under reduced evaluation artifacts, without any help from the prompt examples or the option labels. Across multiple benchmarks and models, this protocol substantially improves the robustness to answer permutations, reducing mean accuracy variance $3\times$ with only a minimal decrease in the mean model's performance. Through ablation studies on various embedding models and similarity functions, we show that the method is more robust than the standard ones.
\end{abstract}
\section{Introduction}
Multiple-choice questions (MCQs) are a standard paradigm for evaluating the reasoning and question-answering capabilities of large language models (LLMs), spanning benchmarks from commonsense reasoning (CommonsenseQA; CSQA) \cite{talmor-etal-2019-commonsenseqa} to graduate-level, Google-proof scientific questions (GPQA) \cite{rein2024gpqa}. However, more broadly, many real-world decision-making scenarios can be cast as MCQs by just enumerating candidate options. However, LLMs are known to be sensitive to superficial MCQ artifacts.

\Citet{zheng2024large} shows that an “answer-moving attack”, which simply moves the golden answer to a specific position, causes LLMs’ dramatic performance fluctuations. Building upon this work, \Citet{zhou2024revisiting} evaluates the influence of position shuffling, label replacement with emoji symbols, and format question change to True/False, showing that the decline in accuracy is more pronounced for option label replacement than for position shuffle. While \cite{zheng2024large} demonstrated position and label bias through "answer-moving attacks", they did not investigate how models exploit patterns in few-shot examples. Building upon both works, we reveal a critical dimension not explored in prior research: few-shot prompt distribution bias. 


We propose \textit{NonsenseQA}, a simple synthetic dataset of random-word questions and options with a randomly assigned correct answer, to quantify evaluation biases. Using NonsenseQA, we demonstrate that different models exhibit fundamentally different behaviors: some models explicitly exploit the few-shot answer distribution in their reasoning to reach $>95\%$ accuracy on meaningless inputs, while others achieve only $\sim 40\%$ without explicitly referencing the prompt examples, showing that MCQ bias is more complex and model-specific than previously.

Building on insights from NonsenseQA, we propose a bias-reduced evaluation protocol for multiple-choice LLM assessment. The protocol uses homogenous, unordered option labels and requires full-text answer generation, minimizing exploitable artifacts while improving robustness. Under this setting, models achieve near-chance accuracy on NonsenseQA while incurring only minimal performance degradation on standard MCQ benchmarks. To accommodate paraphrased answers, we match generated responses to candidate options using semantic similarity. 

We evaluate the protocol across 13 LLMs and five benchmarks -– MMLU-Pro \cite{wang2024mmlu}, GPQA \cite{rein2024gpqa}, CommonsenseQA \cite{talmor-etal-2019-commonsenseqa}, ARC \cite{clark2018arc}, and a subset of INCLUDE \cite{romanouinclude}. The results show an improved robustness, reflected in the combination of a higher SCORE \cite{nalbandyan-etal-2025-score} with reduced variance ratio. Ablation studies confirm that performance is stable across similarity models and matching functions.

The \textbf{contributions} of this paper are:
\begin{itemize}
\item \textbf{Simple practical evaluation debiasing solution}: We propose a simple, single-pass, bias-reduced MCQ evaluation standard that combines uniform option labels, full-text answer generation, and semantic matching, requiring neither model fine-tuning nor access to raw output probabilities, while substantially reducing shortcut exploitation and only requiring an additional 3\% computation time. We evaluate our design on multiple benchmarks, motivating our design choices with thorough ablation studies.
\item \textbf{NonsenseQA}: We propose a debiasing tool, the NonsenseQA dataset, that uses random words to show different biases present within LLMs when it comes to question answering, ranging from labels used to present the answer, the mode of the answer, the position of the answer, and the distribution of the correct answers present in the few-shot prompt.
\item \textbf{Few-shot prompt and answer format bias analysis}: We identify and characterize previously underexplored biases in MCQ-based LLM evaluation, including few-shot answer distribution bias and answer-mode bias, and demonstrate that these artifacts can be systematically exploited even on semantically meaningless inputs using our new diagnostic dataset.
\end{itemize}
\section{Related work}
Prior LLM evaluation on multiple-choice questions (MCQs) has largely relied on cloze-style formulations, where models select the option with the highest probability. However, predicting option symbols (e.g, “A”) through multiple-choice symbol binding (MCSB) was shown to improve the LLM’s performance \cite{robinsonleveraging}, and first-token prediction was demonstrated to be brittle to prompt phrasing and misaligned with full-text answer preferences in instruction-tuned models \cite{wang2024look}.

Nevertheless, standard MCSB-based MCQ protocols suffer from a range of its own biases, including sensitivity to option labels \cite{zheng2024large, zhou2024revisiting}, answer position \cite{jeong2025scope, egressy2025setllm, pezeshkpour2024large, mcilroy-young2024orderindependence, sandan-etal-2025-knockout, brown2025order}, question phrasing \cite{lunardi2025robustness}, and relative prompt and chain-of-thought structure \cite{raman2026reasoning}. Most existing approaches focus on position and label biases, typically require model fine-tuning, access to logits or attention mechanisms, or multiple evaluation passes \cite{egressy2025setllm, brown2025order, zheng2024large, mcilroy-young2024orderindependence, sandan-etal-2025-knockout, pezeshkpour2024large}. In contrast, our approach addresses multiple MCQ-induced biases in a single forward pass without fine-tuning or access to internal model states.

Recent work has proposed abandoning the MCQ question format in favor of free-form answer generation \cite{whichofthese}, including two-pass schemes that map generated answers back to options \cite{chandak2025answer} or regenerate the MCQ label after the first reasoning pass \cite{jo2025finding}. However, such approaches are not universally applicable, as some questions lack unambiguous free-form answers and matching generated responses to candidate options can be NP-hard \cite{chandak2025answer}.

Nonetheless, MCQ evaluation offers its own distinct advantages, such as supporting constrained decision spaces \cite{rahmani2025selfcorrectinglargelanguagemodels} and “None of the above” distractors \cite{elhady-etal-2025-wicked, tam-etal-2025-none}, motivating its continued relevance \cite{zhang2025rethinking}. Building on this insight, we retain the MCQ formulation while requiring models to generate full answer text rather than selecting a label, which is then matched to candidate options. This hybrid approach reduces previously identified biases while preserving the practical benefits of MCQ benchmarks.

Our work is inspired by previous research on free-form answer generation, which examines answer equivalence through semantic similarity to enhance evaluation metrics. \cite{bulian2022tomayto, risch2021semantic}. We build on their insights and adapt these ideas to create an improved MCQ evaluation standard. Similarly, while some of the biases used in this study have been explored in previous literature, with position and label biases extensively explored in \cite{zheng2024large,zhou2024revisiting}, and some of the few-shot example majority and recency biases explored in \cite{zhao2021calibrate}, we differ from their models in our evaluation, as we do not require raw logits and explore a greater combination of biases \cite{zheng2024large, zhao2021calibrate}, with \cite{zheng2024large,zhou2024revisiting} overlooking the few-shot bias. Additionally, while \cite{jo2025finding} explores null prompting, similar to our NonsenseQA, their approach relies on deliberately placing the correct answer in the least likely position, far from semantically similar distractors. In contrast, our method requires no prior knowledge of the correct answer and is applicable in standard evaluation settings. Moreover, we use NonsenseQA as a diagnostic tool to expose a broader range of biases in MCQ evaluation.

\section{Problem Formulation}

We formalize the MCQ evaluation setting to characterize the sources of bias that models may exploit.
An MCQ dataset $\mathcal{D}$ consists of $N$ instances, where each instance $(q_k, \mathcal{A}_k, a_k^*)$ for $k \in \{1, \ldots, N\}$ comprises a question $q_k$, a set of $n$ candidate answer options $\mathcal{A}_k = \{a_k^{(1)}, \ldots, a_k^{(n)}\}$, and a ground-truth answer $a_k^* \in \mathcal{A}_k$.

An evaluation protocol $\mathcal{E} = (\mathcal{P}, \mathcal{L}, \pi, \mathcal{X})$ transforms each MCQ instance into a prompt and extracts a prediction from the model. The protocol is defined by four components, each introducing potential bias:
\begin{enumerate}[label=(\roman*), nosep, leftmargin=*]
    \item \textit{Few-shot prompt}: $\mathcal{P} = \{(q_j, \mathcal{A}_j, a_j^*)\}_{j=1}^{m}$: a set of $m$ exemplar questions with answers, whose answer distribution may be exploited;
    \item \textit{Option labels}: $\mathcal{L} = (\ell_1, \ldots, \ell_n)$: an ordered tuple of symbols assigned to each answer position, which may carry implicit ordering cues;
    \item \textit{Permutation}: $\pi \in S_n$: an element of the symmetric group determining the presentation order of answer options;
    \item \textit{Extraction function}: $\mathcal{X}: \mathcal{Y} \rightarrow \mathcal{A}_k$: a mapping from the space of model outputs $\mathcal{Y}$ to a candidate answer.
\end{enumerate}
Given a language model $f_\theta: \mathcal{T} \rightarrow \mathcal{Y}$ mapping input text $\mathcal{T}$ to output text $\mathcal{Y}$, the predicted answer for question $k$ under protocol $\mathcal{E}$ is:
\begin{equation}
    \hat{a}_k = \mathcal{X}\bigl(f_\theta(\textsc{Prompt}(q_k, \mathcal{A}_k, \mathcal{L}, \pi, \mathcal{P}))\bigr),
\end{equation}
where $\textsc{Prompt}(\cdot)$ constructs the full input according to the protocol specification.

Using these definition, we can formulate the evaluation protocols:
The \textbf{standard evaluation -- Select-and-Letter (S\&L) --} employs distinct labels, for instance with $|\mathcal{A}_k|=4$,  $\mathcal{L} = (\text{A}, \text{B}, \text{C}, \text{D})$ and an extraction function $\mathcal{X}$ that identifies a single letter from the model output.
Our \textbf{proposed protocol -- Matched-and-Dashed (M\&D) --} instead uses uniform labels $\mathcal{L} = (\text{-}, \text{-}, \text{-}, \text{-})$ and defines $\mathcal{X}$ as a semantic similarity matching operation between the model's output and candidate answers $\mathcal{A}_k$. 

\section{Matching answers - bias-reduced evaluation protocol}
\begin{figure*}[ht]
  \begin{center}
    \centerline{\includegraphics[width=\textwidth]{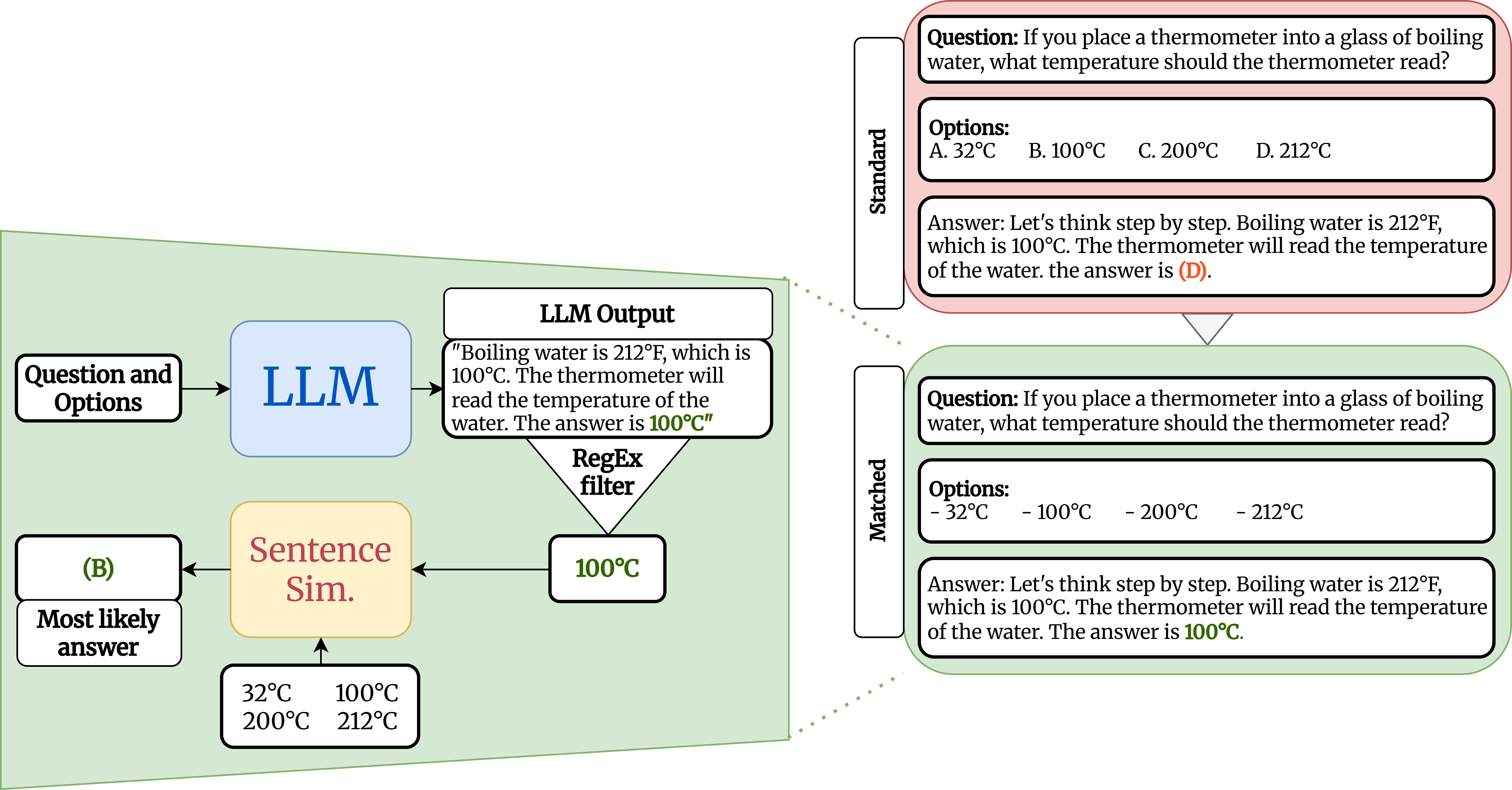}}
    \caption{An example of the wrong conclusion reached by the Gemma-3-12 B-it model on one of the questions from the ARC \cite{clark2018arc} dataset with our proposed method. By changing the presentations from standard evaluation (with different labels and predicting only the answer label) to matched (with uniform labels and predicting a whole answer, shown on the left), we eliminate the label bias present, and the model arrives at a correct conclusion that aligns with its reasoning.}
    \label{fig:example_question}
  \end{center}
\end{figure*}

Inspired by previous work \cite{zheng2024large, risch2021semantic}, we propose a bias-reduced evaluation protocol for LLMs in the MCQ setting. 

Previous work investigated how various answer labels affect models' accuracy and robustness \cite{zheng2024large,zhou2024revisiting}. All previous work focused on distinct-per-answer labels, without considering uniform labels. Nevertheless, the issue primarily stems from distinct per-answer labels, rather than the labels themselves.  Therefore, we propose using standard dashes "-" as labels to mimic an unordered list in a standard Markdown setting and remove any label bias introduced by the natural ordering of labels.

However, without distinct labels, we cannot predict labels as our answers. If we default to predicting ordered labels (e.g., A/B/C/D), we introduce prediction bias. Therefore, to remove any label-dependent bias, we make the model choose the full-text option, as in \cite{zheng2024large}. We use a simple instruction and a set of regular expressions to extract the final, one-sentence answer from the LLM's output. While \citet{zheng2024large} uses a cloze evaluation to extract the final answer, we use standard generation coupled with a sentence similarity model to produce embeddings for the extracted final answer and the possible MCQ options. This allows for variation in the model's output, the use of synonyms, and chain-of-thought reasoning. As the final answer, we choose the option with the highest similarity to the extracted final answer from the LLM's output. 

In standard MCQ benchmark evaluation protocols, a random answer option is selected as a fallback. Similarly, we also allow the final regular expression-based answer extraction to be more random by extracting the last sentence of the LLM's output if the previous regular expressions fail. This serves as a better extraction method than standard ones, as some of the reasoning might align with the output. Nevertheless, we find that this extraction method is rarely used. Moreover, to enforce a full-sentence answer, we slightly modify the original prompt. We depict all the regular expressions used for answer extraction and the modified prompt in Appendix~\ref{appendix:prompt}.

As for the sentence similarity model and the similarity function, we use the Qwen3-Embedding-0.6B \cite{zhang2025qwen3} model with cosine similarity.
Nevertheless, we find that the model and the similarity function choice do not influence the model's performance significantly, as demonstrated in the Figure~\ref{fig:ablation_models} in the Appendix~\ref{appendix:ablation}. A sentence similarity model addresses the shortcomings of a standard cloze evaluation in MCQ question answering, allowing chain-of-thought reasoning and similar answers to be accepted. (We provide a further discussion in Appendix~\ref{appendix:sentence_sim_over_cloze})

\section{Results}
\subsection{Models, datasets, and parameters}
We evaluate 13 open-source LLMs ranging from 8B to 32B parameters, including model families like DeepSeek-R1 \cite{guo2025deepseek}, Qwen3 \cite{yang2025qwen3}, Llama-3.1 \cite{grattafiori2024llama}, Gemma-3 \cite{team2025gemma}, Ministral-3 \cite{liu2026ministral}, Nemotron \cite{wang2025nemotron, blakeman2025nemotron}, Phi-4 \cite{abdin2024phi}, and GPT-OSS \cite{agarwal2025gpt}. We use our synthetic NonsenseQA benchmark and five real-world benchmarks: CSQA \cite{talmor-etal-2019-commonsenseqa}, ARC \cite{clark2018arc}, MMLU-Pro \cite{wang2024mmlu}, GPQA \cite{rein2024gpqa}, and a multilingual subset of INCLUDE \cite{romanouinclude}. For INCLUDE, we report results on four languages shared across all model training data: Spanish, French, Italian, and German. All detailed statistics are provided in the Appendix~\ref{appendix:detailed_results} in Tables \cref{tab:CSQA,tab:ARC,tab:INCLUDE,tab:MMLU-Pro,tab:GPQA}.

For each model, we evaluate the original permutation and $|\mathcal{A}_k|$ answer-moving attacks using a fixed seed and the model’s preferred generation parameters, as previous research has shown the temperature $t \in [0,1]$ does not negatively impact model performance \cite{renze2024effect}; when unavailable, we use a standard configuration described in Appendix~\ref{appendix:parameters}. During each answer-moving attack, we vary the permutation $\pi$ to alter the position of the ground-truth answer $a_k^*$ within both the test question and the few-shot prompt $\mathcal{P}$, except for the GPQA dataset, where we modify $\pi$ only for test questions. Across all permutations, the identity of $a_k^*$ remains unchanged; only its position varies.

In our analysis, we evaluate performance metrics that include the original, dataset-provided option permutation accuracy (represented by dots on the plot) and accuracy under "moving answer" attacks \cite{zheng2024large}. For each model, we present a box plot that summarizes the attack performance metrics, displaying the standard deviation and median with the box, and the minimum and maximum accuracy with the whiskers. This representation allows us to assess the robustness of the model's performance. By comparing both the median accuracy of the attack permutations and the original permutation performance, we can quantify the difference between the standard and anomalous prompts.
Additionally, to provide a comprehensive evaluation, we compute a SCORE \cite{nalbandyan-etal-2025-score} robustness metric across all permutations, including both original and attack scenarios. Let $\Pi$ denote the set of permutations under which we evaluate, and let $\hat{\mathcal{Y}}_k = \{\hat{a}_k^{(\pi)} : \pi \in \Pi\}$ be the set of predictions for question $k$ across all permutations. The SCORE is defined as:
\begin{equation}
    \text{SCORE} = \frac{1}{N}\sum_{k=1}^{N}\sum_{\hat{a}_i \in \hat{\mathcal{Y}}_k}\sum_{\substack{\hat{a}_j \in \hat{\mathcal{Y}}_k\\ i\neq j}}\frac{sim(\hat{a}_i, \hat{a}_j)}{\binom{|\hat{\mathcal{Y}}_k|}{2}}
\end{equation}
where $N$ is the number of questions in the dataset $\mathcal{D}$; $\hat{a}_i$ and $\hat{a}_j$ are predictions for question $k$ under different permutations; and $sim(\hat{a}_i, \hat{a}_j)$ is a similarity function.
Moreover, to measure the effects of our method on the stability of the model's performance, we propose to estimate a variance ratio $\sigma^2_{R}$ defined as:
\begin{equation}
\label{eq:var_ratio}
    \sigma^2_{R} = \frac{\sigma^2_{M\&D}}{\sigma^2_{S\&L}},
\end{equation}
where $\sigma^2_{S\&L}$ and $\sigma^2_{M\&D}$ is variance under S\&L and M\&D evaluation protocol, respectively. 
\subsection{NonsenseQA} \label{section:nonsenseqa}

\begin{wrapfigure}{r}{0.5\textwidth}
  \begin{center}  \centerline{\includegraphics[width=0.5\textwidth]{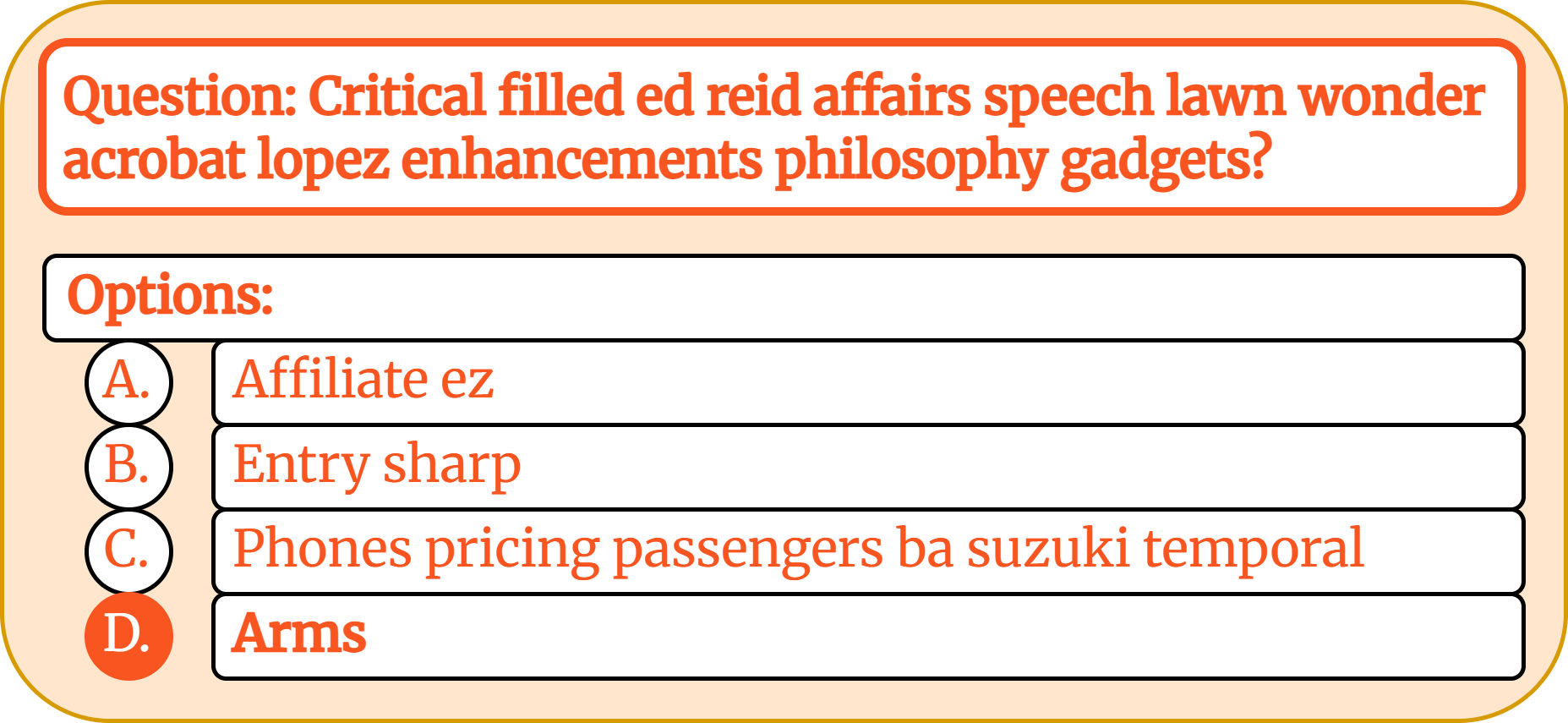}}
    \caption{An example question from NonsenseQA with random answers and a golden answer chosen at random as "D. Arms".}
    \label{fig:nonsenseqa_example_question}
  \end{center}
\end{wrapfigure}

\begin{figure*}[ht]
  \begin{center}
    \centerline{\includegraphics[width=\textwidth]{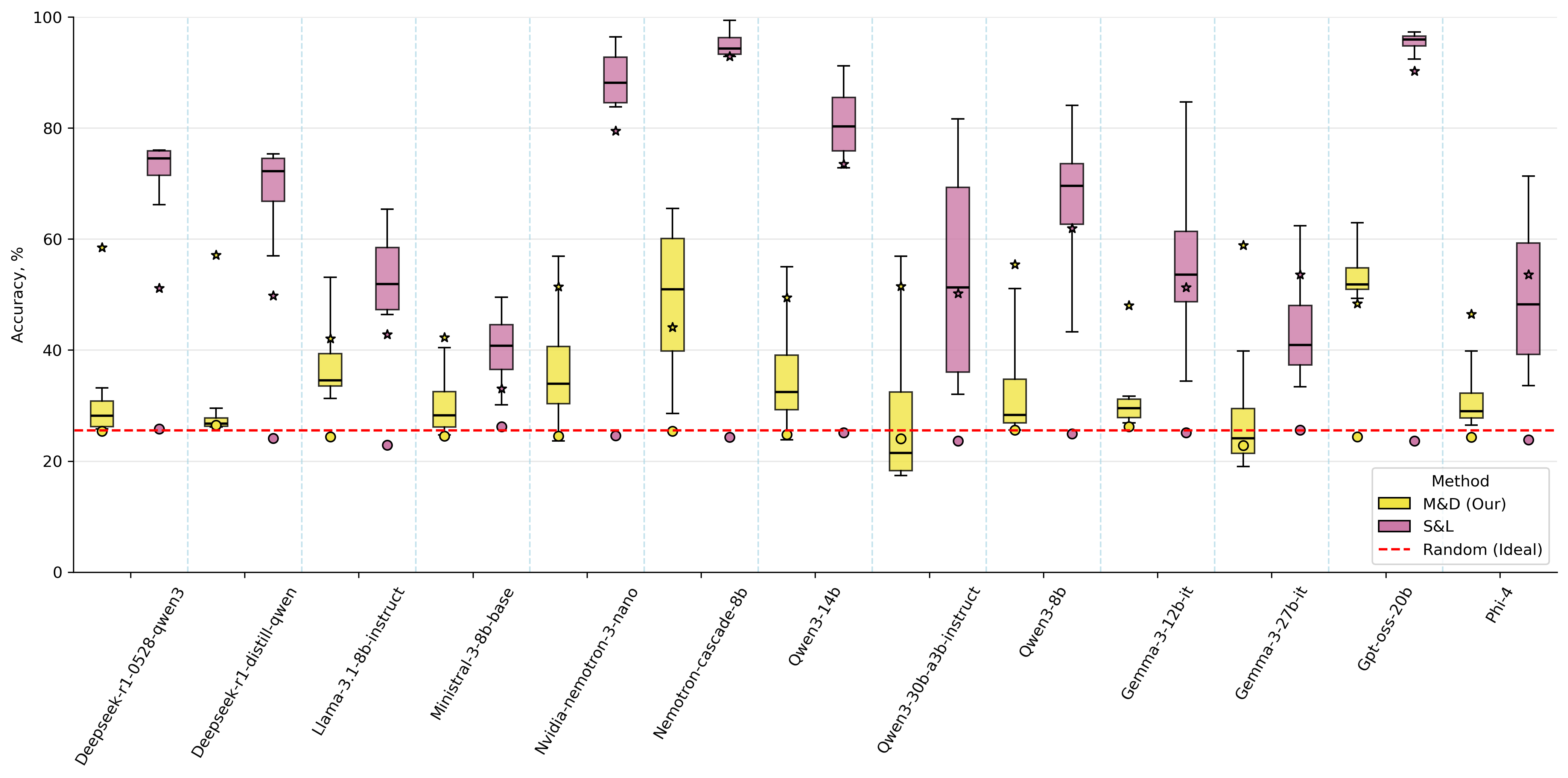}}
    \caption{Comparison of the matched prediction with dashes as labels (M\&D; our method) with standard letter prediction with letters (A/B/C/D) as labels (S\&L) on NonsenseQA with a 5-shot prompt. The boxes illustrate the model performance under all possibilities of "answer-moving attacks", where the whiskers indicate the minimum and maximum accuracy for each model. Each dot represents the performance of the original permutations. Additionally, each star symbolizes a SCORE \cite{nalbandyan-etal-2025-score} robustness metric.}
    \label{fig:NonsenseQA-comparison}
  \end{center}
\end{figure*}

Before evaluating our bias-robust evaluation protocol on real benchmarks, we first introduce \textit{NonsenseQA} a synthetic dataset that allows us to quantify the influence of various biases on a model's output, motivating the design of our approach. NonsenseQA consists of 1,000 random questions, each with four answer options $\forall k \, , |\mathcal{A}_k| = 4$. Each question is formed using a selection of five to twenty random words, while each answer consists of one to six random words.
To ensure the use of actual words, we utilize the "Wordlist 10,000" for each component.
Additionally, we assign a pseudo-golden answer to each question in a way that maintains a uniform distribution of answers, ensuring that no bias is introduced in the original data, we provide an example in Figure~\ref{fig:nonsenseqa_example_question}.
We create a smaller validation dataset of $100$ questions, where each question can be used for the few-shot prompting when querying the NonsenseQA.

We designed the benchmark to demonstrate various types of biases, as by design the accuracy on NonsenseQA should be close to the chance level of 25\%.
However, under the standard evaluation protocol, we observe a different behavior.
The results are presented in Figure~\ref{fig:NonsenseQA-comparison}, with our proposed method of matching answers with uniform labels, labeled as "M\&D," in comparison to the standard letter prediction and letter label methods, which are marked as "S\&L".
In the Appendix~\ref{appendix:nonsenseQA}, we examine each component's influence on the accuracy of NonsenseQA, with additional statistics from only matching but preserving the letters (M\&L) and the standard letter prediction but using dashes as labels (S\&D).

We can observe \textit{three types of models} when it comes to the few-shot and label bias under the S\&L method (\cref{appendix:models_types}). 
\begin{enumerate}
    \item \textbf{Explicit bias models}:
    use answers to the few-shot prompt questions to reason about the provided question. Within the model's output, we see that the model directly references answers to other questions to answer the test question. An example model is the GPT-OSS \cite{agarwal2025gpt}.
    
    \item \textbf{Implicit bias models}:
    use answers to the few-shot prompt questions to reason about the provided answer. An example model could be the Qwen3-8B \cite{yang2025qwen3}. These models do not directly reason about other answers, but achieve a performance greater than 50\% consistently, showing that there is bias when it comes to the distribution present in the few-shot prompt.
    
    \item \textbf{Models unable to exploit the bias}:
    only marginally exploit the few-shot bias implicitly. These models, while still experiencing higher than usual performance with S\&L, cannot utilize the bias consistently, achieving high performance only a few times with median scores at around 50\%. An example model would be Gemma-3-27b-it \cite{team2025gemma}. However, none of the models achieve a mean lower than 40\%.
\end{enumerate}

Under the proposed M\&D protocol, median accuracy on NonsenseQA drops sharply, with several models approaching chance-level performance of 25\%, indicating effective suppression of bias exploitation. By using uniform labels and full-text answer generation, the protocol blurs few-shot patterns and limits models reliance on them. While a subset of models still exhibits some biases, with median accuracy $\sim 50\%$, this remains substantially lower than the $>95\%$ medians observed under standard evaluation.

Across all models, median accuracy under S\&L evaluation is significantly higher than under M\&D. A paired Wilcoxon test confirms this difference (p-value $< 10^{-4}$; Cohen’s $d=2.2$), with per-model median reductions ranging from 12.5\% (Ministral-3) to 52\% (Nemotron-3-Nano), demonstrating that M\&D reduces bias-driven performance gains.

\subsection{Reasoning benchmarks: CSQA and ARC.}

\begin{figure*}[ht]
  \begin{center}
    \centerline{\includegraphics[width=\textwidth]{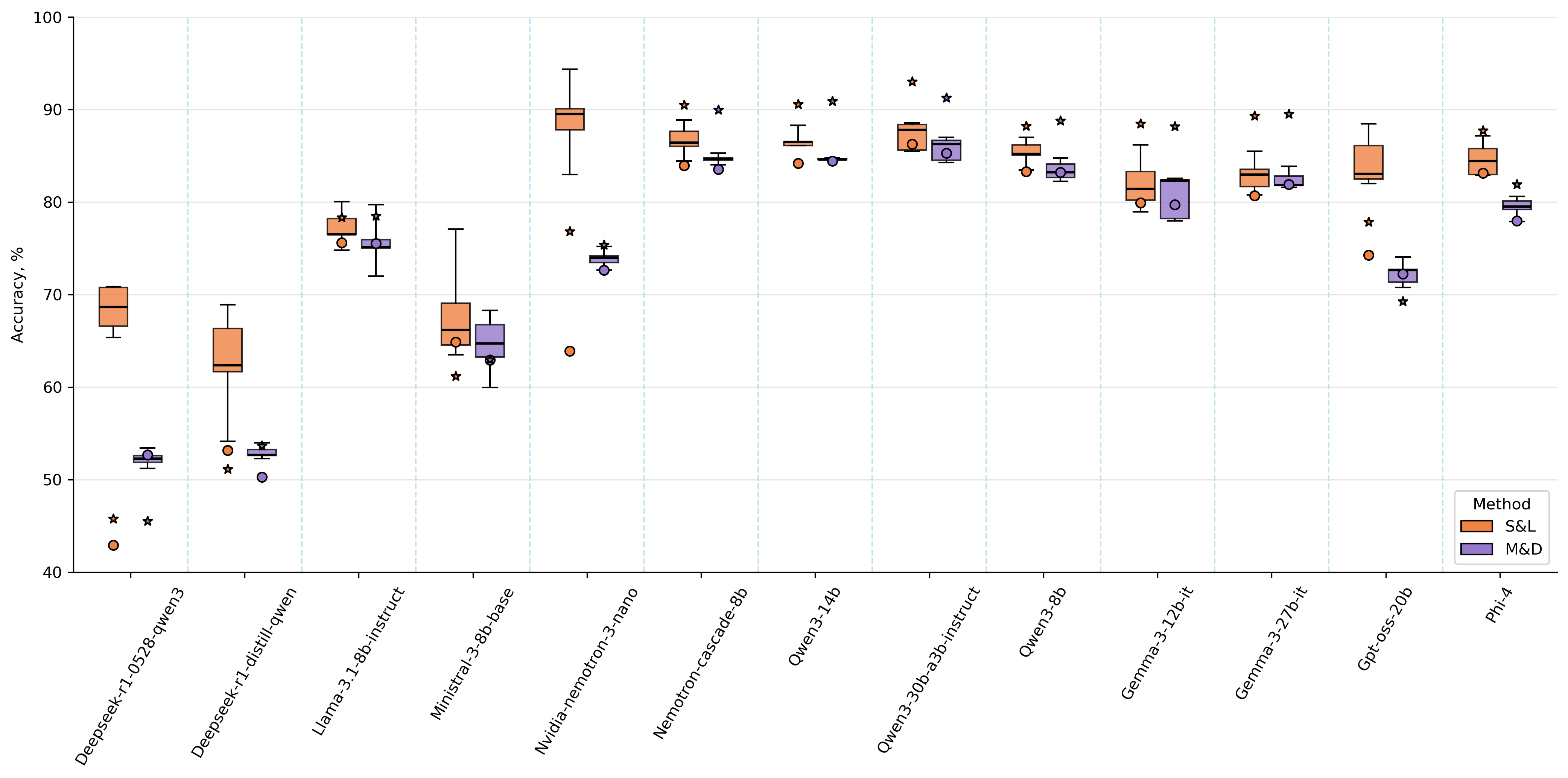}}
    \caption{Comparison of the matched prediction with dashes as labels (M\&D; our method) with standard letter prediction with letters as labels (S\&L) on CSQA \cite{talmor-etal-2019-commonsenseqa} with a 5-shot prompt. The boxes illustrate the model performance under all possibilities of ”answer-moving attacks”, where the whiskers indicate the minimum and maximum accuracy for each model. Each dot represents the performance of the original permutations. Additionally, each star symbolizes a SCORE \cite{nalbandyan-etal-2025-score} robustness metric.}
    \label{fig:CommonsenseQA-comparison}
  \end{center}
\end{figure*}

To assess the impact of our evaluation protocol on simple reasoning benchmarks, we report results on CSQA \cite{talmor-etal-2019-commonsenseqa} and ARC \cite{clark2018arc}. Figure~\ref{fig:CommonsenseQA-comparison} compares our M\&D protocol to the standard S\&L evaluation on CSQA. Across all 13 models, M\&D consistently reduces the gap between original permutation accuracy and mean accuracy under answer-moving attacks. Moreover, except for Llama-3.1, our protocol yields substantially lower variance, indicating improved robustness to label and few-shot prompt biases. These results suggest that M\&D mitigates reliance on few-shot answer distributions and label shortcuts, producing more stable performance estimates.

Figure~\ref{fig:ARC-comparison} presents analogous results on ARC. Despite the differences in the number of answers (5 vs. 4) and the domain (commonsense vs. grade-level science), the trends remain consistent – 11 of 13 models exhibit reduced accuracy variance across permutations, and the same number show a smaller gap between original and attacked accuracies under M\&D protocol compared to S\&L. This demonstrates that our evaluation protocol generalizes across reasoning MCQ benchmarks with diverse structural and topical properties.

\begin{table*}[ht]
\caption{We report the variance ratio $\sigma^2_R$ for each model, where $\sigma^2_R<1$ indicates lower variance under the M\&D evaluation protocol, whereas $\sigma^2_R>1$ indicates lower variance under the S\&L evaluation protocol. To summarize overall model-level trends, we compute the geometric mean of $\sigma^2_R$ across all datasets, denoted $\overline{x}_{GEOM}(\sigma^2_R)$. Across all datasets and models, we observe a geometric mean of variance ratio of {\color[HTML]{009901} \underline{\textbf{0.33}}}, demonstrating a $3\times$ reduction in models' accuracy variance under our protocol.}
\centering
\label{tab:var_ratio}
\scriptsize
\begin{tabular}{lcccccc}
\toprule
\multicolumn{1}{c}{\textbf{Model name}} & \textbf{$\sigma^2_{R_{CSQA}}$} & \textbf{$\sigma^2_{R_{ARC}}$} & \textbf{$\sigma^2_{R_{GPQA}}$} & \textbf{$\sigma^2_{R_{INCLUDE}}$} & \textbf{$\sigma^2_{R_{MMLU-PRO}}$} & \textbf{$\overline{x}_{GEOM}(\sigma^2_R)$} \\
\midrule
Deepseek-r1-0528-qwen3 &
  \color[HTML]{009901} 0.005 &
  \color[HTML]{009901} 0.089 &
  \color[HTML]{009901} 0.226 &
  \color[HTML]{009901} 0.033 &
  \color[HTML]{009901} 0.945 &
  \color[HTML]{009901} 0.08 \\
Deepseek-r1-distill-qwen &
  \color[HTML]{009901} 0.038 &
  \color[HTML]{009901} 0.135 &
  \color[HTML]{009901} 0.184 &
  \color[HTML]{009901} 0.118 &
  \color[HTML]{009901} 0.041 &
  \color[HTML]{009901} 0.09 \\
Llama-3.1-8b-instruct &
  1.688 &
  \color[HTML]{009901} 0.104 &
  \color[HTML]{009901} 0.645 &
  3.479 &
  2.121 &
  \color[HTML]{009901} 0.96 \\
Ministral-3-8b-base &
  \color[HTML]{009901} 0.346 &
  \color[HTML]{009901} 0.813 &
  \color[HTML]{009901} 0.658 &
  \color[HTML]{009901} 0.534 &
  \color[HTML]{009901} 0.577 &
  \color[HTML]{009901} 0.56 \\
Nvidia-nemotron-3-nano &
  \color[HTML]{009901} 0.008 &
  \color[HTML]{009901} 0.066 &
  \color[HTML]{009901} 0.073 &
  \color[HTML]{009901} 0.059 &
  \color[HTML]{009901} 0.620 &
  \color[HTML]{009901} 0.07 \\
Nemotron-cascade-8b &
  \color[HTML]{009901} 0.105 &
  \color[HTML]{009901} 0.351 &
  1.540 &
  \color[HTML]{009901} 0.154 &
  1.610 &
  \color[HTML]{009901} 0.43 \\
Qwen3-14b &
  \color[HTML]{009901} 0.007 &
  2.823 &
  3.124 &
  \color[HTML]{009901} 0.401 &
  2.811 &
  \color[HTML]{009901} 0.59 \\
Qwen3-30b-a3b-instruct &
  \color[HTML]{009901} 0.664 &
  1.082 &
  1.486 &
  \color[HTML]{009901} 0.306 &
  3.080 &
  1.00 \\
Qwen3-8b &
  \color[HTML]{009901} 0.414 &
  \color[HTML]{009901} 0.959 &
  \color[HTML]{009901} 0.385 &
  \color[HTML]{009901} 0.188 &
  1.171 &
  \color[HTML]{009901} 0.51 \\
Gemma-3-12b-it &
  \color[HTML]{009901} 0.659 &
  \color[HTML]{009901} 0.373 &
  \color[HTML]{009901} 0.096 &
  \color[HTML]{009901} 0.967 &
  \color[HTML]{009901} 0.598 &
  \color[HTML]{009901} 0.42 \\
Gemma-3-27b-it &
  \color[HTML]{009901} 0.222 &
  \color[HTML]{009901} 0.379 &
  \color[HTML]{009901} 0.871 &
  1.409 &
  \color[HTML]{009901} 0.989 &
  \color[HTML]{009901} 0.63 \\
Gpt-oss-20b &
  \color[HTML]{009901} 0.057 &
  \color[HTML]{009901} 0.068 &
  1.743 &
  \color[HTML]{009901} 0.007 &
  3.933 &
  \color[HTML]{009901} 0.18 \\
Phi-4 &
  \color[HTML]{009901} 0.396 &
  \color[HTML]{009901} 0.748 &
  \color[HTML]{009901} 0.203 &
  \color[HTML]{009901} 0.307 &
  \color[HTML]{009901} 0.763 &
  \color[HTML]{009901} 0.43 \\ \hline
\textbf{$\overline{x}_{GEOM}(\sigma^2_R)$} &
  \color[HTML]{009901} 0.11 &
  \color[HTML]{009901} 0.33 &
  \color[HTML]{009901} 0.49 &
  \color[HTML]{009901} 0.23 &
  \color[HTML]{009901} 0.99 &
  \textbf{\color[HTML]{009901} \underline{\textbf{0.33}}}\\ \hline
\end{tabular}%
\end{table*}

\subsection{Across languages: INCLUDE.}
To evaluate whether MCQ-induced biases persist in non-English settings, we report results on the INCLUDE dataset using a subset of languages explicitly used during model training – Spanish, French, Italian, and German. The results are shown in Figure~\ref{fig:INCLUDE-comparison} and Table~\ref{tab:INCLUDE} in the Appendix~\ref{appendix:detailed_results}.

We adopt the same evaluation protocol as in the English-only experiments, following \cite{romanouinclude}, which uses a mixed-language setup with an English prompt, followed by questions and answer options in the original language, which was shown to improve performance across models. 

Consistent with English benchmarks, most models show reduced variance across permutations and a smaller gap between original and attack-averaged accuracy. This suggests that label and prediction-mode artifacts are not language-specific, but generalize across multilingual settings.

\subsection{Larger number of options: MMLU-Pro.}

While most benchmarks include only four to five answer options, we evaluate MMLU-Pro \cite{wang2024mmlu} to test the applicability of our method in settings with more closely aligned distractors. MMLU-Pro contains approximately 14,000 challenging questions across diverse academic domains, each with ten answer options. We report results in Figure~\ref{fig:MMLU-comparison} and Table~\ref{tab:MMLU-Pro} in the Appendix~\ref{appendix:detailed_results}, using the same evaluation protocol as for ARC and CSQA.

Unlike other benchmarks, many models achieve their highest accuracy under the original permutation under both S\&L and M\&D evaluations, with even minor answer-position changes causing substantial performance drops. This indicates a strong reliance on positional biases, particularly among models that also exhibit explicit or implicit bias exploitation on NonsenseQA.

Importantly, this behavior reflects structural properties of MMLU-Pro rather than a limitation of the proposed protocol. For models less dependent on such shortcuts, our method still reduces accuracy variance in 7 of 13 cases, demonstrating that it selectively exposes MCQ-specific brittleness without uniformly degrading robustness or performance.

\subsection{Difficult questions: GPQA.}

Finally, to show the applicability of our evaluation protocol to the most difficult answers, we extend the evaluation to the GPQA dataset. Unlike the previous benchmarks, for the GPQA dataset, we do not alter the few-shot answer distribution under answer-moving attacks. We decided to preserve the original prompt distribution to demonstrate that our method is still applicable, even when the few-shot prompt answers do not follow a malicious permutation.

Consistent with previous findings, as shown in Figure~\ref{fig:GPQA-comparison} and Table~\ref{tab:GPQA} in Appendix~\ref{appendix:detailed_results}, our protocol reduces the variance between the accuracies in 9 out of 13 models. Furthermore, it reduces the gap between the original permutation accuracy and the mean answer-moving attack accuracy in 8 out of 13 models.

\subsection{SCORE alone overlooks evaluation biases.}
Across most benchmarks, models evaluated under the S\&L protocol achieve higher SCORE values than under the M\&D protocol. This behavior is expected: biases in the few-shot answer distributions enable models to produce highly consistent predictions by exploiting MCQ shortcuts. As a result, S\&L evaluation can overstate model robustness by rewarding shortcut-driven consistency rather than stable reasoning.

For example, two models that achieve identical SCORE may behave fundamentally differently. If one model exploits permutation bias for perfect accuracy in 10 out of 11 runs while completely failing on the original permutations, and the other predicts correctly with 90\% probability across all runs, both would achieve a SCORE of 0.82.

We therefore supplement the SCORE metric with the variance ratio metric defined in Eq.~\ref{eq:var_ratio}, which better separates these behaviors. In the example above, the variance ratio is $\sim0.001$, a $900\times$ reduction, clearly distinguishing shortcut exploitation from genuine consistency. As reported in Table~\ref{tab:var_ratio}, across most models and benchmarks, M\&D yields consistently lower variance, confirming that our protocol suppresses bias-driven prediction shortcuts.

As for the SCORE metric alone, only two models, Ministral-3 \cite{liu2026ministral} and Gemma-3 \cite{team2025gemma}, achieve a higher SCORE on both the MMLU-Pro and the ARC benchmarks. In contrast, under GPQA, where few-shot distributions are not manipulated, 10 of 13 models show improved SCORE under M\&D, clearly demonstrating that robustness gains stem from removing exploitable MCQ structure rather than few-shot artifacts alone. However, across all these datasets, we observe that majority of models achieve a variance ratio below one. We expand on the influence of biases on the SCORE metric in Appendix~\ref {appendix:score_paradox}. 

\subsection{Cross-Benchmark Agreement}
\begin{figure*}[t]
    \centering
    \begin{subfigure}[b]{0.4\textwidth}
        \centering
        \includegraphics[width=\columnwidth]{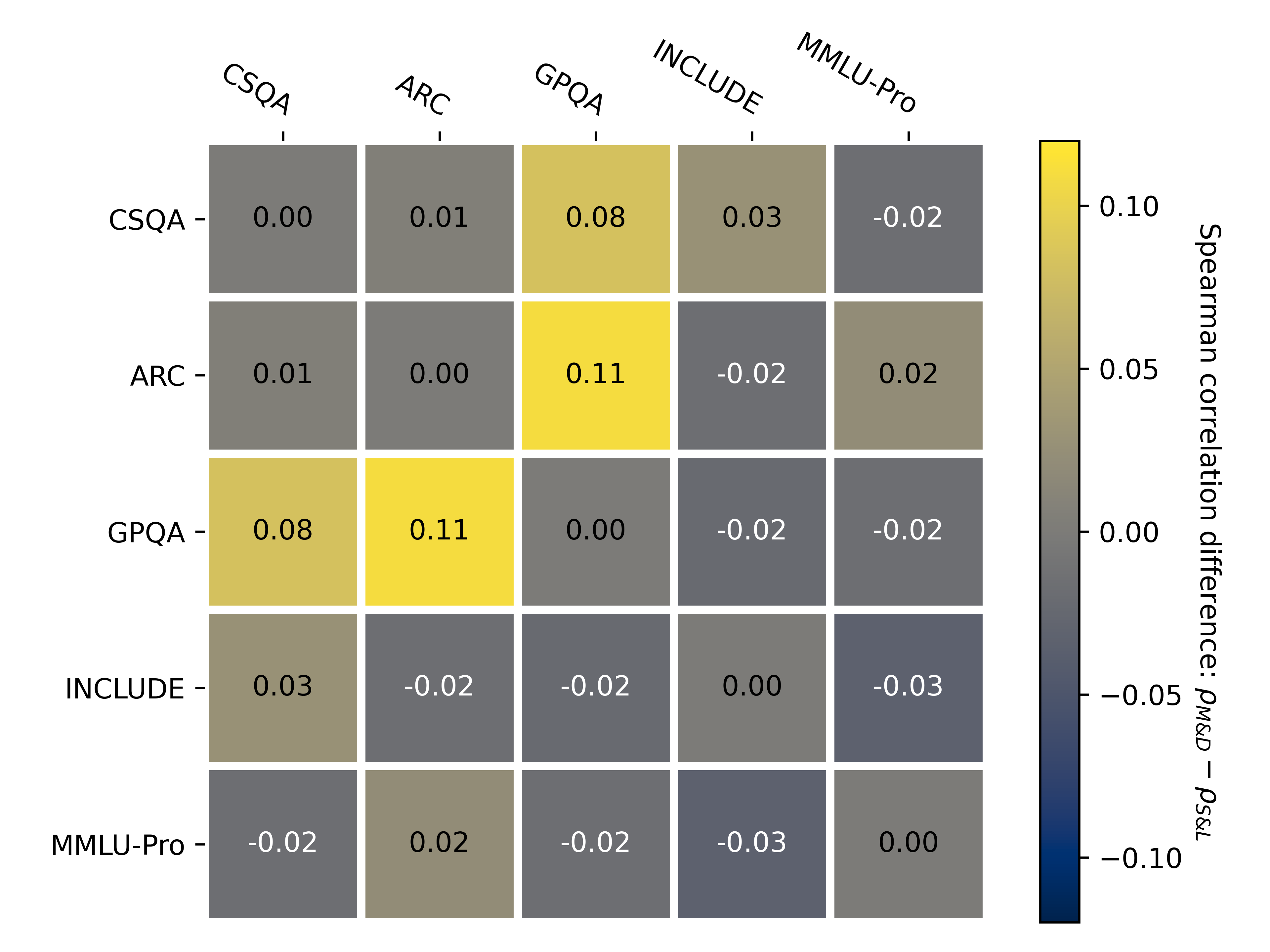}
        \caption{Spearman cross-benchmark correlation difference between $\rho_{M\&D}-\rho_{S\&L}$. A positive value indicates a higher M\&D cross-benchmark correlation, while a negative - higher S\&L correlation.}
        \label{fig:spearman-diff}
    \end{subfigure}
    \hfill
    \begin{subfigure}[b]{0.4\textwidth}
        \centering
        \includegraphics[width=\columnwidth]{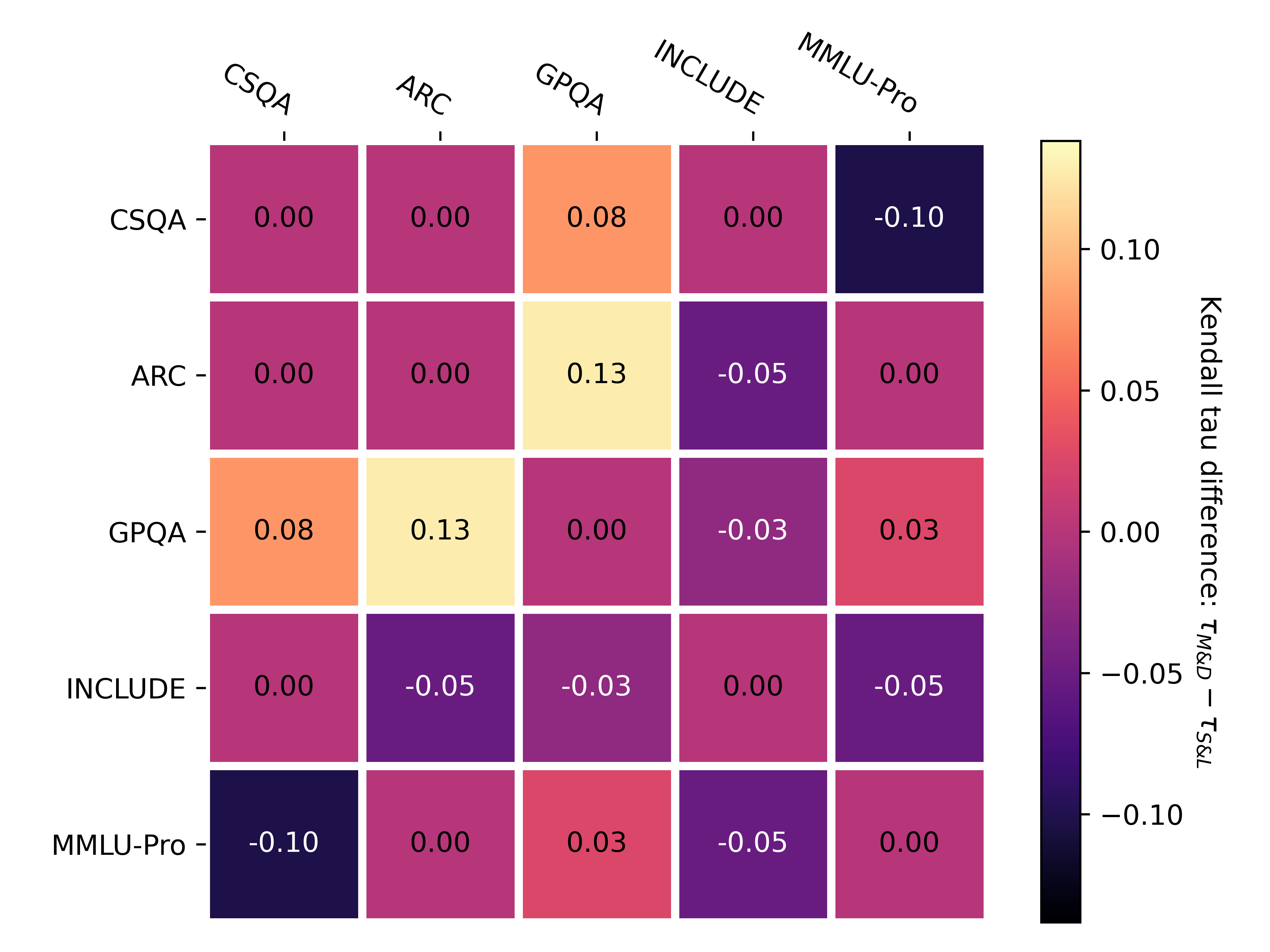}
        \caption{Kendall tau cross-benchmark rank-agreement difference $\tau_{M\&D} - \tau_{S\&L}$. A positive value indicates a higher M\&D cross-benchmark rank agreement, while a negative - higher S\&L rank agreement.}
        \label{fig:tau-diff}
    \end{subfigure}
    \caption{Cross-benchmark Spearman and Kendall Tau rank correlation agreement}
    \label{fig:cross_benchmark_agreement}
\end{figure*}

To assess whether our protocol preserves meaningful benchmark relationships, we compute Spearman ($\rho$) and Kendall ($\tau$) rank correlations between benchmark pairs under both M\&D and S\&L protocols, reporting the differences as e.g., $\rho_{M\&D} - \rho_{S\&L}$. We provide a formal definition of these statistics in Appendix~\ref{appendix:tau_and_spearman} and demonstrate the Spearman and Kendall Tau differences in ~\cref{fig:spearman-diff,fig:tau-diff}, respectively.

As shown in Figure~\ref{fig:spearman-diff}, most cross-benchmark correlations remain stable across protocols. However, reasoning-focused pairs, GPQA–ARC and GPQA–CSQA, show increased agreement under M\&D, suggesting that once evaluation biases are reduced, performance on challenging reasoning tasks aligns more closely with simpler benchmarks.

This trend is reinforced by Figure~\ref{fig:tau-diff}, which shows increased rank consistency between GPQA and ARC, CSQA, and MMLU-Pro under M\&D. This is intuitive, as these benchmarks primarily assess reasoning and academic knowledge, with GPQA representing the most challenging setting.

In contrast, INCLUDE shows decreased agreement with English benchmarks under M\&D, revealing that strong English-language performance does not necessarily translate to comparable multilingual reasoning ability—a relationship previously masked by shared evaluation artifacts

Finally, reduced CSQA–MMLU-Pro agreement under M\&D suggests MMLU-Pro retains knowledge-driven questions requiring minimal reasoning from MMLU \cite{wang2024mmlu}, leading to divergent rankings once biases are removed. Overall, these results show that the proposed protocol exposes meaningful structural differences between benchmarks that are masked under standard evaluation.

\subsection{Ablation studies}
\textbf{Sentence similarity model and function.}
We evaluated various sentence similarity models and functions and did not observe major differences, therefore we use a small model for efficiency (Appendix \ref{appendix:ablation_similairiy}).

\textbf{Option label symbols.}
We evaluated various combinations of option labels $\mathcal{L}$ both homogeneous and heterogeneous, indicating that homogeneous labeling improves evaluation stability over heterogeneous labeling, and that symbols might carry prior semantic associations that can lead to increase variance. (Appendix \ref{appendix:ablation_symbols}).
\section{Conclusion}
MCQ benchmarks are a standard tool for evaluating state-of-the-art LLMs, yet common evaluation protocols introduce artifacts that models can exploit without genuine reasoning. Beyond well-studied position and label biases, we demonstrate how the additional few-shot answer distribution and prediction-mode biases impact the evaluation.

To diagnose these effects, we introduce NonsenseQA, a simple diagnostic benchmark that isolates model behavior under controlled changes to few-shot distributions, labeling schemes, and prediction modes. Guided by these findings, we propose a bias-reduced evaluation standard that preserves the MCQ structure while mitigating these artifacts.

Across diverse models and benchmarks, our approach of using homogeneous labels, full-text answer generation, and semantic similarity-based matching yields more stable and informative evaluations than standard MCQ scoring. In particular, it improves cross-benchmark agreement and more clearly differentiates models by reasoning, simple knowledge retrieval, and multilingual capabilities. We hope this work encourages more careful treatment of evaluation artifacts and advances more robust assessment of LLMs.

\section*{Broader societal impact statement}

The ubiquitous adoption of AI systems and large language models raises important concerns regarding biases and how models may exploit evaluation artifacts when answering questions. 

Let’s consider an example of a professional relying on an LLM to validate a high-stakes decision. When such a query is presented in a multiple-choice format with few-shot prompting, model outputs may be driven by label biases or patterns in the previous examples rather than by genuine reasoning, potentially leading to serious consequences.

Our benchmark and evaluation protocol aim to expose these behaviors by influencing model choices and reducing performance variance attributable to evaluation artifacts, thereby isolating errors that stem from true gaps in model knowledge or reasoning. Widespread adoption of our proposed evaluation protocol and our NonsenseQA benchmark as a diagnostic tool could serve as a practical guiderail, enabling more consistent and reliable LLM evaluation with minimal additional computational cost, and helping practitioners identify models that rely least on few-shot and label-induced biases.

\section*{Acknowledgment}
This research was funded by the Defense Advanced Research Projects Agency (DARPA), under contract W912CG23C0031.

\newpage

\bibliography{references}
\small
\bibliographystyle{plainnat}

\newpage
\appendix
\onecolumn

\startcontents[sections]
\printcontents[sections]{l}{1}{\setcounter{tocdepth}{2}}

\section{Selected datasets} \label{appendix:datasets}
The evaluation of large language models relies on a diverse set of benchmarks designed to probe different reasoning capabilities. In the domain of question answering grounded in common prior knowledge, CommonsenseQA \cite{talmor-etal-2019-commonsenseqa} (CSQA) assesses whether models can reason about simple, everyday scenarios. The AI2 Reasoning Challenge \cite{clark2018arc} (ARC) targets deeper logical reasoning, going beyond surface-level pattern matching and factual recall to require scientific reasoning. To further reduce the likelihood of correct answers by chance and to emphasize deliberate, multi-domain reasoning at the college level, MMLU-Pro \cite{wang2024mmlu} extends the original MMLU  \cite{hendrycksmeasuring} benchmark by filtering out knowledge-heavy questions with minimal reasoning requirements and increasing the number of distractors per question. Finally, GPQA \cite{rein2024gpqa} is designed to test advanced reasoning by introducing PhD-level, Google-proof questions. Despite these increasing levels of difficulty and expanded answer sets, we demonstrate that all of these benchmarks remain susceptible to shortcut exploitation by LLMs, indicating that high performance does not necessarily reflect genuine reasoning. Finally, to test the transferability of these biases to other languages, we test our evaluation protocol on the INCLUDE \cite{romanouinclude} dataset, a comprehensive knowledge- and reasoning-centric benchmark, which includes questions in 44 languages and focuses on MCQ questions extracted from academic and professional exams, covering 57 topics, including regional knowledge. By applying our proposed evaluation protocol, we mitigate these artifacts and substantially improve the robustness of these benchmarks, without any modifications to any of the benchmarks themselves, allowing a more faithful evaluation of the model’s reasoning abilities.

\section{On the advantages of sentence similarity over cloze evaluation.} \label{appendix:sentence_sim_over_cloze}
A sentence similarity model addresses the shortcomings of a standard cloze evaluation in MCQ question answering.
The sentence similarity produces an embedding of a fixed size and allows for chain-of-thought reasoning and for semantically similar answers to be accepted.
On the other hand, cloze-style evaluation needs to normalize the predicted score by the token length of the possible answer and can only evaluate the logits for the tokens predicted right after the prompt.
Moreover, cloze-style evaluation does not compute logits for semantically similar answers.

\section{The different types of models based on few-shot and label biases} \label{appendix:models_types}
We provide examples of models falling under the three models types identified in \cref{section:nonsenseqa}.
(Type 1) Explicit bias models: use answers to the few-shot
prompt questions to reason about the provided question: GPT-OSS.
(Type 2) Implicit bias models: use answers to the few-shot
prompt questions to reason about the provided answer, illustrated by Qwen3-8B.
(Type 3) Models unable to exploit the bias: only marginally
exploit the few-shot bias implicitly, illustrated by Gemma-3-27b-it.

\begin{figure*}[!ht]
  \begin{center}
    \centerline{\includegraphics[width=\textwidth]{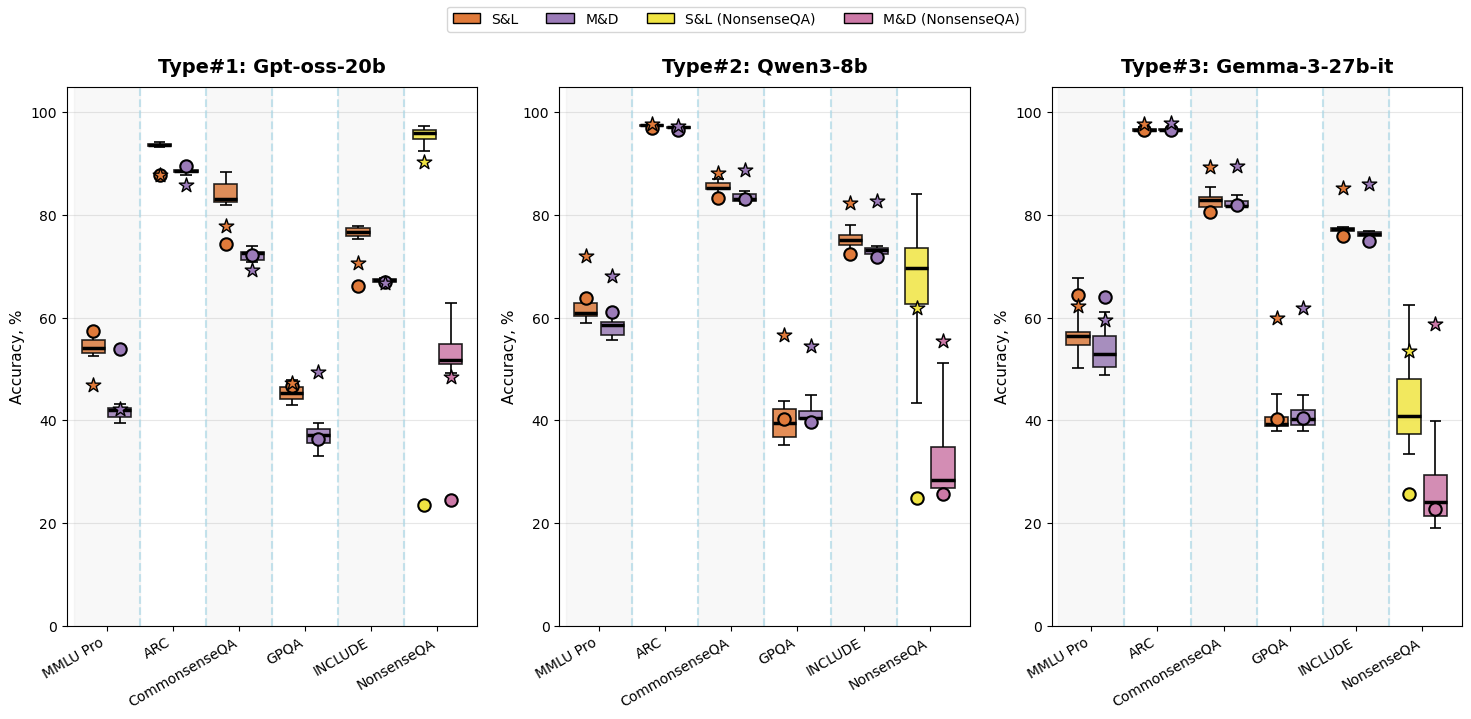}}
    \caption{We identify three categories of models: those that explicitly exploit biases present in the few-shot prompt (e.g., GPT-OSS \cite{agarwal2025gpt}), those that implicitly rely on such biases (e.g., Qwen3-8B \cite{yang2025qwen3}), and those that fail to reliably leverage few-shot prompt bias (e.g., Gemma-3-27b \cite{team2025gemma}). For each category, we provide representative examples and report their performance across all benchmarks. Model categorization is determined using performance on NonsenseQA, shown in the last column, where the expected performance should be close to chance-level (25\%).}
    \label{fig:model_types}
  \end{center}
\end{figure*}

\section{Ablation study}
\label{appendix:ablation}

We varied the sentence similarity model and function to evaluate their impact \cref{appendix:ablation_similairiy}, and considered different types of symbols and the effect of symbol heterogeneity and homogeneity \cref{appendix:ablation_symbols}.

\subsection{The choice of sentence similarity model and function has limited impact.} \label{appendix:ablation_similairiy}

\begin{figure}[!ht]
  \begin{center}
    \centerline{\includegraphics[width=\columnwidth]{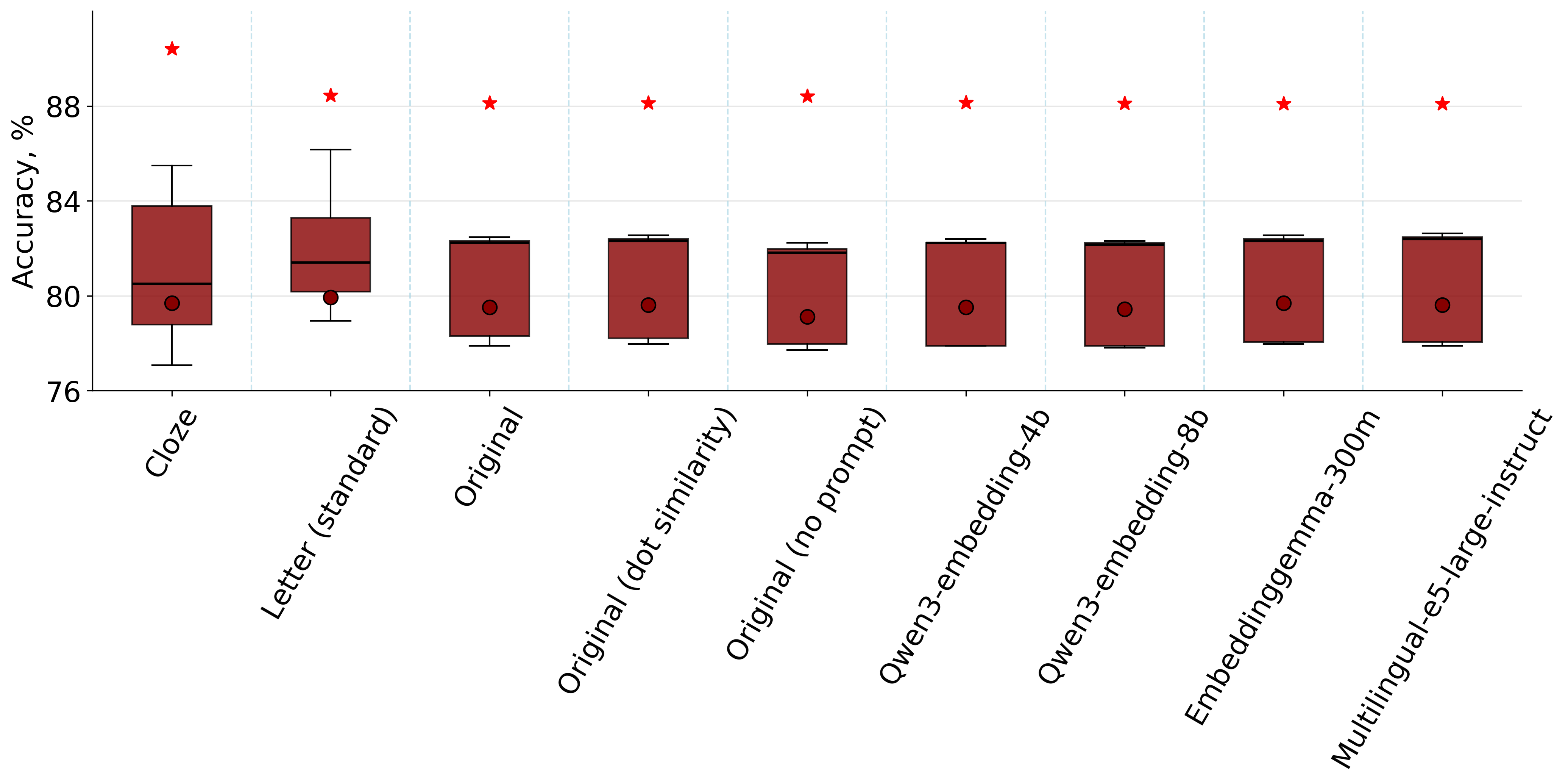}}
    \caption{Ablation study on the effect of the similarity model and the similarity function. For the ablation study, we compare standard cloze-style evaluation ("Cloze") with the letter prediction and matching ("Letter (standard)") to various full-sentence prediction outputs. We compare the original setting of the Qwen3-Embedding-0.6B \cite{zhang2025qwen3} model tested under varying conditions ("Original" experiments) with different model types and sizes.}
    \label{fig:ablation_models}
  \end{center}
\end{figure}

To motivate the design choices of our evaluation protocol, we ablate the effect of both the sentence similarity model and the similarity function on the CommonsenseQA benchmark, with the results demonstrated in Figure~\ref{fig:ablation_models}. We compare our approach against two baselines: the cloze-style evaluation \cite{robinsonleveraging, zheng2024large} and the standard letter-based MCQ evaluation. The “Original” configuration corresponds to the one proposed in the main body of the paper, using the Qwen3-Embedding-0.6B model with cosine similarity. 

Across all ablations, we observe that variations in similarity model size, architecture, prompt formulation, and similarity function lead to only minor changes in evaluation behavior. In particular, the stability of model performance under answer-moving attacks remains largely unaffected by these design choices. However, all ablated variants consistently exhibit greater robustness to answer-position perturbations than both the cloze-style and standard MCQ evaluations. These results indicate that the robustness gains arise primarily from the proposed evaluation protocol itself, rather than from specific choices of similarity model or similarity metric.

\subsection{Impact of the symbols used and the non-uniformity} \label{appendix:ablation_symbols}
\begin{figure}[!ht]
  \begin{center}
    \centerline{\includegraphics[width=\columnwidth]{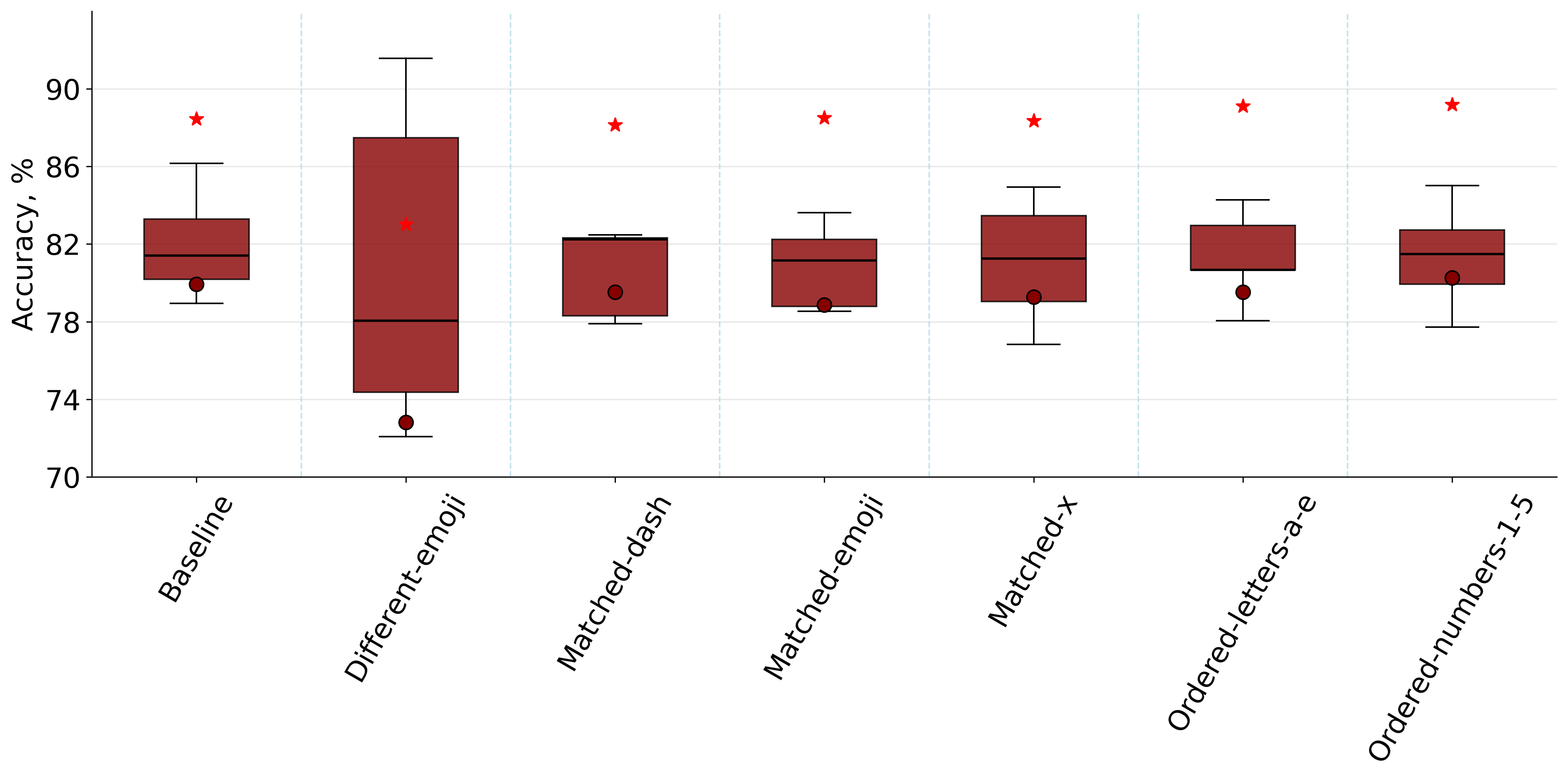}}
    \caption{Ablation study on the effect of different labels used in the proposed matching protocol. For the ablation study, we compare standard letter prediction and matching ("Baseline") to various label settings for the matching model. We compare using homogeneous per-option labels ( like dash ("Matched-dash"), a "\twemoji{1f643}" ("Matched-emoji"), and "X" ("Matched-x") ) to using heterogeneous labels (e.g, (\twemoji{1f641}/\twemoji{1f642}/\twemoji{1f643}/\twemoji{1f644}/\twemoji{1f636}) ("Different-emoji"), (A/B/C/D/E), and (1/2/3/4/5) ).}
    \label{fig:ablation_letters}
  \end{center}
\end{figure}

Similarly, we perform an ablation study on the choice of option labels used in the Matched evaluation protocol, as shown in Figure~\ref{fig:ablation_letters}. As before, we compare our approach against the cloze-style evaluation and the standard letter-based MCQ evaluation. We consider three homogeneous label sets and three heterogeneous label sets.

For homogeneous configurations, we use the same label for all options, using either "\twemoji{1f643}", a dash, or the symbol “X”. For heterogeneous configurations, we use distinct labels for each option, using either (\twemoji{1f641}/\twemoji{1f642}/\twemoji{1f643}/\twemoji{1f644}/\twemoji{1f636}), (1/2/3/4/5), or (A/B/C/D/E). Across these settings, we observe that homogeneous labeling improves evaluation stability relative to heterogeneous labeling. In particular, the emoji and dash labels exhibit comparable robustness. However, the “X” symbol leads to noticeably greater variance.

This behavior could indicate that labels carrying prior semantic associations, such as “X,” which is commonly used to denote incorrect answers or unknowns, can introduce unintended biases into the evaluation. Overall, these results indicate that minimizing unintended semantic content in option labels is beneficial for reducing variance and improving robustness in MCQ-based evaluations.

\section{NonsenseQA - Bias Decomposition}
\label{appendix:nonsenseQA}

To measure the difference between S\&L and our proposed M\&D, we evaluate all the possible combinations (S\&L, S\&D, M\&L, and M\&D) on NonsenseQA, and we show that our uniform label and matched full-text option prediction give the biggest randomness. In \cref{fig:SL-comparison,fig:ML-comparison,fig:SD-comparison,fig:MD-comparison}, we demonstrate that the mode of prediction (S vs. M) has a bigger impact on the bias than the label used (D vs. L). Nevertheless, we find both design choices synergistic, achieving the least biased predictor - M\&D. To simplify all the comparisons, we gather all plots in Figure~\ref{fig:full-comparison}.

\subsection{Standard \& Letter}

\begin{figure*}[!ht]
  \begin{center}
    \centerline{\includegraphics[width=\textwidth]{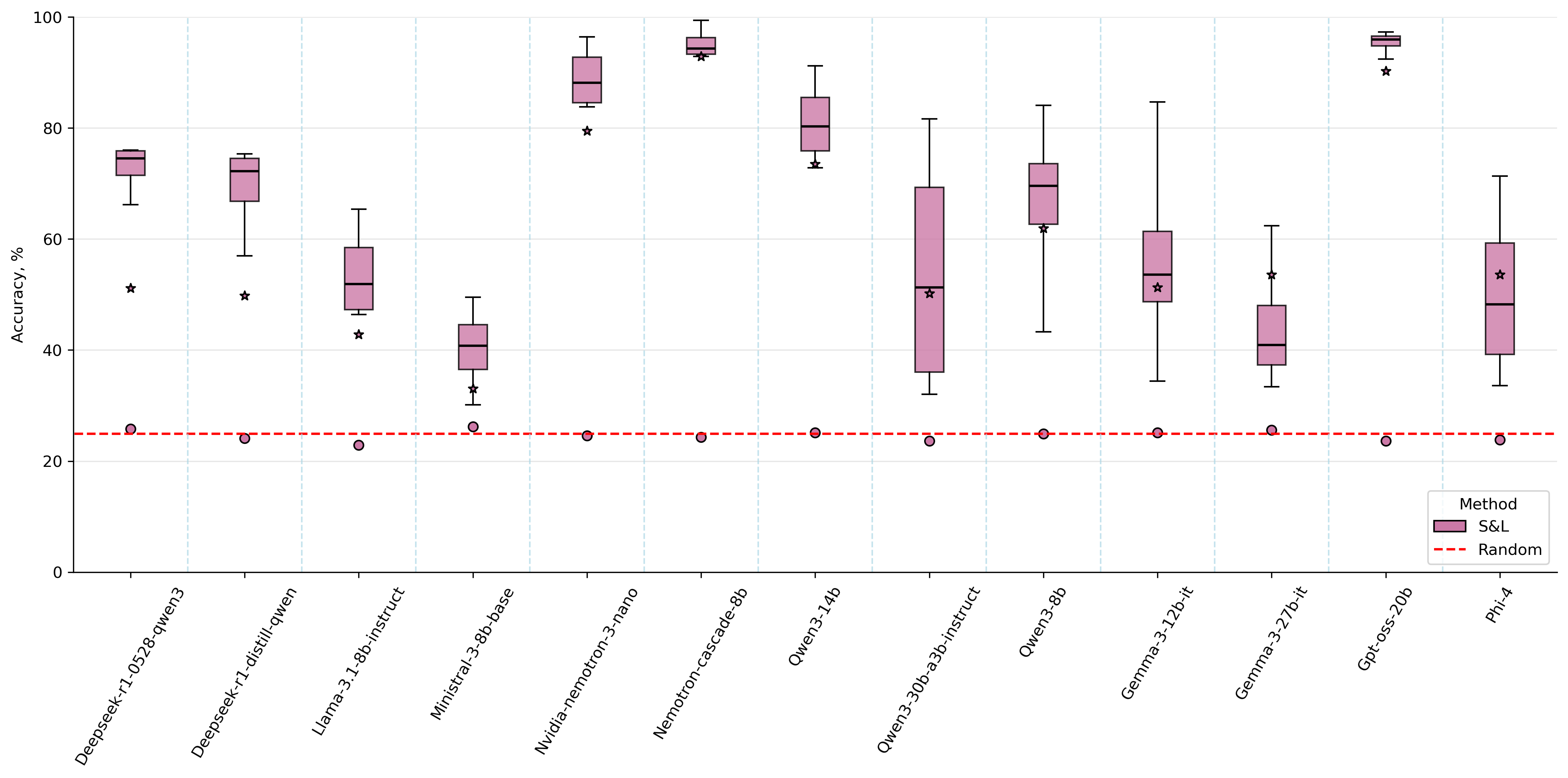}}
    \caption{The performance of Standard-And-Letter (S\&L) on our NonsenseQA dataset with a 5-shot prompt. The boxes illustrate the model performance under all possibilities of ”answer-moving attacks”, where the whiskers indicate the minimum and maximum accuracy for each model. Each dot represents the performance of the original permutations. Additionally, each star symbolizes a SCORE \cite{nalbandyan-etal-2025-score} robustness metric.}
    \label{fig:SL-comparison}
  \end{center}
\end{figure*}

As demonstrated in Figure~\ref{fig:SL-comparison}, none of the models achieve an accuracy of 25\% in all their runs, with some models consistently achieving an accuracy of 95\% and a tight model variance spread. As for the original, random permutation on NonsenseQA, all models achieve an accuracy of around 25\%. This demonstrated that, within our evaluation setup, models can freely exploit the few-shot prompt and label bias.

\subsection{Standard \& Dash}
\begin{figure*}[!ht]
  \begin{center}
    \centerline{\includegraphics[width=\textwidth]{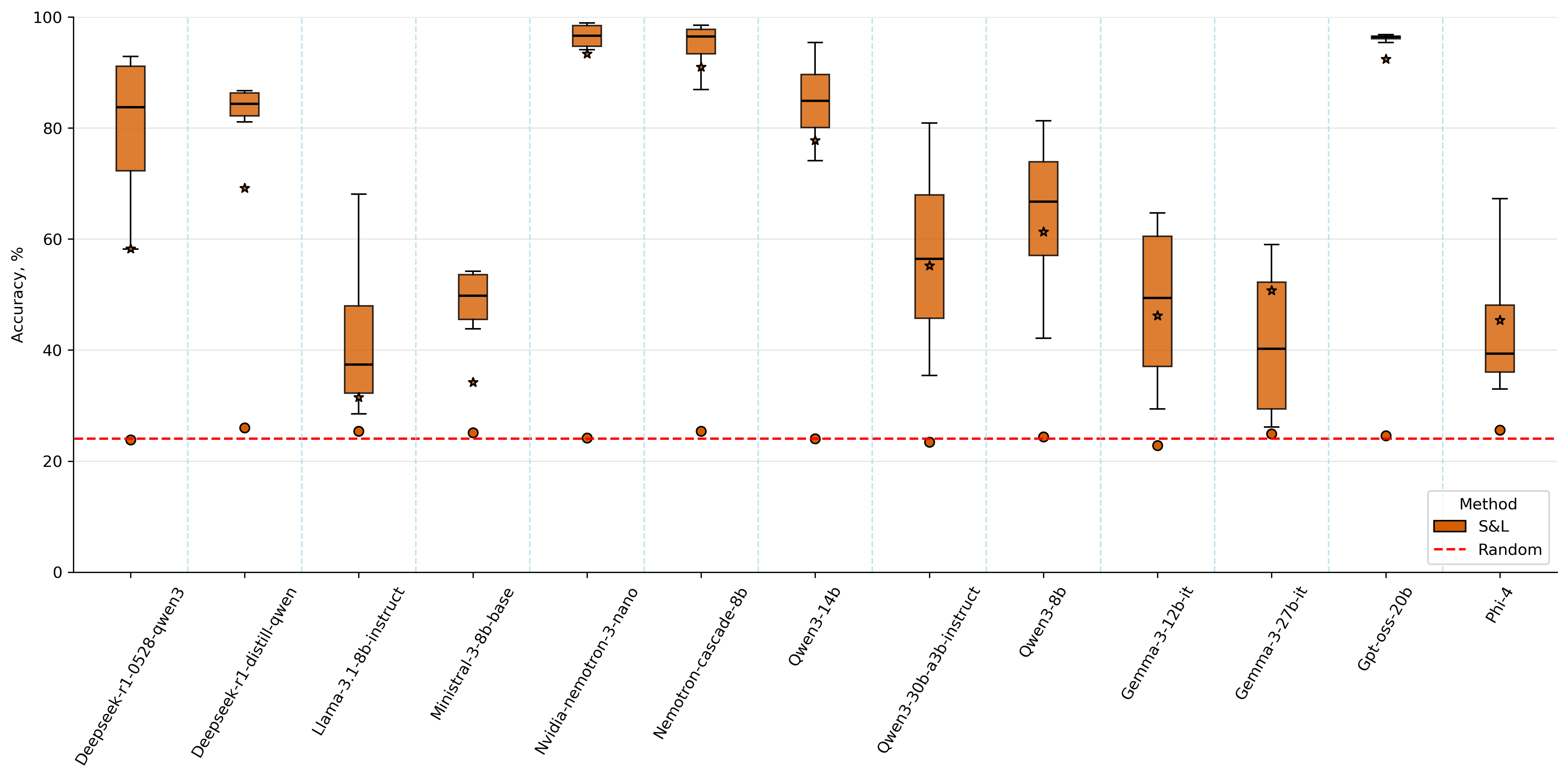}}
    \caption{The performance of Standard-And-Dash (S\&D) on our NonsenseQA dataset with a 5-shot prompt. The boxes illustrate the model performance under all possibilities of ”answer-moving attacks”, where the whiskers indicate the minimum and maximum accuracy for each model. Each dot represents the performance of the original permutations. Additionally, each star symbolizes a SCORE \cite{nalbandyan-etal-2025-score} robustness metric.}
    \label{fig:SD-comparison}
  \end{center}
\end{figure*}

Compared to the S\&L evaluation protocol shown in Figure~\ref{fig:SL-comparison}, S\&D exhibits a noticeable reduction in bias, as shown in Figure~\ref{fig:SD-comparison}. Although several models still achieve relatively high mean accuracy, some permutations approach chance-level performance of 25\%, and a subset of models exhibits an overall lower mean accuracy. This suggests that removing distinct option labels mitigates a portion of the evaluation artifacts. However, the remaining performance inflation indicates that biases induced by the prediction mode and few-shot answer distribution are more influential than label-specific biases alone. In particular, models like Qwen3-14B and GPT-OSS-20B continue to achieve consistently high mean accuracy under S\&D, highlighting that label homogenization is insufficient to suppress few-shot shortcut exploitation.

\subsection{Matched \& Letter}
\begin{figure*}[!ht]
  \begin{center}
    \centerline{\includegraphics[width=\textwidth]{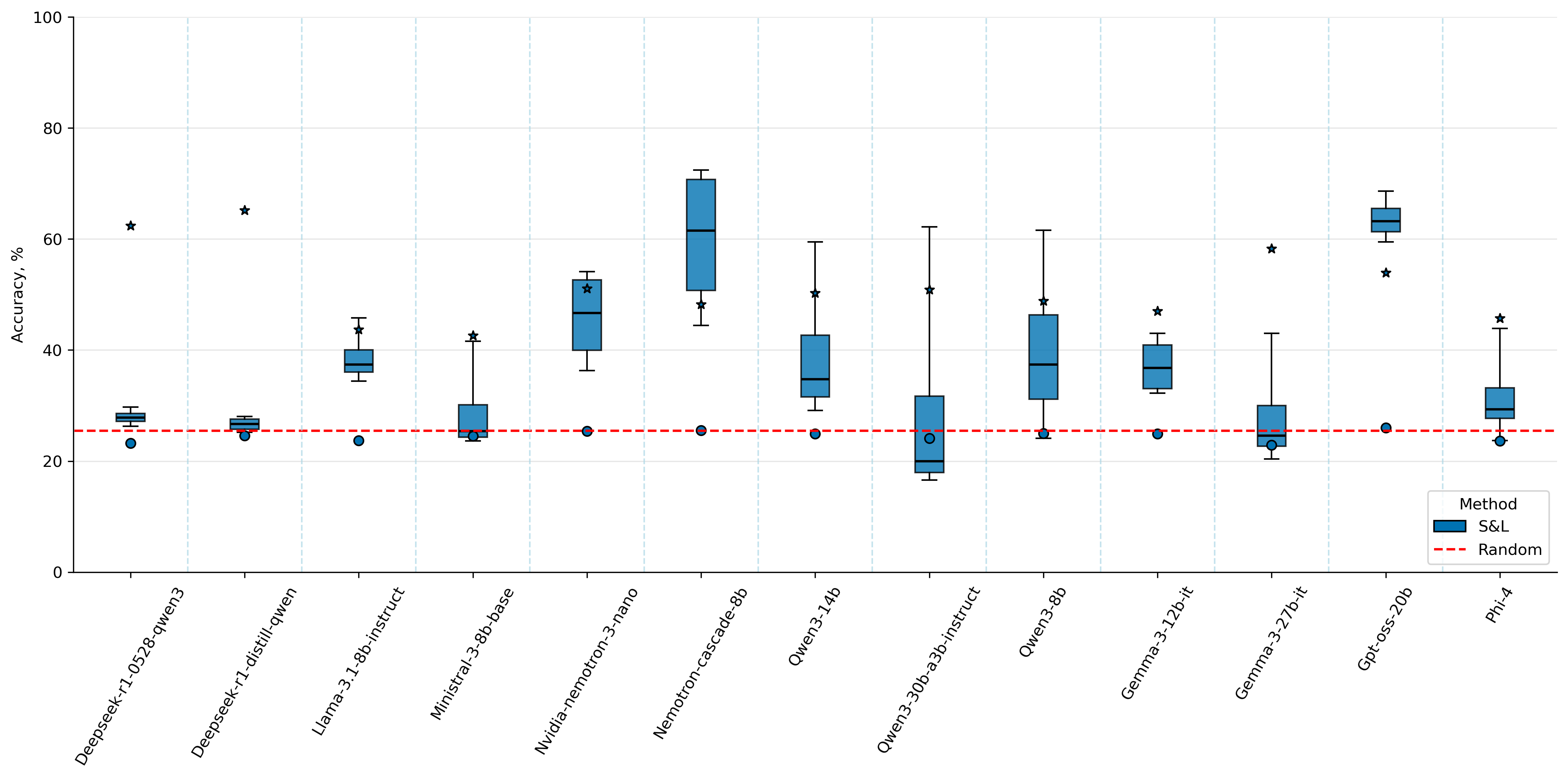}}
    \caption{The performance of Matched-And-Letter (M\&L) on our NonsenseQA dataset with a 5-shot prompt. The boxes illustrate the model performance under all possibilities of ”answer-moving attacks”, where the whiskers indicate the minimum and maximum accuracy for each model. Each dot represents the performance of the original permutations. Additionally, each star symbolizes a SCORE \cite{nalbandyan-etal-2025-score} robustness metric.}
    \label{fig:ML-comparison}
  \end{center}
\end{figure*}

To isolate the effect of prediction mode while preserving letter-based option labels, we evaluate model performance on NonsenseQA under the M\&L protocol and report accuracy under answer-moving attacks in Figure~\ref{fig:ML-comparison}. Relative to the S\&L results shown in Figure~\ref{fig:SL-comparison}, M\&L yields a substantial reduction in mean under-attack accuracy, indicating that models are less able to exploit answer distributions present in the few-shot prompt when required to generate full-text answers. Notably, six of the thirteen evaluated models achieve a mean accuracy close to the chance level, reflecting effectively random behavior on random inputs. However, several models (e.g, Nemotron-Cascade-8B and GPT-OSS-20B) never approach chance-level accuracy across any of their permutations, suggesting that some biases persist even when prediction-mode shortcuts are partially mitigated.

\subsection{Matched \& Dash}

\begin{figure*}[!ht]
  \begin{center}
    \centerline{\includegraphics[width=\textwidth]{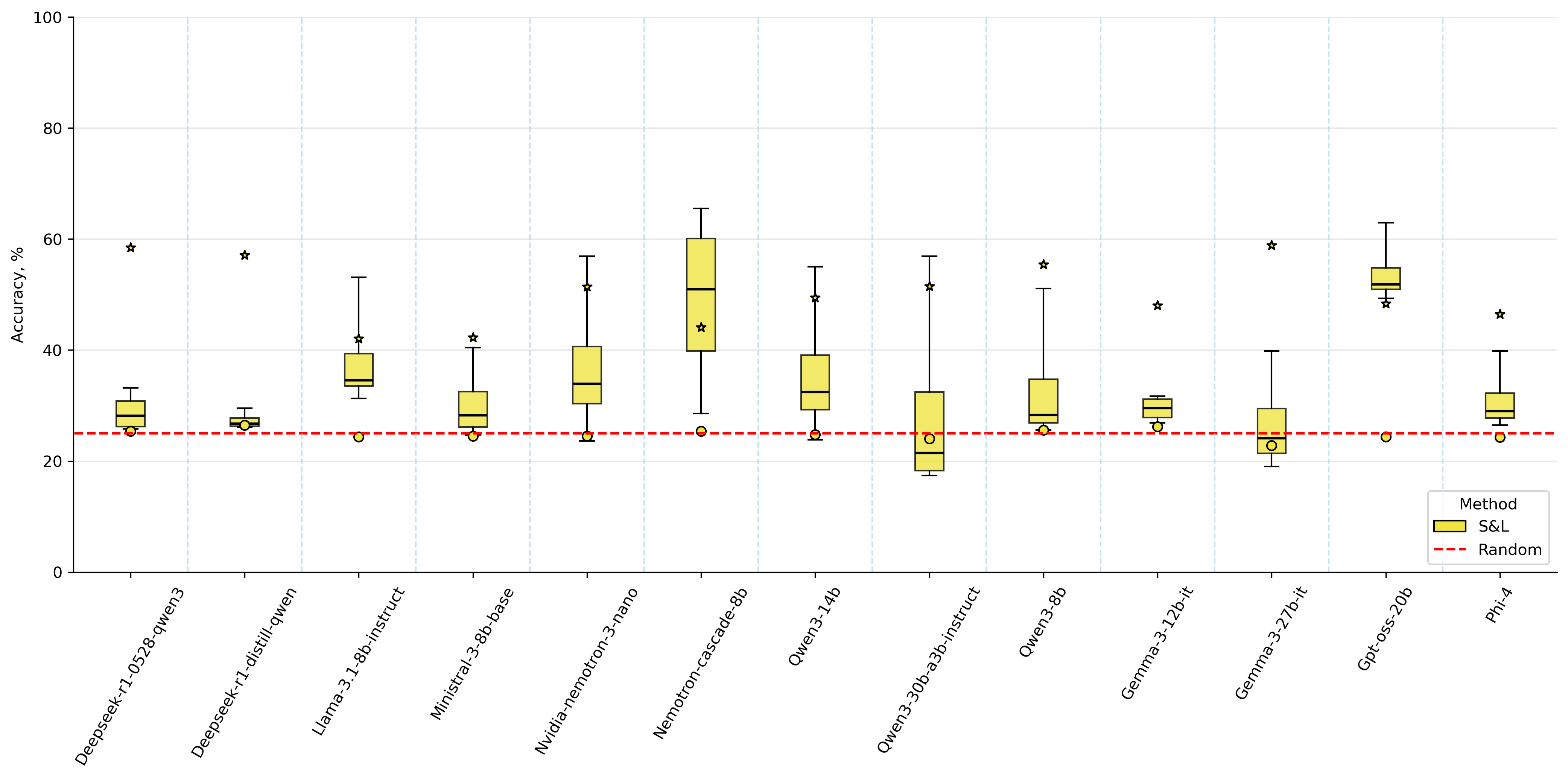}}
    \caption{The performance of Matched-And-Dash (M\&D) on our NonsenseQA dataset with a 5-shot prompt. The boxes illustrate the model performance under all possibilities of ”answer-moving attacks”, where the whiskers indicate the minimum and maximum accuracy for each model. Each dot represents the performance of the original permutations. Additionally, each star symbolizes a SCORE \cite{nalbandyan-etal-2025-score} robustness metric.}
    \label{fig:MD-comparison}
  \end{center}
\end{figure*}

Finally, we demonstrate the performance of our method, M\&D, on NonsenseQA in Figure~\ref{fig:MD-comparison}. Compared to the previous design choices presented in \cref{fig:SL-comparison,fig:SD-comparison,fig:ML-comparison}, we observe that a full-text prediction mode with homogenous option labels leads to the most chance-like performance, with eight models achieving a mean under-attack accuracy of 25\%. Moreover, we observe an increase in the number of models achieving an accuracy of 25\% within one of their permutations, with 11 out of 13 models falling into that category. Nevertheless, two models - Llama 3.1 and GP-OSS-20B - still achieve a mean performance high above 25\% with none of the permutations reaching a chance-like score. This demonstrates that some biases are still present within these models; however, under all measured evaluation protocols, M\&D achieves the lowest mean under-attack accuracy for these models as well.

\subsection{Full comparison}

\begin{figure*}[!ht]
  \begin{center}
    \centerline{\includegraphics[width=\textwidth]{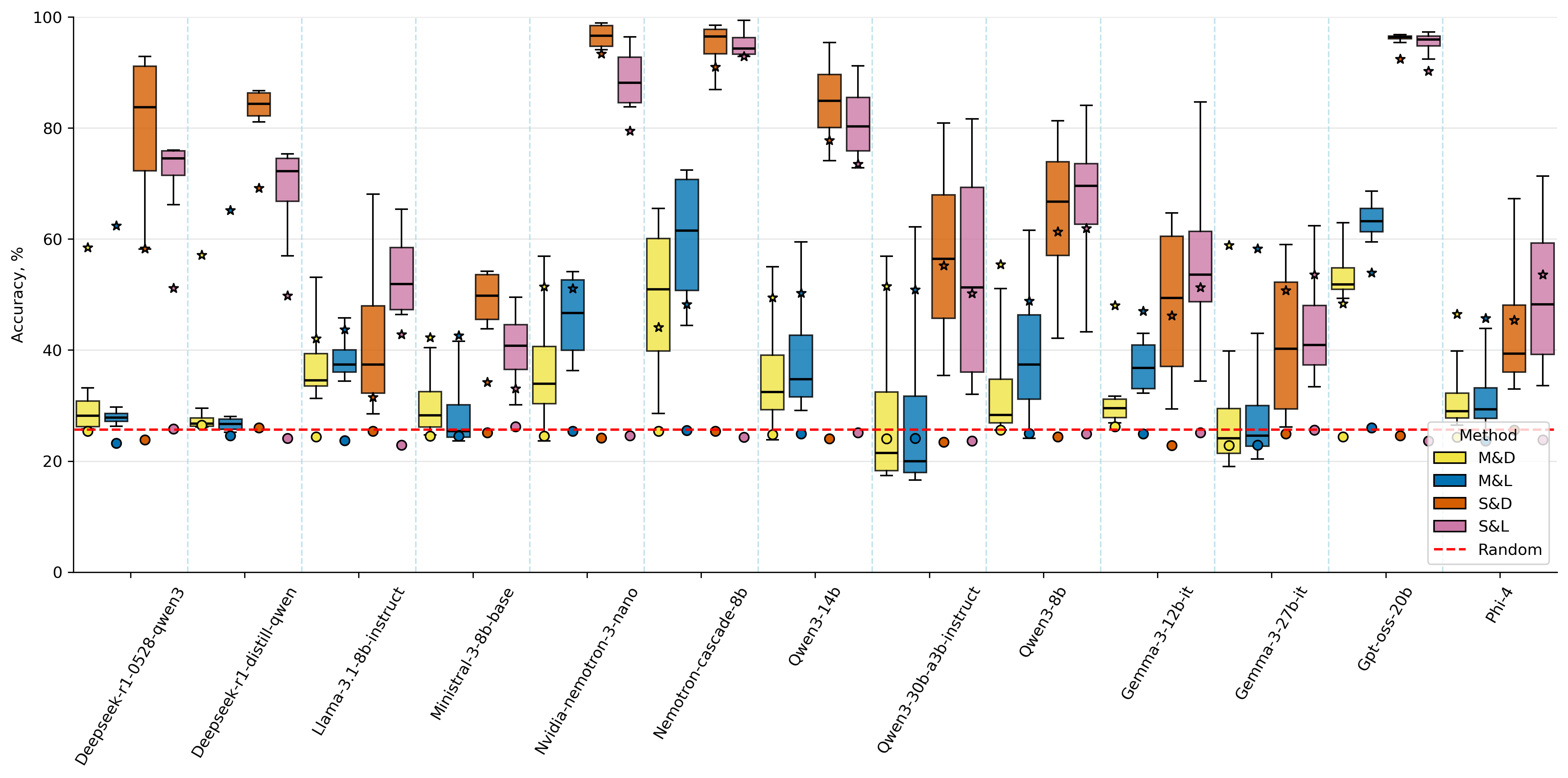}}
    \caption{The performance comparison under all the possible evaluation protocols (S\&L, S\&D, M\&L, and M\&D) on our NonsenseQA dataset with a 5-shot prompt. The boxes illustrate the model performance under all possibilities of ”answer-moving attacks”, where the whiskers indicate the minimum and maximum accuracy for each model. Each dot represents the performance of the original permutations. Additionally, each star symbolizes a SCORE \cite{nalbandyan-etal-2025-score} robustness metric. Each color is consistent with the previous \cref{fig:SL-comparison,fig:SD-comparison,fig:ML-comparison,fig:MD-comparison}.}
    \label{fig:full-comparison}
  \end{center}
\end{figure*}

To enable direct comparison across evaluation protocols on NonsenseQA, we consolidate all previously reported results in Figure~\ref{fig:full-comparison}. For each protocol, we use consistent color coding across figures to aid visual comparison. As demonstrated, the S\&L and M\&D protocols exhibit the highest and lowest levels of bias, respectively. Under S\&L, none of the evaluated models approach chance-level performance, whereas under M\&D, the majority of models exhibit near-random accuracy. This contrast indicates that evaluation artifacts arise jointly from the prediction mode and the option-label presentation. Together, these findings motivate the design of our protocol, which suppresses both sources of bias.

\section{Evaluation details}
\label{appendix:parameters}

\subsection{Computational cost}

In this subsection, we assess the practical overhead of the proposed evaluation protocol. Table~\ref{tab:computation_overhead} compares the empirical computational cost of evaluating CommonsenseQA with the Gemma-3-12b-it model under the S\&L and M\&D protocols. 

Overall, the total evaluation time under M\&D is comparable to that of the standard S\&L protocol. Notably, M\&D exhibits a slightly lower mean generation time than S\&L. One possible explanation is that the required reasoning output under M\&D is simpler, as the model does not encode its final decision into an explicit MCQ label.
 
The additional answer extraction step introduced by M\&D is inexpensive, as the similarity-based extraction with a sentence similarity model on the entire CommonsenseQA dataset takes less than three seconds, accounting for approximately 7\% of the total evaluation time under M\&D. As the generation time under M\&D is marginally shorter than under S\&L, the net increase in overall computation time is limited to approximately 3\%, as reported in the “Ratio” column of Table~\ref{tab:computation_overhead}.

\begin{table}[ht]
\caption{The computational cost of answer generation and extraction for the Gemma-3-12b-it model on CommonsenseQA under the S\&L and M\&D evaluation protocols across all six answer permutations. For each setting, we report the average runtime in seconds, together with the corresponding standard deviation. The results show that the additional extraction step required by M\&D accounts for only a small fraction of the total runtime, indicating that the proposed protocol introduces minimal computational overhead and remains practical for large-scale evaluation.}

\resizebox{\textwidth}{!}{%
\begin{tabular}{lclllcllll}
\cline{2-9}
 &
  \multicolumn{4}{c}{\textbf{S\&L}} &
  \multicolumn{4}{c}{\textbf{M\&D}} &
   \\ \cline{2-9} 
\multirow{-2}{*}{} &
  \multicolumn{1}{c}{\textbf{Generation (s)}} &
  \multicolumn{1}{c}{\textbf{Extraction (s)}} &
  \multicolumn{1}{c}{\textbf{Total (s)}} &
  \multicolumn{1}{c}{\textbf{Extraction cost (\%)}} &
  \multicolumn{1}{c}{\textbf{Generation (s)}} &
  \multicolumn{1}{c}{\textbf{Extraction (s)}} &
  \multicolumn{1}{c}{\textbf{Total (s)}} &
  \multicolumn{1}{c}{\textbf{Extraction cost (\%)}} &
  \multicolumn{1}{l}{\textbf{Ratio}} \\ \hline
\multicolumn{1}{l}{\textbf{Gemma-3-12b-it}} &
  \multicolumn{1}{l}{$36.31 (\pm 1.47)$} &
  \multicolumn{1}{l}{-} &
  \multicolumn{1}{l}{$36.31 (\pm 1.47)$} &
  - &
  \multicolumn{1}{l}{$34.65 (\pm 1.42)$} &
  \multicolumn{1}{l}{$2.70 (\pm0.04)$} &
  \multicolumn{1}{l}{$37.35 (\pm1.43)$} &
  $7\%$ &
  \multicolumn{1}{l}{$1.03$} \\ \hline
\end{tabular}%
}

\label{tab:computation_overhead}
\end{table}

\subsection{Regular expression used}
In this section, we demonstrate the types of regular expressions used to extract the model, showing the usage frequency of each extraction regular expression on MMLU-Pro for all models in Table~\ref{tab:extraction_analysis}. For each regular expression, we extract the last occurrence of that expression within the output if there is a match. All of our regular expressions are based on similar counterparts within the standard evaluation protocol used for MMLU-Pro \cite{wang2024mmlu}. Below, we show the exact regular expressions used in the order of testing:
\begin{itemize}
    \item \textbf{Expression \#1}: \verb/answer is (?!.*answer is ).+/

    The first regular expression mirrors the one used in the MMLU-Pro evaluation code (\verb/answer is \(?([A-J])\)?/), which searches for an explicitly instructed answer match within the model output. In practice, the majority of extractions are resolved at this stage.
    \item \textbf{Expression \#2}: \verb/.*[aA]nswer:\s*(?!.*[aA]nswer:\s*).+/ 
    
    The second regular expression is similarly adapted from MMLU-Pro (\verb/.*[aA]nswer:\s*([A-J])/) and permits minor deviations in the answer prefix, increasing robustness to formatting variation.
    \item \textbf{Expression \#3}: A regular expression verbosely searching each answer within the model's output as is.

    Following the MMLU-Pro approach, which allows random A–J characters to be selected when earlier extraction attempts fail (\verb/\b[A-J]\b(?!.*\b[A-J]\b)/), we instead allow the full, verbose answer to be captured if the first two extraction steps are unsuccessful.
    \item \textbf{Expression \#4}: \verb/([^.!?]+[.!?]*$)/

    As a final fallback, we extract the last sentence of the model’s output. While this step serves a role analogous to MMLU-Pro’s random-choice fallback, we hypothesize that using the last generated sentence preserves semantically relevant information that can more informatively guide the similarity-based matching process.
\end{itemize}

We observe that the majority of models satisfy Extraction Rule \#1 in over 95\% of generated outputs. The two notable exceptions are the two DeepSeek models, which achieve extraction success rates of approximately 70–80\%. Manual inspection reveals that these failures are primarily due to formatting issues, such as the absence of whitespace in the generated answers, which complicates reliable extraction.

If the model does not produce an answer, we choose a random answer. However, this has only occurred once in all of the permutations over all of the models on MMLU-Pro, as demonstrated in Table~\ref{tab:extraction_analysis}. 
\begin{table}[ht]

\caption{Extraction frequency of each proposed regular expression on MMLU-Pro \cite{wang2024mmlu} across models and all permutations evaluated. We observe that, for most models, approximately 98\% of outputs adhere to the intended answer format and are successfully extracted by the primary matching rules, indicating that models generally follow the instructed output format reliably.}

\resizebox{\textwidth}{!}{%
\begin{tabular}{lccccc}
\cline{2-6}
 &
  \multicolumn{5}{c}{\textbf{Frequency, \%}} \\ \hline

\multicolumn{1}{c}{\textbf{Model Name}} &
  \multicolumn{1}{l}{\textbf{Extraction \#1}} &
  \multicolumn{1}{l}{\textbf{Extraction \#2}} &
  \multicolumn{1}{l}{\textbf{Extraction \#3}} &
  \multicolumn{1}{l}{\textbf{Extraction \#4}} &
  \multicolumn{1}{l}{\textbf{Failed}} \\ \hline
\multicolumn{1}{l}{DeepSeek-R1-0528-Qwen3-8B} &
  \multicolumn{1}{c}{72.75} &
  \multicolumn{1}{c}{1.92} &
  \multicolumn{1}{c}{8.24} &
  \multicolumn{1}{c}{17.09} &
  0.00 \\ \hline
\multicolumn{1}{l}{DeepSeek-R1-Distill-Qwen-7B} &
  \multicolumn{1}{c}{83.75} &
  \multicolumn{1}{c}{2.73} &
  \multicolumn{1}{c}{3.74} &
  \multicolumn{1}{c}{9.78} &
  0.00 \\ \hline
\multicolumn{1}{l}{Llama-3.1-8B-Instruct} &
  \multicolumn{1}{c}{95.81} &
  \multicolumn{1}{c}{0.00} &
  \multicolumn{1}{c}{0.88} &
  \multicolumn{1}{c}{3.30} &
  0.00 \\ \hline
\multicolumn{1}{l}{Ministral-3-8B-Base-2512} &
  \multicolumn{1}{c}{97.34} &
  \multicolumn{1}{c}{0.04} &
  \multicolumn{1}{c}{1.26} &
  \multicolumn{1}{c}{1.36} &
  0.00 \\ \hline
\multicolumn{1}{l}{NVIDIA-Nemotron-3-Nano-30B-A3B-BF16} &
  \multicolumn{1}{c}{95.99} &
  \multicolumn{1}{c}{0.70} &
  \multicolumn{1}{c}{2.08} &
  \multicolumn{1}{c}{1.23} &
  0.00 \\ \hline
\multicolumn{1}{l}{Nemotron-Cascade-8B} &
  \multicolumn{1}{c}{99.75} &
  \multicolumn{1}{c}{0.00} &
  \multicolumn{1}{c}{0.12} &
  \multicolumn{1}{c}{0.12} &
  0.00 \\ \hline
\multicolumn{1}{l}{Qwen3-14B} &
  \multicolumn{1}{c}{98.57} &
  \multicolumn{1}{c}{0.07} &
  \multicolumn{1}{c}{0.51} &
  \multicolumn{1}{c}{0.86} &
  0.00 \\ \hline
\multicolumn{1}{l}{Qwen3-30B-A3B-Instruct-2507} &
  \multicolumn{1}{c}{95.62} &
  \multicolumn{1}{c}{0.01} &
  \multicolumn{1}{c}{2.59} &
  \multicolumn{1}{c}{1.78} &
  0.00 \\ \hline
\multicolumn{1}{l}{Qwen3-8B} &
  \multicolumn{1}{c}{98.56} &
  \multicolumn{1}{c}{0.00} &
  \multicolumn{1}{c}{0.67} &
  \multicolumn{1}{c}{0.76} &
  0.00 \\ \hline
\multicolumn{1}{l}{gemma-3-12b-it} &
  \multicolumn{1}{c}{99.62} &
  \multicolumn{1}{c}{0.01} &
  \multicolumn{1}{c}{0.12} &
  \multicolumn{1}{c}{0.25} &
  0.00 \\ \hline
\multicolumn{1}{l}{gemma-3-27b-it} &
  \multicolumn{1}{c}{98.69} &
  \multicolumn{1}{c}{0.00} &
  \multicolumn{1}{c}{0.45} &
  \multicolumn{1}{c}{0.86} &
  0.01 \\ \hline
\multicolumn{1}{l}{gpt-oss-20b} &
  \multicolumn{1}{c}{91.82} &
  \multicolumn{1}{c}{3.42} &
  \multicolumn{1}{c}{1.12} &
  \multicolumn{1}{c}{3.64} &
  0.00 \\ \hline
\multicolumn{1}{l}{phi-4} &
  \multicolumn{1}{c}{99.12} &
  \multicolumn{1}{c}{0.11} &
  \multicolumn{1}{c}{0.31} &
  \multicolumn{1}{c}{0.46} &
  0.00 \\ \hline
\end{tabular}%
}

\label{tab:extraction_analysis}
\end{table}

\subsection{Default parameters and prompt modification}
\label{appendix:prompt}

When generating results, we use models' preferred generation parameters provided on Huggingface, as it has been proven that they do not yield significant differences in MCQ answering. However, if no generation configuration is provided, we use the set of default parameters shown in Table~\ref{tab:parameters} below.

\begin{table}[ht]
\caption{Default set of parameters used when the generation configuration is not provided on Huggingface.}

\centering
\begin{tabular}{ccccc}
\toprule
\textbf{Parameter}                     & \textbf{Temperature} & \textbf{Top K} & \textbf{Top P} & \textbf{Min P} \\
\midrule
\textbf{Value} & 0.6                  & 20             & 0.95           & 0              \\ 
\bottomrule
\end{tabular}%

\label{tab:parameters}
\end{table}

\begin{figure*}[ht]
  \begin{center}
    \centerline{\includegraphics[width=\textwidth]{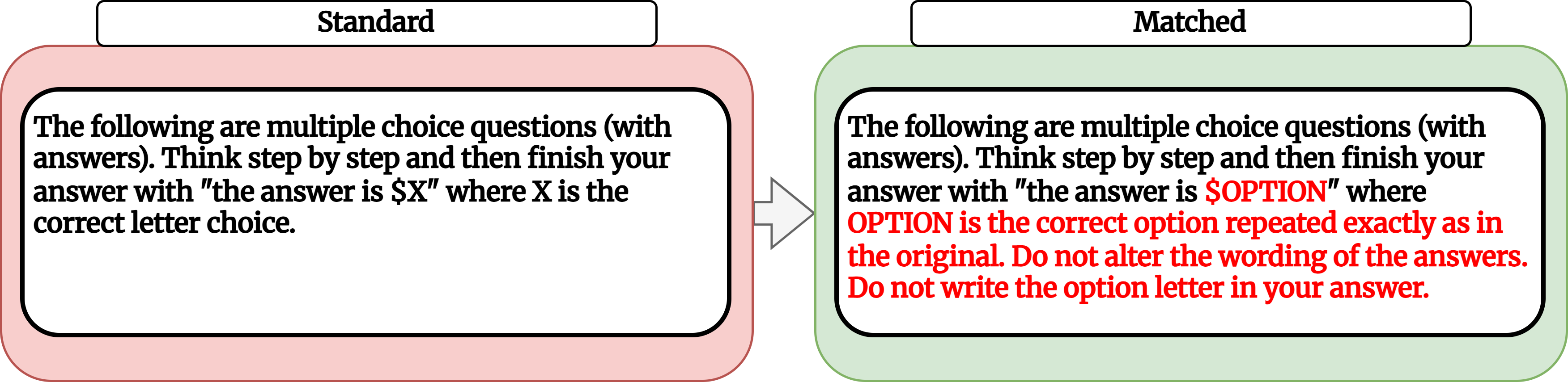}}
    \caption{The proposed prompt modification with all changes in red. We find that by changing "\$X" to "\$OPTION" and adding the additional constraint forces the model to switch from letter prediction to full-text option prediction without mentioning any letters in their answers.}
    \label{fig:prompt_update}
  \end{center}
\end{figure*}

To enforce a full-text model generation and adherence to the option provided, we modify the generation instruction slightly, as shown in Figure~\ref{fig:prompt_update}. We find that by substituting “\$X” with “\$OPTION”, the model is less likely to answer using a letter, while the additional sentences improve the model output and further remove the letters used.

\subsection{Prompt example - Matched}
\begin{figure*}[ht]
  \begin{center}
    \centerline{\includegraphics[height=0.9\textheight]{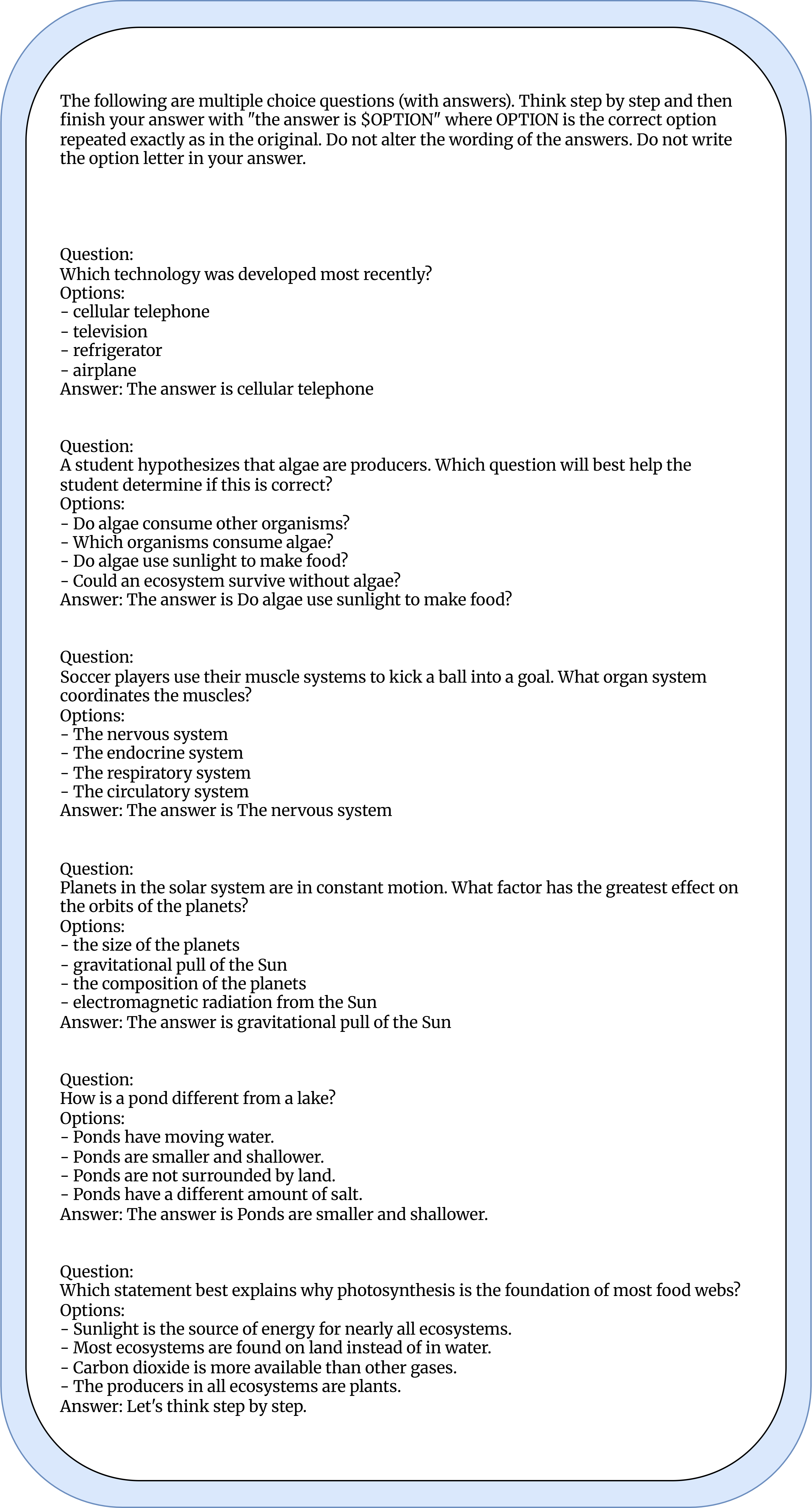}}
    \caption{An example few-shot prompt with a question from the ARC-Easy \cite{clark2018arc} dataset within our evaluation protocol. As with the MMLU-Pro \cite{wang2024mmlu} prompt, we add "Let's think step by step." to induce Chain-of-thought reasoning.}
    \label{fig:example_prompt}
  \end{center}
\end{figure*}

We provide an example few-shot question prompt under M\&D protocol used for the ARC-Easy \cite{clark2018arc} benchmark evaluation in  Figure~\ref{fig:example_prompt}.

\section{Code and benchmark release} \label{appendix:code_release}
For the code release, we provide the evaluation code used to run CommonsenseQA \cite{talmor-etal-2019-commonsenseqa} under both the S\&L and M\&D protocols, demonstrating how easily existing benchmarks can be adapted to our evaluation framework.

For NonsenseQA, we do not release a fixed benchmark instance. Instead, we provide lightweight code for generating the dataset. Releasing a static version of NonsenseQA would undermine its purpose as a diagnostic tool, as models could be explicitly trained on it to conceal underlying biases rather than expose them.

\section{Limitations} \label{appendix:limitations}
While we propose a bias-reduced evaluation protocol, our work has several limitations.

\begin{itemize}
    \item \textbf{Position bias}: Firstly, we do not attempt to mitigate position bias. Our moving-answer attack explicitly demonstrates the existence and impact of positional sensitivity, but the proposed protocol is designed to expose this bias, rather than correct it. In most benchmarks, position bias has a limited effect on model performance, as observed in the main body of the paper. However, in MMLU-Pro \cite{wang2024mmlu}, the original answer position appears to influence evaluation significantly. This effect is clearly illustrated in Figure~\ref{fig:MMLU-comparison}. Moreover, this could be the bias that prevents some models in NonsenseQA experiments from achieving 25\% mean accuracy, as demonstrated in Figure~\ref{fig:NonsenseQA-comparison}. Nevertheless, our evaluation protocol is orthogonal to some of the position bias reduction solutions and can be combined with them. Developing methods to mitigate position bias without additional training or access to model logits, consistent with our work, remains an open direction for future research.

    \item \textbf{Generation vs. selection}: Secondly, we do not directly quantify the gap between answer generation and answer selection. This would more explicitly characterize the difference between MCQ-induced bias and natural LLM reasoning. A fair comparison would require transforming existing benchmarks to support free-form answer generation while preserving their original semantics, including the removal or redesign of questions that inherently rely on multiple-choice structure (e.g., “Which of the following is the smallest?”). Constructing such datasets is beyond the scope of this paper, and we encourage future work to develop benchmarks that jointly support MCQ-based and generative evaluation.

    \item \textbf{English bias}: Finally, although we use the INCLUDE \cite{romanouinclude} benchmark’s questions and answers in their original languages, we retain English-only instructions rather than using the original language-specific prompts. To rigorously evaluate the English language-induced biases, it would require designing additional language-specific answer extraction rules for the multilingual settings. While feasible, this would require linguistic expertise. We therefore leave a systematic study of instruction language and answer extraction effects in multilingual evaluation as an avenue for future work.
\end{itemize}

\section{Detailed results}
\label{appendix:detailed_results}
In \cref{tab:NonsenseQA,tab:CSQA,tab:ARC,tab:GPQA,tab:MMLU-Pro,tab:INCLUDE}, we report detailed per-benchmark results under both the S\&L and M\&D evaluation protocols. Table~\ref{tab:NonsenseQA} presents results on the NonsenseQA benchmark, while the remaining tables correspond to real-world benchmarks. For each protocol and model, we report the mean accuracy under attack ($ACC_{Mean}$), the original permutation accuracy ($ACC_{ORG}$), the absolute difference between original and attacked performance ($|\Delta ACC|$), the variance across all permutations ($\sigma^2$), the minimum and maximum accuracies ($ACC_{min}$ and $ACC_{max}$, respecitvely), the resulting accuracy range ($\Delta_{max-min}$), and the SCORE metric (\textbf{SCORE}).
In addition, for each model, we compute
\begin{itemize}
    \item the variance ratio, defined as: 

    \begin{equation}
        \sigma^2_{R} = \frac{\sigma^2_{M\&D}}{\sigma^2_{S\&L}},
    \end{equation}
    \item the difference in the SCORE metric, defined as: 

    \begin{equation}
        \Delta_{SCORE} = SCORE_{M\&D} - SCORE_{S\&L}.
    \end{equation}
\end{itemize}
A variance ratio below one indicates reduced variance under the M\&D protocol, while a negative SCORE difference indicates improved SCORE relative to S\&L. The improvements in the significant results are highlighted in bold to facilitate visual comparison across models and protocols. We use the SCORE difference as a diagnostic tool, showing which models achieve higher reasoning consistency under M\&D protocol (negative score) and which models achieve higher consistency under S\&L protocol (positive score). The improvements in the significant results are highlighted in bold to facilitate visual comparison across models and protocols.

\subsection{NonsenseQA - detailed results}
For the NonsenseQA results reported in Table~\ref{tab:NonsenseQA}, we focus on two key statistics - the mean accuracy under attack and the SCORE difference. Given the synthetic nature of the dataset, we emphasize results in which the mean accuracy is closer to the chance level of 25\%, as well as with models achieving negative SCORE differences, which indicate models that behave more randomly while exhibiting higher consistency under the M\&D protocol.
\begin{table}[ht]
\caption{Detailed results on \textbf{NonsenseQA}.}
\resizebox{\textwidth}{!}{%
\begin{tabular}{lcccccccccccccccccc}
\cline{2-17}
 &
  \multicolumn{8}{c}{\textbf{S\&L}} &
  \multicolumn{8}{c}{\textbf{M\&D}} &
  \multicolumn{1}{l}{} &
  \multicolumn{1}{l}{} \\ \hline
 
\multicolumn{1}{c}{\textbf{Model name}} &
  \multicolumn{1}{c}{\textbf{$ACC_{Mean}$}} &
  \multicolumn{1}{c}{\textbf{$ACC_{ORG}$}} &
  \multicolumn{1}{c}{\textbf{$|\Delta ACC|$}} &
  \multicolumn{1}{c}{\textbf{$\sigma^2$}} &
  \multicolumn{1}{c}{\textbf{$ACC_{min}$}} &
  \multicolumn{1}{c}{\textbf{$ACC_{max}$}} &
  \multicolumn{1}{c}{\textbf{$\Delta_{max-min}$}} &
  \textbf{SCORE} &
  \multicolumn{1}{c}{\textbf{$ACC_{Mean}$}} &
  \multicolumn{1}{c}{\textbf{$ACC_{ORG}$}} &
  \multicolumn{1}{c}{\textbf{$|\Delta ACC|$}} &
  \multicolumn{1}{c}{\textbf{$\sigma^2$}} &
  \multicolumn{1}{c}{\textbf{$ACC_{min}$}} &
  \multicolumn{1}{c}{\textbf{$ACC_{max}$}} &
  \multicolumn{1}{c}{\textbf{$\Delta_{max-min}$}} &
  \textbf{SCORE} &
  \multicolumn{1}{c}{\textbf{$\sigma^2_{Ratio}$}} &
  \multicolumn{1}{c}{\textbf{$\Delta_{SCORE}$}} \\ \hline
\multicolumn{1}{l}{Deepseek-r1-0528-qwen3} &
  72.80 &
  76.00 &
  4.27 &
  15.74 &
  66.20 &
  76.00 &
  9.80 &
  0.39 &
  \textbf{28.82} &
  30.00 &
  1.57 &
  9.01 &
  25.80 &
  33.20 &
  7.40 &
  0.59 &
  0.57 &
  \multicolumn{1}{c}{\textbf{-0.195}} \\ \hline
\multicolumn{1}{l}{Deepseek-r1-distill-qwen} &
  69.17 &
  75.30 &
  8.17 &
  53.22 &
  57.00 &
  75.30 &
  18.30 &
  0.42 &
  \textbf{27.28} &
  27.20 &
  0.10 &
  1.82 &
  26.10 &
  29.50 &
  3.40 &
  0.58 &
  0.03 &
  \multicolumn{1}{c}{\textbf{-0.154}} \\ \hline
\multicolumn{1}{l}{Llama-3.1-8b-instruct} &
  53.88 &
  56.10 &
  2.97 &
  58.26 &
  46.40 &
  65.40 &
  19.00 &
  0.38 &
  \textbf{38.38} &
  53.10 &
  19.63 &
  74.07 &
  31.30 &
  53.10 &
  21.80 &
  0.43 &
  1.27 &
  \multicolumn{1}{c}{\textbf{-0.052}} \\ \hline
\multicolumn{1}{l}{Ministral-3-8b-base} &
  40.27 &
  42.90 &
  3.50 &
  49.58 &
  30.10 &
  49.50 &
  19.40 &
  0.34 &
  \textbf{30.40} &
  40.40 &
  13.33 &
  36.80 &
  24.70 &
  40.40 &
  15.70 &
  0.44 &
  0.74 &
  \multicolumn{1}{c}{\textbf{-0.106}} \\ \hline
\multicolumn{1}{l}{Nvidia-nemotron-3-nano} &
  89.12 &
  96.40 &
  9.70 &
  26.41 &
  83.80 &
  96.40 &
  12.60 &
  0.58 &
  \textbf{37.08} &
  56.90 &
  26.43 &
  149.54 &
  23.60 &
  56.90 &
  33.30 &
  0.53 &
  5.66 &
  \multicolumn{1}{c}{0.050} \\ \hline
\multicolumn{1}{l}{Nemotron-cascade-8b} &
  95.22 &
  99.40 &
  5.57 &
  6.54 &
  92.90 &
  99.40 &
  6.50 &
  0.67 &
  \textbf{49.00} &
  58.30 &
  12.40 &
  201.01 &
  28.60 &
  65.50 &
  36.90 &
  0.44 &
  30.73 &
  \multicolumn{1}{c}{0.226} \\ \hline
\multicolumn{1}{l}{Qwen3-14b} &
  81.12 &
  91.20 &
  13.43 &
  48.70 &
  72.80 &
  91.20 &
  18.40 &
  0.57 &
  \textbf{35.93} &
  55.00 &
  25.43 &
  134.67 &
  23.80 &
  55.00 &
  31.20 &
  0.50 &
  2.77 &
  \multicolumn{1}{c}{0.070} \\ \hline
\multicolumn{1}{l}{Qwen3-30b-a3b-instruct} &
  54.05 &
  81.60 &
  36.73 &
  411.69 &
  32.00 &
  81.60 &
  49.60 &
  0.47 &
  \textbf{29.30} &
  56.90 &
  36.80 &
  260.71 &
  17.40 &
  56.90 &
  39.50 &
  0.53 &
  0.63 &
  \multicolumn{1}{c}{\textbf{-0.059}} \\ \hline
\multicolumn{1}{l}{Qwen3-8b} &
  66.65 &
  84.10 &
  23.27 &
  216.91 &
  43.30 &
  84.10 &
  40.80 &
  0.54 &
  \textbf{33.33} &
  51.10 &
  23.70 &
  107.03 &
  25.60 &
  51.10 &
  25.50 &
  0.57 &
  0.49 &
  \multicolumn{1}{c}{\textbf{-0.034}} \\ \hline
\multicolumn{1}{l}{Gemma-3-12b-it} &
  56.55 &
  34.40 &
  29.53 &
  325.26 &
  34.40 &
  84.70 &
  50.30 &
  0.49 &
  \textbf{29.43} &
  26.90 &
  3.37 &
  3.95 &
  26.90 &
  31.70 &
  4.80 &
  0.51 &
  0.01 &
  \multicolumn{1}{c}{\textbf{-0.025}} \\ \hline
\multicolumn{1}{l}{Gemma-3-27b-it} &
  44.40 &
  43.20 &
  1.60 &
  120.02 &
  33.40 &
  62.40 &
  29.00 &
  0.54 &
  \textbf{26.75} &
  26.00 &
  1.00 &
  62.91 &
  19.00 &
  39.80 &
  20.80 &
  0.62 &
  0.52 &
  \multicolumn{1}{c}{\textbf{-0.075}} \\ \hline
\multicolumn{1}{l}{Gpt-oss-20b} &
  95.40 &
  96.30 &
  1.20 &
  3.36 &
  92.40 &
  97.30 &
  4.90 &
  0.63 &
  \textbf{53.95} &
  52.10 &
  2.47 &
  27.79 &
  49.30 &
  62.90 &
  13.60 &
  0.46 &
  8.26 &
  \multicolumn{1}{c}{0.176} \\ \hline
\multicolumn{1}{l}{Phi-4} &
  50.33 &
  71.30 &
  27.97 &
  207.38 &
  33.60 &
  71.30 &
  37.70 &
  0.51 &
  \textbf{31.05} &
  39.80 &
  11.67 &
  26.80 &
  26.50 &
  39.80 &
  13.30 &
  0.48 &
  0.13 &
  \multicolumn{1}{c}{0.029} \\ \hline
\end{tabular}%
}

\label{tab:NonsenseQA}
\end{table}

\subsection{Real benchmark - detailed results}
\begin{figure*}[!ht]
  \begin{center}
    \centerline{\includegraphics[width=\textwidth]{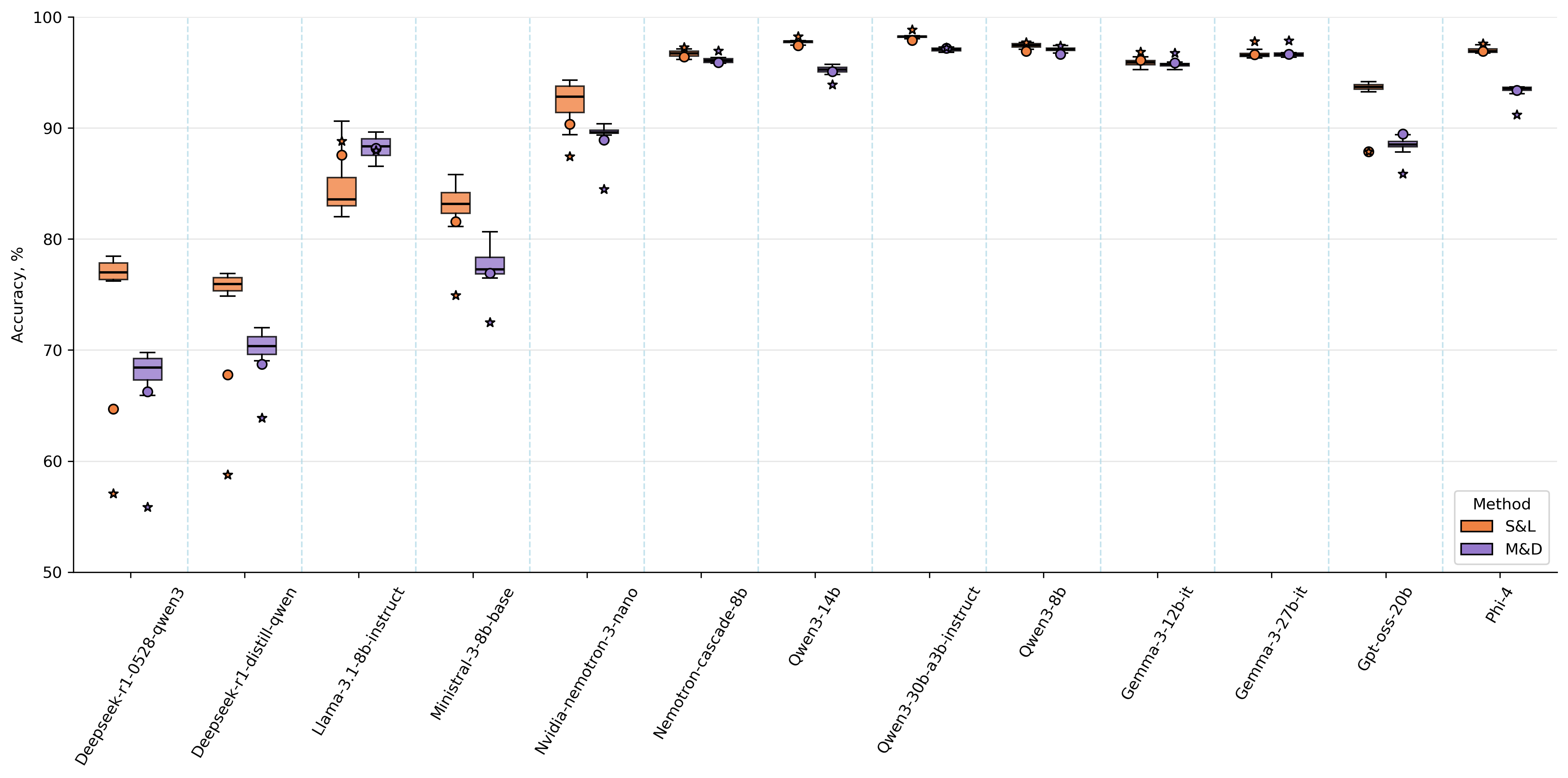}}
    \caption{Comparison of the matched prediction with dashes as labels (M\&D; our method) with standard letter prediction with letters as labels (S\&L) on ARC \cite{clark2018arc} with a 5-shot prompt. The boxes illustrate the model performance under all possibilities of ”answer-moving attacks”, where the whiskers indicate the minimum and maximum accuracy for each model. Each dot represents the performance of the original permutations. Additionally, each star symbolizes a SCORE \cite{nalbandyan-etal-2025-score} robustness metric.}
    \label{fig:ARC-comparison}
  \end{center}
\end{figure*}

\begin{figure*}[!ht]
  \begin{center}
    \centerline{\includegraphics[width=\textwidth]{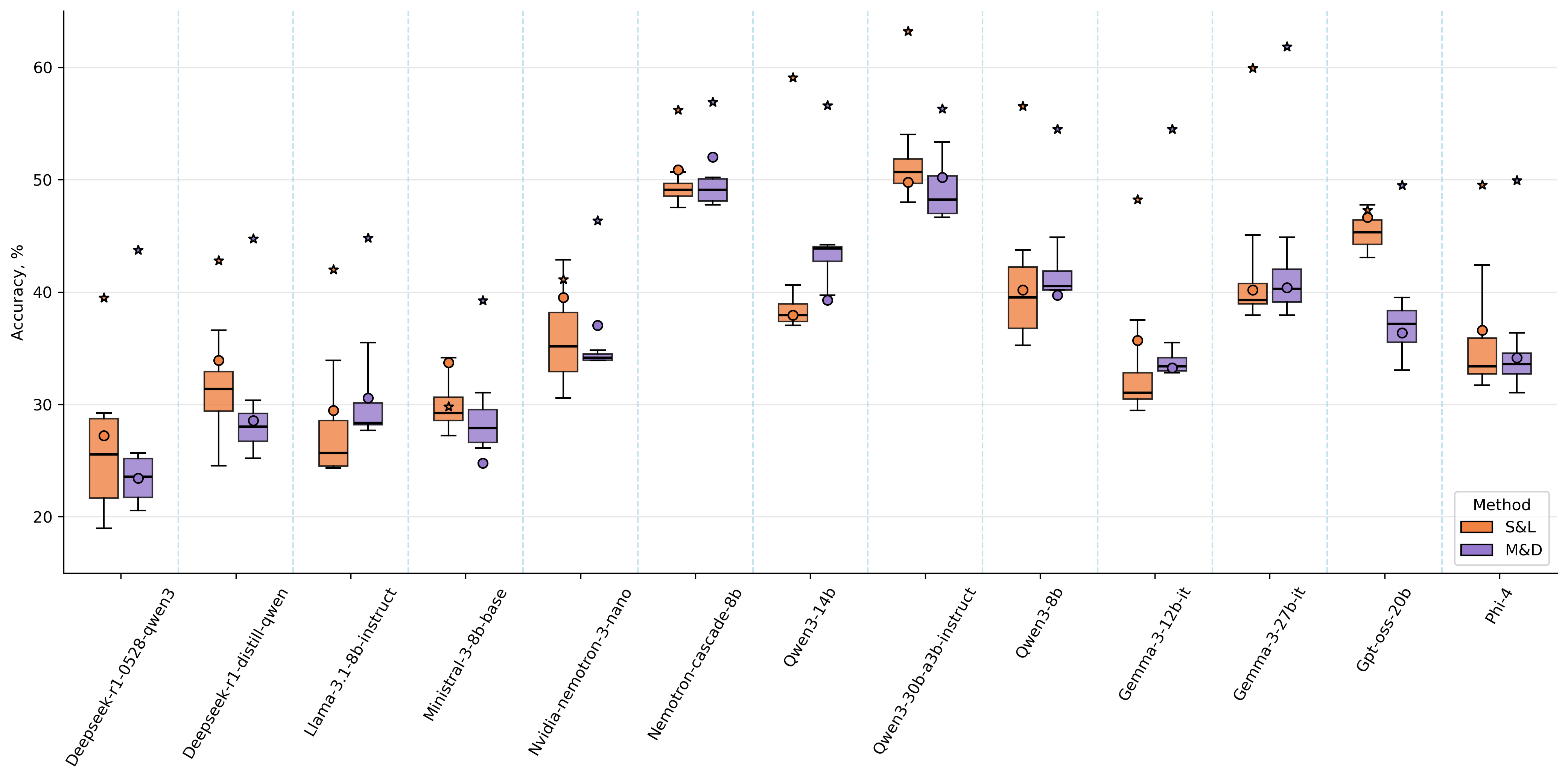}}
    \caption{Comparison of the matched prediction with dashes as labels (M\&D; our method) with standard letter prediction with letters as labels (S\&L) on GPQA \cite{rein2024gpqa} with a 5-shot prompt. The boxes illustrate the model performance under all possibilities of ”answer-moving attacks”, where the whiskers indicate the minimum and maximum accuracy for each model. Each dot represents the performance of the original permutations. Additionally, each star symbolizes a SCORE \cite{nalbandyan-etal-2025-score} robustness metric.}
    \label{fig:GPQA-comparison}
  \end{center}
\end{figure*}

\begin{figure*}[!ht]
  \begin{center}
    \centerline{\includegraphics[width=\textwidth]{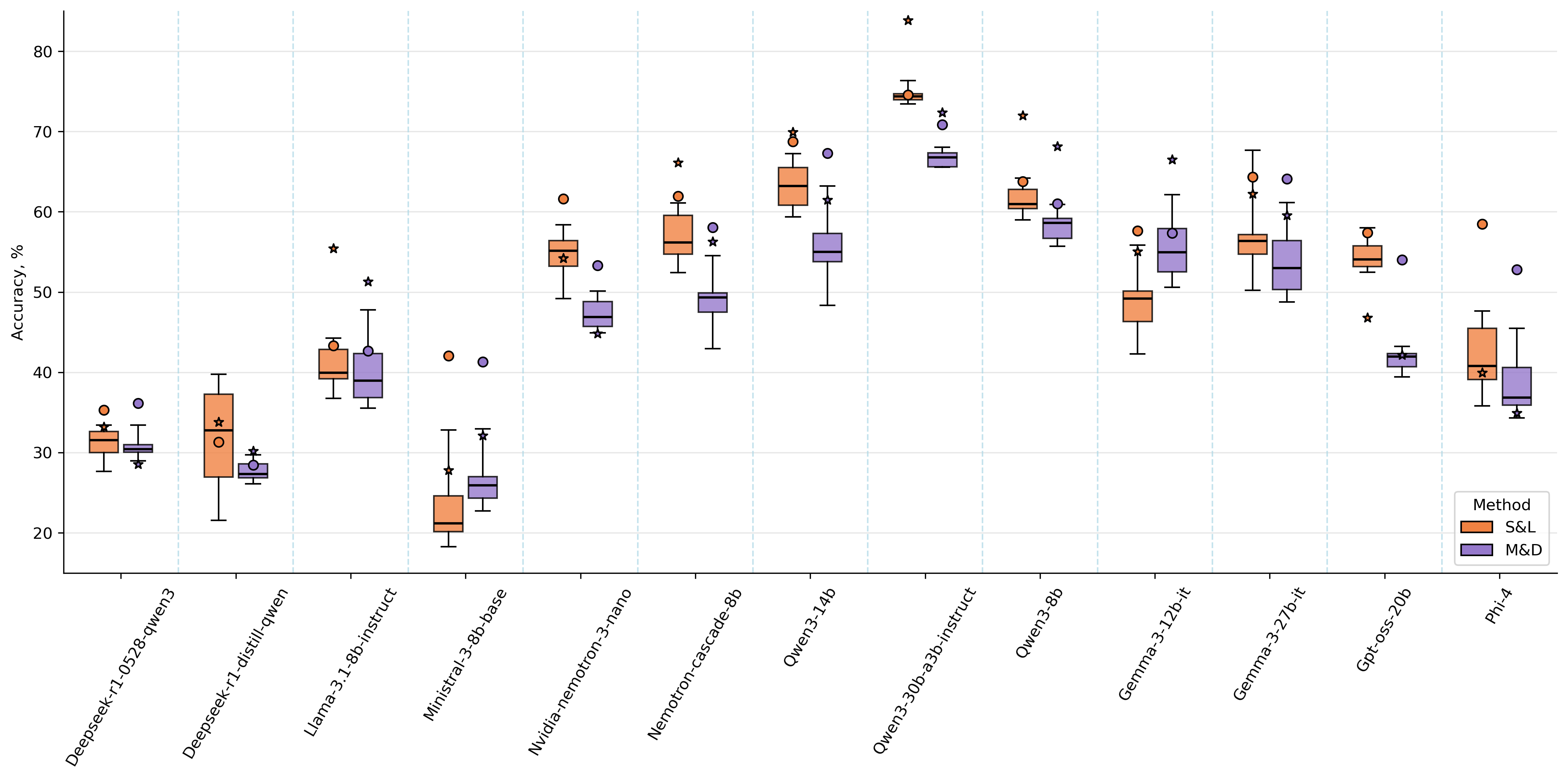}}
    \caption{Comparison of the matched prediction with dashes as labels (M\&D; our method) with standard letter prediction with letters as labels (S\&L) on MMLU-Pro \cite{wang2024mmlu} with a 5-shot prompt. The boxes illustrate the model performance under all possibilities of ”answer-moving attacks”, where the whiskers indicate the minimum and maximum accuracy for each model. Each dot represents the performance of the original permutations. Additionally, each star symbolizes a SCORE \cite{nalbandyan-etal-2025-score} robustness metric.}
    \label{fig:MMLU-comparison}
  \end{center}
\end{figure*}

\begin{figure*}[!ht]
  \begin{center}
    \centerline{\includegraphics[width=\textwidth]{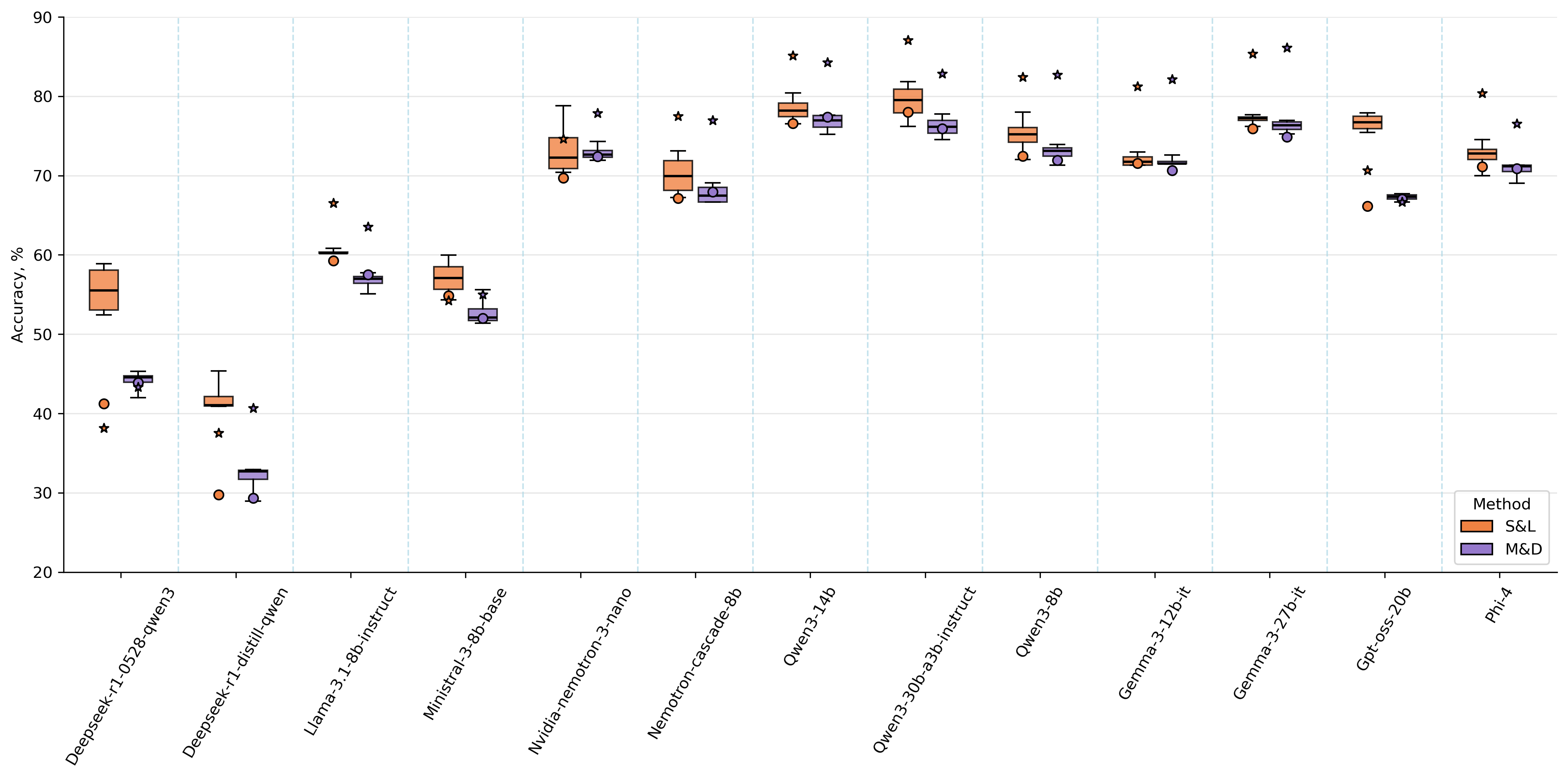}}
    \caption{Comparison of the matched prediction with dashes as labels (M\&D; our method) with standard letter prediction with letters as labels (S\&L) on the subset of common languages on the INCLUDE benchmark \cite{romanouinclude} with a 5-shot prompt. The boxes illustrate the model performance under all possibilities of ”answer-moving attacks”, where the whiskers indicate the minimum and maximum accuracy for each model. Each dot represents the performance of the original permutations. Additionally, each star symbolizes a SCORE \cite{nalbandyan-etal-2025-score} robustness metric.}
    \label{fig:INCLUDE-comparison}
  \end{center}
\end{figure*}

For all the real-life results reported in \cref{tab:CSQA,tab:ARC,tab:GPQA,tab:MMLU-Pro,tab:INCLUDE} and \cref{fig:CommonsenseQA-comparison,fig:ARC-comparison,fig:GPQA-comparison,fig:MMLU-comparison,fig:INCLUDE-comparison}, we highlight the absolute difference between original and attacked performance closer to zero, a lower variance ratio score, and negative SCORE values. These statistics demonstrate which models achieve more stable performance results while consistently choosing similar answers. 

\begin{table}[ht]

\caption{Detailed results on \textbf{CommonsenseQA} \cite{talmor-etal-2019-commonsenseqa}.}

\resizebox{\textwidth}{!}{%
\begin{tabular}{lcccccccccccccccccc}
\cline{2-17}
 &
  \multicolumn{8}{c}{\textbf{S\&L}} &
  \multicolumn{8}{c}{\textbf{M\&D}} &
  \multicolumn{1}{l}{} &
  \multicolumn{1}{l}{} \\ \hline
 
\multicolumn{1}{c}{\textbf{Model name}} &
  \multicolumn{1}{c}{\textbf{$ACC_{Mean}$}} &
  \multicolumn{1}{c}{\textbf{$ACC_{ORG}$}} &
  \multicolumn{1}{c}{\textbf{$|\Delta ACC|$}} &
  \multicolumn{1}{c}{\textbf{$\sigma^2$}} &
  \multicolumn{1}{c}{\textbf{$ACC_{min}$}} &
  \multicolumn{1}{c}{\textbf{$ACC_{max}$}} &
  \multicolumn{1}{c}{\textbf{$\Delta_{max-min}$}} &
  \textbf{SCORE} &
  \multicolumn{1}{c}{\textbf{$ACC_{Mean}$}} &
  \multicolumn{1}{c}{\textbf{$ACC_{ORG}$}} &
  \multicolumn{1}{c}{\textbf{$|\Delta ACC|$}} &
  \multicolumn{1}{c}{\textbf{$\sigma^2$}} &
  \multicolumn{1}{c}{\textbf{$ACC_{min}$}} &
  \multicolumn{1}{c}{\textbf{$ACC_{max}$}} &
  \multicolumn{1}{c}{\textbf{$\Delta_{max-min}$}} &
  \textbf{SCORE} &
  \multicolumn{1}{c}{\textbf{$\sigma^2_{Ratio}$}} &
  \multicolumn{1}{c}{\textbf{$\Delta_{SCORE}$}} \\ \hline
\multicolumn{1}{l}{Deepseek-r1-0528-qwen3} &
  64.18 &
  42.92 &
  25.51 &
  94.43 &
  42.92 &
  70.84 &
  27.92 &
  0.46 &
  52.32 &
  52.66 &
  \textbf{0.41} &
  0.48 &
  51.19 &
  53.40 &
  2.21 &
  0.46 &
  \textbf{0.01} &
  \multicolumn{1}{c}{0.002} \\ \hline
\multicolumn{1}{l}{Deepseek-r1-distill-qwen} &
  61.08 &
  53.15 &
  9.52 &
  33.58 &
  53.15 &
  68.88 &
  15.73 &
  0.51 &
  52.50 &
  50.29 &
  \textbf{2.65} &
  1.28 &
  50.29 &
  53.97 &
  3.68 &
  0.54 &
  \textbf{0.04} &
  \multicolumn{1}{c}{\textbf{-0.025}} \\ \hline
\multicolumn{1}{l}{Llama-3.1-8b-instruct} &
  76.93 &
  75.59 &
  1.61 &
  3.01 &
  74.77 &
  80.02 &
  5.25 &
  0.78 &
  75.54 &
  75.51 &
  \textbf{0.03} &
  5.07 &
  71.99 &
  79.69 &
  7.70 &
  0.78 &
  1.69 &
  \multicolumn{1}{c}{\textbf{-0.001}} \\ \hline
\multicolumn{1}{l}{Ministral-3-8b-base} &
  67.53 &
  64.86 &
  3.20 &
  21.28 &
  63.47 &
  77.07 &
  13.60 &
  0.61 &
  64.30 &
  62.90 &
  \textbf{1.69} &
  7.36 &
  59.95 &
  68.30 &
  8.35 &
  0.63 &
  \textbf{0.35} &
  \multicolumn{1}{c}{\textbf{-0.018}} \\ \hline
\multicolumn{1}{l}{Nvidia-nemotron-3-nano} &
  84.77 &
  63.88 &
  25.06 &
  98.58 &
  63.88 &
  94.35 &
  30.47 &
  0.77 &
  73.68 &
  72.65 &
  \textbf{1.24} &
  0.79 &
  72.65 &
  75.18 &
  2.53 &
  0.75 &
  \textbf{0.01} &
  \multicolumn{1}{c}{0.015} \\ \hline
\multicolumn{1}{l}{Nemotron-cascade-8b} &
  86.21 &
  83.95 &
  2.72 &
  2.89 &
  83.95 &
  88.86 &
  4.91 &
  0.90 &
  84.47 &
  83.54 &
  \textbf{1.11} &
  0.30 &
  83.54 &
  85.26 &
  1.72 &
  0.90 &
  \textbf{0.10} &
  \multicolumn{1}{c}{0.005} \\ \hline
\multicolumn{1}{l}{Qwen3-14b} &
  86.27 &
  84.19 &
  2.50 &
  1.43 &
  84.19 &
  88.29 &
  4.10 &
  0.91 &
  84.61 &
  84.44 &
  \textbf{0.21} &
  0.01 &
  84.44 &
  84.77 &
  0.33 &
  0.91 &
  \textbf{0.01} &
  \multicolumn{1}{c}{\textbf{-0.003}} \\ \hline
\multicolumn{1}{l}{Qwen3-30b-a3b-instruct} &
  87.00 &
  86.24 &
  0.92 &
  1.61 &
  85.50 &
  88.53 &
  3.03 &
  0.93 &
  85.66 &
  85.26 &
  \textbf{0.47} &
  1.07 &
  84.28 &
  86.98 &
  2.70 &
  0.91 &
  \textbf{0.66} &
  \multicolumn{1}{c}{0.018} \\ \hline
\multicolumn{1}{l}{Qwen3-8b} &
  85.03 &
  83.29 &
  2.08 &
  1.77 &
  83.29 &
  86.98 &
  3.69 &
  0.88 &
  83.36 &
  83.21 &
  \textbf{0.18} &
  0.73 &
  82.23 &
  84.77 &
  2.54 &
  0.89 &
  \textbf{0.41} &
  \multicolumn{1}{c}{\textbf{-0.005}} \\ \hline
\multicolumn{1}{l}{Gemma-3-12b-it} &
  81.65 &
  79.93 &
  2.07 &
  5.92 &
  78.95 &
  86.16 &
  7.21 &
  0.88 &
  80.52 &
  79.69 &
  \textbf{1.00} &
  3.90 &
  77.97 &
  82.56 &
  4.59 &
  0.88 &
  \textbf{0.66} &
  \multicolumn{1}{c}{0.003} \\ \hline
\multicolumn{1}{l}{Gemma-3-27b-it} &
  82.51 &
  80.67 &
  2.21 &
  2.90 &
  80.67 &
  85.50 &
  4.83 &
  0.89 &
  82.30 &
  81.90 &
  \textbf{0.48} &
  0.64 &
  81.57 &
  83.87 &
  2.30 &
  0.90 &
  0.22 &
  \multicolumn{1}{c}{\textbf{-0.002}} \\ \hline
\multicolumn{1}{l}{Gpt-oss-20b} &
  82.72 &
  74.28 &
  10.13 &
  19.35 &
  74.28 &
  88.45 &
  14.17 &
  0.78 &
  72.29 &
  72.24 &
  \textbf{0.06} &
  1.11 &
  70.76 &
  74.04 &
  3.28 &
  0.69 &
  \textbf{0.06} &
  \multicolumn{1}{c}{0.086} \\ \hline
\multicolumn{1}{l}{Phi-4} &
  84.38 &
  83.13 &
  1.50 &
  2.55 &
  82.88 &
  87.14 &
  4.26 &
  0.88 &
  79.21 &
  77.97 &
  \textbf{1.49} &
  1.01 &
  77.89 &
  80.59 &
  2.70 &
  0.82 &
  \textbf{0.40} &
  \multicolumn{1}{c}{0.058} \\ \hline
\end{tabular}%
}

\label{tab:CSQA}
\end{table}
\begin{table}[ht]

\caption{Detailed results on \textbf{ARC} \cite{clark2018arc}.}
\resizebox{\textwidth}{!}{%
\begin{tabular}{lcccccccccccccccccc}
\cline{2-17}
 &
  \multicolumn{8}{c}{\textbf{S\&L}} &
  \multicolumn{8}{c}{\textbf{M\&D}} &
  \multicolumn{1}{l}{} &
  \multicolumn{1}{l}{} \\ \hline
 
\multicolumn{1}{c}{\textbf{Model name}} &
  \multicolumn{1}{c}{\textbf{$ACC_{Mean}$}} &
  \multicolumn{1}{c}{\textbf{$ACC_{ORG}$}} &
  \multicolumn{1}{c}{\textbf{$|\Delta ACC|$}} &
  \multicolumn{1}{c}{\textbf{$\sigma^2$}} &
  \multicolumn{1}{c}{\textbf{$ACC_{min}$}} &
  \multicolumn{1}{c}{\textbf{$ACC_{max}$}} &
  \multicolumn{1}{c}{\textbf{$\Delta_{max-min}$}} &
  \textbf{SCORE} &
  \multicolumn{1}{c}{\textbf{$ACC_{Mean}$}} &
  \multicolumn{1}{c}{\textbf{$ACC_{ORG}$}} &
  \multicolumn{1}{c}{\textbf{$|\Delta ACC|$}} &
  \multicolumn{1}{c}{\textbf{$\sigma^2$}} &
  \multicolumn{1}{c}{\textbf{$ACC_{min}$}} &
  \multicolumn{1}{c}{\textbf{$ACC_{max}$}} &
  \multicolumn{1}{c}{\textbf{$\Delta_{max-min}$}} &
  \textbf{SCORE} &
  \multicolumn{1}{c}{\textbf{$\sigma^2_{Ratio}$}} &
  \multicolumn{1}{c}{\textbf{$\Delta_{SCORE}$}} \\ \hline
\multicolumn{1}{l}{Deepseek-r1-0528-qwen3} &
  74.67 &
  64.68 &
  12.48 &
  25.60 &
  64.68 &
  78.44 &
  13.76 &
  0.57 &
  67.75 &
  66.26 &
  \textbf{1.86} &
  2.27 &
  65.90 &
  69.76 &
  3.86 &
  0.56 &
  \textbf{0.09} &
  \multicolumn{1}{c}{0.012} \\ \hline
\multicolumn{1}{l}{Deepseek-r1-distill-qwen} &
  74.28 &
  67.78 &
  8.12 &
  11.05 &
  67.78 &
  76.89 &
  9.11 &
  0.59 &
  70.10 &
  68.74 &
  \textbf{1.70} &
  1.49 &
  68.74 &
  72.01 &
  3.27 &
  0.64 &
  \textbf{0.14} &
  \multicolumn{1}{c}{\textbf{-0.051}} \\ \hline
\multicolumn{1}{l}{Llama-3.1-8b-instruct} &
  85.47 &
  87.57 &
  2.63 &
  10.05 &
  81.99 &
  90.61 &
  8.62 &
  0.89 &
  88.21 &
  88.19 &
  \textbf{0.02} &
  1.05 &
  86.56 &
  89.63 &
  3.07 &
  0.88 &
  \textbf{0.10} &
  \multicolumn{1}{c}{0.009} \\ \hline
\multicolumn{1}{l}{Ministral-3-8b-base} &
  82.96 &
  81.57 &
  1.74 &
  2.78 &
  81.12 &
  85.79 &
  4.67 &
  0.75 &
  77.72 &
  76.94 &
  \textbf{0.98} &
  2.26 &
  76.47 &
  80.64 &
  4.17 &
  0.72 &
  \textbf{0.81} &
  \multicolumn{1}{c}{0.024} \\ \hline
\multicolumn{1}{l}{Nvidia-nemotron-3-nano} &
  91.93 &
  90.33 &
  2.00 &
  3.47 &
  89.40 &
  94.31 &
  4.91 &
  0.87 &
  89.57 &
  88.92 &
  \textbf{0.81} &
  0.23 &
  88.92 &
  90.39 &
  1.47 &
  0.84 &
  \textbf{0.07} &
  \multicolumn{1}{c}{0.029} \\ \hline
\multicolumn{1}{l}{Nemotron-cascade-8b} &
  96.63 &
  96.42 &
  0.26 &
  0.11 &
  96.17 &
  97.13 &
  0.96 &
  0.97 &
  96.04 &
  95.91 &
  \textbf{0.16} &
  0.04 &
  95.83 &
  96.34 &
  0.51 &
  0.97 &
  \textbf{0.35} &
  \multicolumn{1}{c}{0.003} \\ \hline
\multicolumn{1}{l}{Qwen3-14b} &
  97.67 &
  97.44 &
  0.29 &
  0.03 &
  97.44 &
  97.86 &
  0.42 &
  0.98 &
  95.22 &
  95.07 &
  \textbf{0.19} &
  0.10 &
  94.81 &
  95.72 &
  0.91 &
  0.94 &
  2.82 &
  \multicolumn{1}{c}{0.044} \\ \hline
\multicolumn{1}{l}{Qwen3-30b-a3b-instruct} &
  98.14 &
  97.89 &
  0.31 &
  0.02 &
  97.89 &
  98.31 &
  0.42 &
  0.99 &
  97.09 &
  97.18 &
  \textbf{0.12} &
  0.03 &
  96.82 &
  97.29 &
  0.47 &
  0.97 &
  1.08 &
  \multicolumn{1}{c}{0.016} \\ \hline
\multicolumn{1}{l}{Qwen3-8b} &
  97.32 &
  96.93 &
  0.48 &
  0.08 &
  96.93 &
  97.69 &
  0.76 &
  0.98 &
  97.00 &
  96.65 &
  \textbf{0.44} &
  0.08 &
  96.65 &
  97.44 &
  0.79 &
  0.97 &
  \textbf{0.96} &
  \multicolumn{1}{c}{0.003} \\ \hline
\multicolumn{1}{l}{Gemma-3-12b-it} &
  95.92 &
  96.11 &
  0.24 &
  0.15 &
  95.24 &
  96.42 &
  1.18 &
  0.97 &
  95.71 &
  95.86 &
  \textbf{0.19} &
  0.06 &
  95.26 &
  95.94 &
  0.68 &
  0.97 &
  \textbf{0.37} &
  \multicolumn{1}{c}{0.001} \\ \hline
\multicolumn{1}{l}{Gemma-3-27b-it} &
  96.61 &
  96.62 &
  \textbf{0.01} &
  0.06 &
  96.31 &
  97.07 &
  0.76 &
  0.98 &
  96.61 &
  96.65 &
  0.05 &
  0.02 &
  96.36 &
  96.79 &
  0.43 &
  0.98 &
  \textbf{0.38} &
  \multicolumn{1}{c}{\textbf{-0.001}} \\ \hline
\multicolumn{1}{l}{Gpt-oss-20b} &
  92.53 &
  87.88 &
  5.82 &
  5.50 &
  87.88 &
  94.17 &
  6.29 &
  0.88 &
  88.75 &
  89.46 &
  \textbf{0.89} &
  0.37 &
  87.85 &
  89.46 &
  1.61 &
  0.86 &
  \textbf{0.07} &
  \multicolumn{1}{c}{0.020} \\ \hline
\multicolumn{1}{l}{Phi-4} &
  97.00 &
  96.93 &
  0.09 &
  0.07 &
  96.76 &
  97.49 &
  0.73 &
  0.98 &
  93.46 &
  93.38 &
  0.09 &
  0.05 &
  93.07 &
  93.71 &
  0.64 &
  0.91 &
  \textbf{0.75} &
  \multicolumn{1}{c}{0.064} \\ \hline
\end{tabular}%
}

\label{tab:ARC}
\end{table}
\begin{table}[ht]
\caption{Detailed results on \textbf{GPQA} \cite{rein2024gpqa}.}
\resizebox{\textwidth}{!}{%
\begin{tabular}{lcccccccccccccccccc}
\cline{2-17}
 &
  \multicolumn{8}{c}{\textbf{S\&L}} &
  \multicolumn{8}{c}{\textbf{M\&D}} &
  \multicolumn{1}{l}{} &
  \multicolumn{1}{l}{} \\ \hline
 
\multicolumn{1}{c}{\textbf{Model name}} &
  \multicolumn{1}{c}{\textbf{$ACC_{Mean}$}} &
  \multicolumn{1}{c}{\textbf{$ACC_{ORG}$}} &
  \multicolumn{1}{c}{\textbf{$|\Delta ACC|$}} &
  \multicolumn{1}{c}{\textbf{$\sigma^2$}} &
  \multicolumn{1}{c}{\textbf{$ACC_{min}$}} &
  \multicolumn{1}{c}{\textbf{$ACC_{max}$}} &
  \multicolumn{1}{c}{\textbf{$\Delta_{max-min}$}} &
  \textbf{SCORE} &
  \multicolumn{1}{c}{\textbf{$ACC_{Mean}$}} &
  \multicolumn{1}{c}{\textbf{$ACC_{ORG}$}} &
  \multicolumn{1}{c}{\textbf{$|\Delta ACC|$}} &
  \multicolumn{1}{c}{\textbf{$\sigma^2$}} &
  \multicolumn{1}{c}{\textbf{$ACC_{min}$}} &
  \multicolumn{1}{c}{\textbf{$ACC_{max}$}} &
  \multicolumn{1}{c}{\textbf{$\Delta_{max-min}$}} &
  \textbf{SCORE} &
  \multicolumn{1}{c}{\textbf{$\sigma^2_{Ratio}$}} &
  \multicolumn{1}{c}{\textbf{$\Delta_{SCORE}$}} \\ \hline
\multicolumn{1}{l}{Deepseek-r1-0528-qwen3} &
  25.31 &
  27.23 &
  2.40 &
  15.53 &
  18.97 &
  29.24 &
  10.27 &
  0.39 &
  23.35 &
  23.44 &
  \textbf{0.11} &
  3.51 &
  20.54 &
  25.67 &
  5.13 &
  0.44 &
  \textbf{0.23} &
  \multicolumn{1}{c}{\textbf{-0.043}} \\ \hline
\multicolumn{1}{l}{Deepseek-r1-distill-qwen} &
  31.56 &
  33.93 &
  2.96 &
  16.11 &
  24.55 &
  36.61 &
  12.06 &
  0.43 &
  28.03 &
  28.57 &
  \textbf{0.67} &
  2.97 &
  25.22 &
  30.36 &
  5.14 &
  0.45 &
  \textbf{0.18} &
  \multicolumn{1}{c}{\textbf{-0.019}} \\ \hline
\multicolumn{1}{l}{Llama-3.1-8b-instruct} &
  27.81 &
  29.46 &
  2.06 &
  12.79 &
  24.33 &
  33.93 &
  9.60 &
  0.42 &
  30.09 &
  30.58 &
  \textbf{0.61} &
  8.25 &
  27.68 &
  35.49 &
  7.81 &
  0.45 &
  \textbf{0.65} &
  \multicolumn{1}{c}{\textbf{-0.028}} \\ \hline
\multicolumn{1}{l}{Ministral-3-8b-base} &
  30.71 &
  33.71 &
  3.75 &
  7.47 &
  27.23 &
  34.15 &
  6.92 &
  0.30 &
  27.55 &
  24.78 &
  \textbf{3.46} &
  4.91 &
  24.78 &
  31.03 &
  6.25 &
  0.39 &
  \textbf{0.66} &
  \multicolumn{1}{c}{\textbf{-0.095}} \\ \hline
\multicolumn{1}{l}{Nvidia-nemotron-3-nano} &
  36.65 &
  39.51 &
  3.57 &
  18.45 &
  30.58 &
  42.86 &
  12.28 &
  0.41 &
  34.82 &
  37.05 &
  \textbf{2.79} &
  1.35 &
  33.93 &
  37.05 &
  3.12 &
  0.46 &
  \textbf{0.07} &
  \multicolumn{1}{c}{\textbf{-0.052}} \\ \hline
\multicolumn{1}{l}{Nemotron-cascade-8b} &
  49.46 &
  50.89 &
  \textbf{1.78} &
  1.51 &
  47.54 &
  50.89 &
  3.35 &
  0.56 &
  49.64 &
  52.01 &
  2.96 &
  2.32 &
  47.77 &
  52.01 &
  4.24 &
  0.57 &
  1.54 &
  \multicolumn{1}{c}{\textbf{-0.007}} \\ \hline
\multicolumn{1}{l}{Qwen3-14b} &
  38.30 &
  37.95 &
  \textbf{0.44} &
  1.54 &
  37.05 &
  40.62 &
  3.57 &
  0.59 &
  42.19 &
  39.29 &
  3.62 &
  4.82 &
  39.29 &
  44.20 &
  4.91 &
  0.57 &
  3.12 &
  \multicolumn{1}{c}{0.025} \\ \hline
\multicolumn{1}{l}{Qwen3-30b-a3b-instruct} &
  50.63 &
  49.78 &
  \textbf{1.06} &
  3.92 &
  47.99 &
  54.02 &
  6.03 &
  0.63 &
  49.33 &
  50.22 &
  1.11 &
  5.82 &
  46.65 &
  53.35 &
  6.70 &
  0.56 &
  1.49 &
  \multicolumn{1}{c}{0.069} \\ \hline
\multicolumn{1}{l}{Qwen3-8b} &
  39.64 &
  40.18 &
  \textbf{0.67} &
  9.25 &
  35.27 &
  43.75 &
  8.48 &
  0.57 &
  41.16 &
  39.73 &
  1.79 &
  3.57 &
  39.73 &
  44.87 &
  5.14 &
  0.55 &
  \textbf{0.39} &
  \multicolumn{1}{c}{0.020} \\ \hline
\multicolumn{1}{l}{Gemma-3-12b-it} &
  32.94 &
  35.71 &
  3.46 &
  9.60 &
  29.46 &
  37.50 &
  8.04 &
  0.48 &
  33.66 &
  33.26 &
  \textbf{0.50} &
  0.92 &
  32.81 &
  35.49 &
  2.68 &
  0.55 &
  \textbf{0.10} &
  \multicolumn{1}{c}{\textbf{-0.063}} \\ \hline
\multicolumn{1}{l}{Gemma-3-27b-it} &
  40.36 &
  40.18 &
  \textbf{0.23} &
  6.10 &
  37.95 &
  45.09 &
  7.14 &
  0.60 &
  40.76 &
  40.40 &
  0.45 &
  5.32 &
  37.95 &
  44.87 &
  6.92 &
  0.62 &
  \textbf{0.87} &
  \multicolumn{1}{c}{\textbf{-0.019}} \\ \hline
\multicolumn{1}{l}{Gpt-oss-20b} &
  45.62 &
  46.65 &
  1.28 &
  2.64 &
  43.08 &
  47.77 &
  4.69 &
  0.47 &
  36.65 &
  36.38 &
  \textbf{0.34} &
  4.61 &
  33.04 &
  39.51 &
  6.47 &
  0.49 &
  1.74 &
  \multicolumn{1}{c}{\textbf{-0.022}} \\ \hline
\multicolumn{1}{l}{Phi-4} &
  35.49 &
  36.61 &
  1.39 &
  14.54 &
  31.70 &
  42.41 &
  10.71 &
  0.50 &
  33.75 &
  34.15 &
  \textbf{0.50} &
  2.95 &
  31.03 &
  36.38 &
  5.35 &
  0.50 &
  \textbf{0.20} &
  \multicolumn{1}{c}{\textbf{-0.004}} \\ \hline
\end{tabular}%
}

\label{tab:GPQA}
\end{table}
\begin{table}[ht]
\caption{Detailed results on \textbf{MMLU-Pro} \cite{wang2024mmlu}.}

\resizebox{\textwidth}{!}{%
\begin{tabular}{lcccccccccccccccccc}
\cline{2-17}
 &
  \multicolumn{8}{c}{\textbf{S\&L}} &
  \multicolumn{8}{c}{\textbf{M\&D}} &
  \multicolumn{1}{l}{} &
  \multicolumn{1}{l}{} \\ \hline
 
\multicolumn{1}{c}{\textbf{Model name}} &
  \multicolumn{1}{c}{\textbf{$ACC_{Mean}$}} &
  \multicolumn{1}{c}{\textbf{$ACC_{ORG}$}} &
  \multicolumn{1}{c}{\textbf{$|\Delta ACC|$}} &
  \multicolumn{1}{c}{\textbf{$\sigma^2$}} &
  \multicolumn{1}{c}{\textbf{$ACC_{min}$}} &
  \multicolumn{1}{c}{\textbf{$ACC_{max}$}} &
  \multicolumn{1}{c}{\textbf{$\Delta_{max-min}$}} &
  \textbf{SCORE} &
  \multicolumn{1}{c}{\textbf{$ACC_{Mean}$}} &
  \multicolumn{1}{c}{\textbf{$ACC_{ORG}$}} &
  \multicolumn{1}{c}{\textbf{$|\Delta ACC|$}} &
  \multicolumn{1}{c}{\textbf{$\sigma^2$}} &
  \multicolumn{1}{c}{\textbf{$ACC_{min}$}} &
  \multicolumn{1}{c}{\textbf{$ACC_{max}$}} &
  \multicolumn{1}{c}{\textbf{$\Delta_{max-min}$}} &
  \textbf{SCORE} &
  \multicolumn{1}{c}{\textbf{$\sigma^2_{Ratio}$}} &
  \multicolumn{1}{c}{\textbf{$\Delta_{SCORE}$}} \\ \hline
\multicolumn{1}{l}{Deepseek-r1-0528-qwen3} &
  31.57 &
  35.30 &
  \textbf{4.11} &
  4.07 &
  27.66 &
  35.30 &
  7.64 &
  0.33 &
  31.29 &
  36.12 &
  5.31 &
  3.85 &
  28.95 &
  36.12 &
  7.17 &
  0.29 &
  \textbf{0.94} &
  \multicolumn{1}{l}{0.047} \\ \hline
\multicolumn{1}{l}{Deepseek-r1-distill-qwen} &
  31.56 &
  31.33 &
  \textbf{0.25} &
  33.83 &
  21.53 &
  39.74 &
  18.21 &
  0.34 &
  27.71 &
  28.47 &
  0.83 &
  1.39 &
  26.10 &
  29.72 &
  3.62 &
  0.30 &
  \textbf{0.04} &
  \multicolumn{1}{l}{0.036} \\ \hline
\multicolumn{1}{l}{Llama-3.1-8b-instruct} &
  40.79 &
  43.33 &
  \textbf{2.79} &
  6.24 &
  36.77 &
  44.27 &
  7.50 &
  0.55 &
  40.07 &
  42.67 &
  2.86 &
  13.23 &
  35.52 &
  47.78 &
  12.26 &
  0.51 &
  2.12 &
  \multicolumn{1}{l}{0.041} \\ \hline
\multicolumn{1}{l}{Ministral-3-8b-base} &
  24.48 &
  42.03 &
  19.30 &
  45.36 &
  18.28 &
  42.03 &
  23.75 &
  0.28 &
  27.60 &
  41.29 &
  \textbf{15.06} &
  26.15 &
  22.74 &
  41.29 &
  18.55 &
  0.32 &
  \textbf{0.58} &
  \multicolumn{1}{l}{\textbf{-0.043}} \\ \hline
\multicolumn{1}{l}{Nvidia-nemotron-3-nano} &
  55.31 &
  61.61 &
  6.93 &
  9.62 &
  49.20 &
  61.61 &
  12.41 &
  0.54 &
  47.77 &
  53.31 &
  \textbf{6.10} &
  5.96 &
  44.92 &
  53.31 &
  8.39 &
  0.45 &
  \textbf{0.62} &
  \multicolumn{1}{l}{0.094} \\ \hline
\multicolumn{1}{l}{Nemotron-cascade-8b} &
  57.26 &
  61.96 &
  \textbf{5.17} &
  9.43 &
  52.41 &
  61.96 &
  9.55 &
  0.66 &
  49.55 &
  58.03 &
  9.33 &
  15.18 &
  42.93 &
  58.03 &
  15.10 &
  0.56 &
  1.61 &
  \multicolumn{1}{l}{0.098} \\ \hline
\multicolumn{1}{l}{Qwen3-14b} &
  63.69 &
  68.74 &
  \textbf{5.55} &
  8.77 &
  59.35 &
  68.74 &
  9.39 &
  0.70 &
  56.58 &
  67.30 &
  11.79 &
  24.65 &
  48.35 &
  67.30 &
  18.95 &
  0.61 &
  2.81 &
  \multicolumn{1}{l}{0.085} \\ \hline
\multicolumn{1}{l}{Qwen3-30b-a3b-instruct} &
  74.53 &
  74.55 &
  \textbf{0.02} &
  0.72 &
  73.44 &
  76.33 &
  2.89 &
  0.84 &
  66.97 &
  70.85 &
  4.26 &
  2.22 &
  65.54 &
  70.85 &
  5.31 &
  0.72 &
  3.08 &
  \multicolumn{1}{l}{0.115} \\ \hline
\multicolumn{1}{l}{Qwen3-8b} &
  61.67 &
  63.78 &
  \textbf{2.33} &
  2.73 &
  58.97 &
  64.18 &
  5.21 &
  0.72 &
  58.35 &
  61.02 &
  2.94 &
  3.20 &
  55.69 &
  61.02 &
  5.33 &
  0.68 &
  1.17 &
  \multicolumn{1}{l}{0.039} \\ \hline
\multicolumn{1}{l}{Gemma-3-12b-it} &
  49.17 &
  57.62 &
  9.30 &
  19.75 &
  42.28 &
  57.62 &
  15.34 &
  0.55 &
  55.58 &
  57.36 &
  \textbf{1.96} &
  11.81 &
  50.58 &
  62.14 &
  11.56 &
  0.66 &
  \textbf{0.60} &
  \multicolumn{1}{l}{\textbf{-0.114}} \\ \hline
\multicolumn{1}{l}{Gemma-3-27b-it} &
  57.49 &
  64.35 &
  \textbf{7.55} &
  22.67 &
  50.20 &
  67.65 &
  17.45 &
  0.62 &
  54.65 &
  64.10 &
  10.40 &
  22.44 &
  48.75 &
  64.10 &
  15.35 &
  0.60 &
  \textbf{0.99} &
  \multicolumn{1}{l}{0.027} \\ \hline
\multicolumn{1}{l}{Gpt-oss-20b} &
  54.84 &
  57.38 &
  \textbf{2.80} &
  3.53 &
  52.48 &
  58.01 &
  5.53 &
  0.47 &
  42.78 &
  54.00 &
  12.34 &
  13.89 &
  39.44 &
  54.00 &
  14.56 &
  0.42 &
  3.93 &
  \multicolumn{1}{l}{0.047} \\ \hline
\multicolumn{1}{l}{Phi-4} &
  43.18 &
  58.47 &
  16.82 &
  37.01 &
  35.83 &
  58.47 &
  22.64 &
  0.40 &
  39.58 &
  52.79 &
  \textbf{14.54} &
  28.26 &
  34.32 &
  52.79 &
  18.47 &
  0.35 &
  \textbf{0.76} &
  \multicolumn{1}{l}{0.050} \\ \hline
\end{tabular}%
}

\label{tab:MMLU-Pro}
\end{table}
\begin{table}[ht]

\caption{Detailed results on \textbf{the subset of INCLUDE} \cite{romanouinclude} consisting of French, German, Italian and Spanish languages.}

\resizebox{\textwidth}{!}{%
\begin{tabular}{lcccccccccccccccccc}
\cline{2-17}
 &
  \multicolumn{8}{c}{\textbf{S\&L}} &
  \multicolumn{8}{c}{\textbf{M\&D}} &
  \multicolumn{1}{l}{} &
  \multicolumn{1}{l}{} \\ \hline
 
\multicolumn{1}{c}{\textbf{Model name}} &
  \multicolumn{1}{c}{\textbf{$ACC_{Mean}$}} &
  \multicolumn{1}{c}{\textbf{$ACC_{ORG}$}} &
  \multicolumn{1}{c}{\textbf{$|\Delta ACC|$}} &
  \multicolumn{1}{c}{\textbf{$\sigma^2$}} &
  \multicolumn{1}{c}{\textbf{$ACC_{min}$}} &
  \multicolumn{1}{c}{\textbf{$ACC_{max}$}} &
  \multicolumn{1}{c}{\textbf{$\Delta_{max-min}$}} &
  \textbf{SCORE} &
  \multicolumn{1}{c}{\textbf{$ACC_{Mean}$}} &
  \multicolumn{1}{c}{\textbf{$ACC_{ORG}$}} &
  \multicolumn{1}{c}{\textbf{$|\Delta ACC|$}} &
  \multicolumn{1}{c}{\textbf{$\sigma^2$}} &
  \multicolumn{1}{c}{\textbf{$ACC_{min}$}} &
  \multicolumn{1}{c}{\textbf{$ACC_{max}$}} &
  \multicolumn{1}{c}{\textbf{$\Delta_{max-min}$}} &
  \textbf{SCORE} &
  \multicolumn{1}{c}{\textbf{$\sigma^2_{Ratio}$}} &
  \multicolumn{1}{c}{\textbf{$\Delta_{SCORE}$}} \\ \hline
\multicolumn{1}{l}{Deepseek-r1-0528-qwen3} &
  52.72 &
  41.24 &
  14.35 &
  39.16 &
  41.24 &
  58.88 &
  17.64 &
  0.38 &
  44.06 &
  43.90 &
  \textbf{0.20} &
  1.29 &
  41.97 &
  45.29 &
  3.32 &
  0.43 &
  \textbf{0.03} &
  \multicolumn{1}{c}{\textbf{-0.052}} \\ \hline
\multicolumn{1}{l}{Deepseek-r1-distill-qwen} &
  39.61 &
  29.77 &
  12.30 &
  27.08 &
  29.77 &
  45.35 &
  15.58 &
  0.38 &
  31.32 &
  29.35 &
  \textbf{2.46} &
  3.19 &
  28.93 &
  32.91 &
  3.98 &
  0.41 &
  \textbf{0.12} &
  \multicolumn{1}{c}{\textbf{-0.032}} \\ \hline
\multicolumn{1}{l}{Llama-3.1-8b-instruct} &
  60.14 &
  59.24 &
  1.12 &
  0.25 &
  59.24 &
  60.81 &
  1.57 &
  0.67 &
  56.86 &
  57.49 &
  \textbf{0.79} &
  0.89 &
  55.07 &
  57.73 &
  2.66 &
  0.64 &
  3.48 &
  \multicolumn{1}{c}{0.030} \\ \hline
\multicolumn{1}{l}{Ministral-3-8b-base} &
  56.65 &
  54.83 &
  2.28 &
  4.35 &
  54.35 &
  59.96 &
  5.61 &
  0.54 &
  52.63 &
  51.99 &
  \textbf{0.80} &
  2.33 &
  51.39 &
  55.62 &
  4.23 &
  0.55 &
  \textbf{0.53} &
  \multicolumn{1}{c}{\textbf{-0.008}} \\ \hline
\multicolumn{1}{l}{Nvidia-nemotron-3-nano} &
  72.68 &
  69.69 &
  3.74 &
  10.94 &
  69.69 &
  78.80 &
  9.11 &
  0.75 &
  72.77 &
  72.40 &
  \textbf{0.46} &
  0.65 &
  71.92 &
  74.28 &
  2.36 &
  0.78 &
  \textbf{0.06} &
  \multicolumn{1}{c}{\textbf{-0.032}} \\ \hline
\multicolumn{1}{l}{Nemotron-cascade-8b} &
  69.47 &
  67.15 &
  2.90 &
  5.77 &
  67.15 &
  73.13 &
  5.98 &
  0.77 &
  67.73 &
  67.93 &
  \textbf{0.25} &
  0.89 &
  66.67 &
  69.08 &
  2.41 &
  0.77 &
  \textbf{0.15} &
  \multicolumn{1}{c}{0.005} \\ \hline
\multicolumn{1}{l}{Qwen3-14b} &
  77.98 &
  76.57 &
  1.77 &
  2.14 &
  76.51 &
  80.43 &
  3.92 &
  0.85 &
  76.81 &
  77.36 &
  \textbf{0.68} &
  0.86 &
  75.18 &
  77.60 &
  2.42 &
  0.84 &
  \textbf{0.40} &
  \multicolumn{1}{c}{0.009} \\ \hline
\multicolumn{1}{l}{Qwen3-30b-a3b-instruct} &
  79.01 &
  78.02 &
  1.24 &
  3.89 &
  76.21 &
  81.82 &
  5.61 &
  0.87 &
  76.10 &
  75.91 &
  \textbf{0.24} &
  1.19 &
  74.52 &
  77.78 &
  3.26 &
  0.83 &
  \textbf{0.31} &
  \multicolumn{1}{c}{0.043} \\ \hline
\multicolumn{1}{l}{Qwen3-8b} &
  74.58 &
  72.46 &
  2.64 &
  4.72 &
  72.04 &
  78.02 &
  5.98 &
  0.82 &
  72.67 &
  71.92 &
  \textbf{0.94} &
  0.89 &
  71.32 &
  73.91 &
  2.59 &
  0.83 &
  \textbf{0.19} &
  \multicolumn{1}{c}{\textbf{-0.003}} \\ \hline
\multicolumn{1}{l}{Gemma-3-12b-it} &
  71.86 &
  71.56 &
  0.38 &
  \textbf{0.39} &
  71.32 &
  72.95 &
  1.63 &
  0.81 &
  71.53 &
  70.65 &
  1.10 &
  0.38 &
  70.65 &
  72.58 &
  1.93 &
  0.82 &
  \textbf{0.97} &
  \multicolumn{1}{c}{\textbf{-0.009}} \\ \hline
\multicolumn{1}{l}{Gemma-3-27b-it} &
  76.85 &
  75.91 &
  1.17 &
  \textbf{0.45} &
  75.91 &
  77.66 &
  1.75 &
  0.85 &
  75.95 &
  74.88 &
  1.34 &
  0.63 &
  74.88 &
  76.93 &
  2.05 &
  0.86 &
  1.41 &
  \multicolumn{1}{c}{\textbf{-0.008}} \\ \hline
\multicolumn{1}{l}{Gpt-oss-20b} &
  74.58 &
  66.12 &
  10.57 &
  18.66 &
  66.12 &
  77.90 &
  11.78 &
  0.71 &
  67.21 &
  67.03 &
  \textbf{0.22} &
  0.13 &
  66.67 &
  67.69 &
  1.02 &
  0.67 &
  \textbf{0.01} &
  \multicolumn{1}{c}{0.040} \\ \hline
\multicolumn{1}{l}{Phi-4} &
  72.25 &
  71.14 &
  1.39 &
  2.42 &
  69.99 &
  74.52 &
  4.53 &
  0.80 &
  70.71 &
  70.89 &
  \textbf{0.22} &
  0.74 &
  69.02 &
  71.32 &
  2.30 &
  0.77 &
  \textbf{0.31} &
  \multicolumn{1}{c}{0.038} \\ \hline
\end{tabular}%
}

\label{tab:INCLUDE}
\end{table}

\subsection{On the SCORE paradox - capturing bias}
\label{appendix:score_paradox}
\begin{figure*}[!ht]
  \begin{center}
    \centerline{\includegraphics[width=\textwidth]{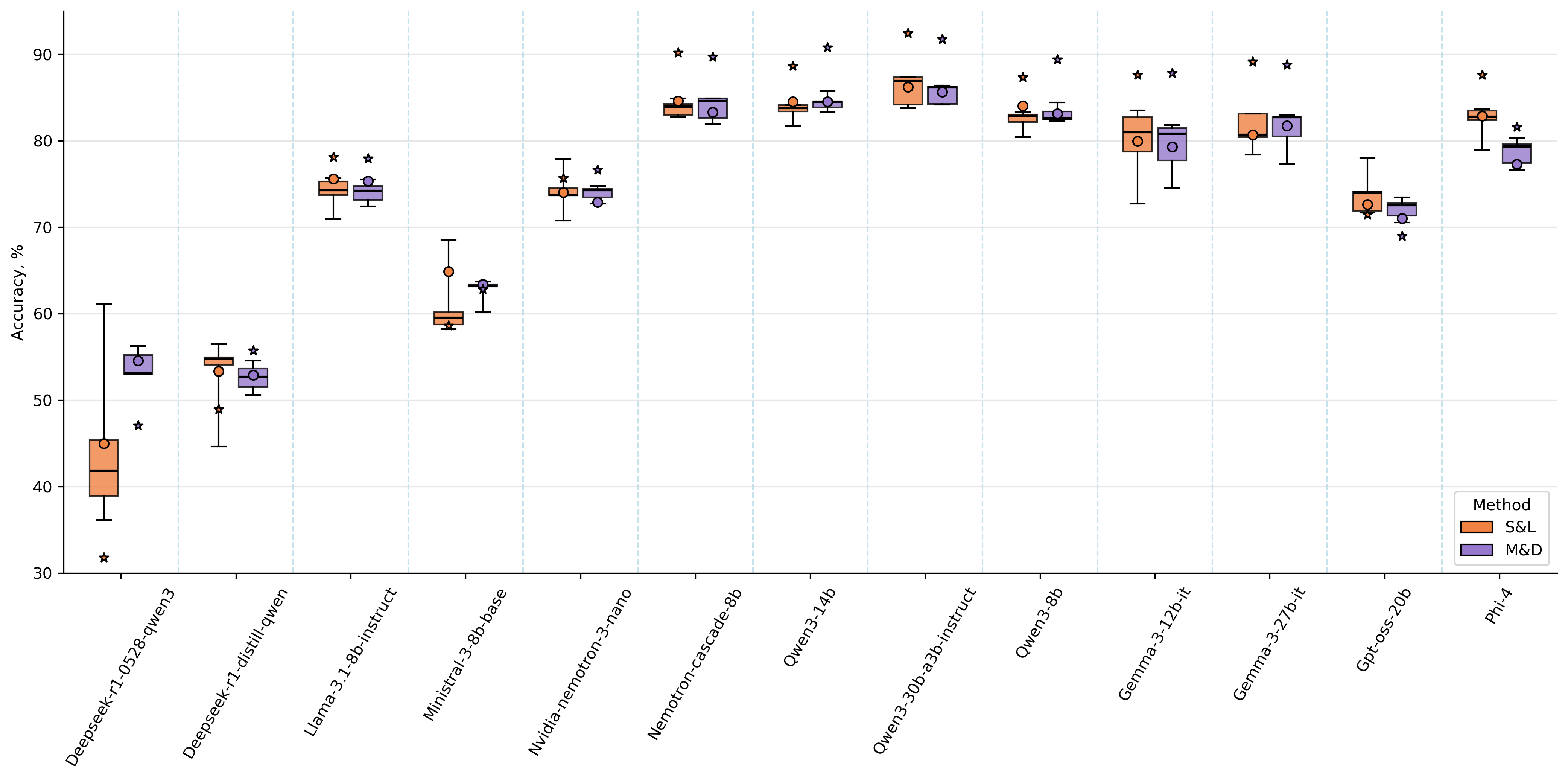}}
    \caption{Comparison of the matched prediction with dashes as labels (M\&D; our method) with standard letter prediction with letters
    as labels (S\&L) on CommonsenseQA \cite{talmor-etal-2019-commonsenseqa} with a 5-shot prompt \textbf{that preserves the original few-shot answers distribution}. The boxes illustrate the model performance under all possibilities of ”answer-moving attacks”, where the whiskers indicate the minimum and maximum accuracy for each model. Each dot represents the performance of the original permutations. Additionally, each star symbolizes a SCORE \cite{nalbandyan-etal-2025-score} robustness metric.}
    \label{fig:noVal_CSQA}
  \end{center}
\end{figure*}

While positive SCORE difference values on real-life benchmarks might look like failure modes, we note that a model that gets every question wrong in their original permutation, but correctly identifies the present biases under-attack permutations, would achieve a SCORE of 0.6, 0.66, and 0.81, when the number of possible answers is four, five, and ten, respectively. To show this paradox, we demonstrate additional result on CommonsenseQA \cite{talmor-etal-2019-commonsenseqa} without changing the few-shot prompt examples, just like in GPQA \cite{rein2024gpqa} evaluation. We present the results in Figure~\ref{fig:noVal_CSQA} and the detailed results in Table~\ref{tab:CSQA_val} and Table~\ref{tab:score_diff}.

As shown in Figure~\ref{fig:noVal_CSQA} and Table~\ref{tab:score_diff}, the biases introduced by the few-shot prompt also significantly affect the SCORE score under the S\&L metric. Under bias, SCORE M\&D is lower than SCORE S\&L by 1\%. However, without permuting any examples from the original few-shot prompt, we observe an increased mean M\&D SCORE compared to S\&L by 1.4\%. Interestingly, we observe that SCORE under the S\&L benchmark changes drastically depending on whether the few-shot prompt contains a malicious answer distribution. In contrast, the M\&D score remains relatively stable, demonstrating that our evaluation protocol is robust to the few-shot prompt bias.

\begin{table}[ht]

\caption{Detailed results on \textbf{CommonsenseQA} \cite{talmor-etal-2019-commonsenseqa} without inducing malicious permutations on the few-shot prompt.}

\resizebox{\textwidth}{!}{%
\begin{tabular}{lcccccccccccccccccc}
\cline{2-17}
 &
  \multicolumn{8}{c}{\textbf{S\&L}} &
  \multicolumn{8}{c}{\textbf{M\&D}} &
  \multicolumn{1}{l}{} &
  \multicolumn{1}{l}{} \\ \hline
 
\multicolumn{1}{c}{\textbf{Model name}} &
  \multicolumn{1}{c}{\textbf{$ACC_{Mean}$}} &
  \multicolumn{1}{c}{\textbf{$ACC_{ORG}$}} &
  \multicolumn{1}{c}{\textbf{$|\Delta ACC|$}} &
  \multicolumn{1}{c}{\textbf{$\sigma^2$}} &
  \multicolumn{1}{c}{\textbf{$ACC_{min}$}} &
  \multicolumn{1}{c}{\textbf{$ACC_{max}$}} &
  \multicolumn{1}{c}{\textbf{$\Delta_{max-min}$}} &
  \textbf{SCORE} &
  \multicolumn{1}{c}{\textbf{$ACC_{Mean}$}} &
  \multicolumn{1}{c}{\textbf{$ACC_{ORG}$}} &
  \multicolumn{1}{c}{\textbf{$|\Delta ACC|$}} &
  \multicolumn{1}{c}{\textbf{$\sigma^2$}} &
  \multicolumn{1}{c}{\textbf{$ACC_{min}$}} &
  \multicolumn{1}{c}{\textbf{$ACC_{max}$}} &
  \multicolumn{1}{c}{\textbf{$\Delta_{max-min}$}} &
  \textbf{SCORE} &
  \multicolumn{1}{c}{\textbf{$\sigma^2_{Ratio}$}} &
  \multicolumn{1}{c}{\textbf{$\Delta_{SCORE}$}} \\ \hline
\multicolumn{1}{l}{Deepseek-r1-0528-qwen3} &
  44.72 &
  44.96 &
  \textbf{0.29} &
  64.14 &
  36.12 &
  61.10 &
  24.98 &
  0.32 &
  54.18 &
  54.55 &
  0.45 &
  1.60 &
  52.99 &
  56.27 &
  3.28 &
  0.47 &
  \textbf{0.02} &
  \multicolumn{1}{c}{\textbf{-0.153}} \\ \hline
\multicolumn{1}{l}{Deepseek-r1-distill-qwen} &
  53.04 &
  53.32 &
  0.33 &
  15.07 &
  44.64 &
  56.51 &
  11.87 &
  0.49 &
  52.65 &
  52.91 &
  \textbf{0.31} &
  1.68 &
  50.61 &
  54.55 &
  3.94 &
  0.56 &
  \textbf{0.11} &
  \multicolumn{1}{c}{\textbf{-0.068}} \\ \hline
\multicolumn{1}{l}{Llama-3.1-8b-instruct} &
  74.24 &
  75.59 &
  1.62 &
  2.70 &
  70.93 &
  75.68 &
  4.75 &
  0.78 &
  74.23 &
  75.35 &
  \textbf{1.35} &
  1.29 &
  72.40 &
  75.51 &
  3.11 &
  0.78 &
  \textbf{0.48} &
  \multicolumn{1}{c}{0.002} \\ \hline
\multicolumn{1}{l}{Ministral-3-8b-base} &
  61.68 &
  64.86 &
  3.81 &
  14.12 &
  58.23 &
  68.55 &
  10.32 &
  0.59 &
  62.84 &
  63.39 &
  \textbf{0.65} &
  1.43 &
  60.20 &
  63.72 &
  3.52 &
  0.63 &
  \textbf{0.10} &
  \multicolumn{1}{c}{\textbf{-0.042}} \\ \hline
\multicolumn{1}{l}{Nvidia-nemotron-3-nano} &
  74.11 &
  74.04 &
  \textbf{0.08} &
  4.34 &
  70.76 &
  77.89 &
  7.13 &
  0.76 &
  73.76 &
  72.89 &
  1.05 &
  0.61 &
  72.73 &
  74.77 &
  2.04 &
  0.77 &
  \textbf{0.14} &
  \multicolumn{1}{c}{\textbf{-0.010}} \\ \hline
\multicolumn{1}{l}{Nemotron-cascade-8b} &
  83.91 &
  84.60 &
  0.83 &
  0.66 &
  82.72 &
  84.93 &
  2.21 &
  0.90 &
  83.71 &
  83.29 &
  \textbf{0.51} &
  1.39 &
  81.90 &
  84.93 &
  3.03 &
  0.90 &
  2.11 &
  \multicolumn{1}{c}{0.005} \\ \hline
\multicolumn{1}{l}{Qwen3-14b} &
  83.61 &
  84.52 &
  1.10 &
  0.82 &
  81.74 &
  84.52 &
  2.78 &
  0.89 &
  84.41 &
  84.52 &
  \textbf{0.13} &
  0.56 &
  83.29 &
  85.75 &
  2.46 &
  0.91 &
  \textbf{0.69} &
  \multicolumn{1}{c}{\textbf{-0.021}} \\ \hline
\multicolumn{1}{l}{Qwen3-30b-a3b-instruct} &
  85.98 &
  86.24 &
  0.31 &
  2.16 &
  83.78 &
  87.39 &
  3.61 &
  0.92 &
  85.49 &
  85.67 &
  \textbf{0.22} &
  0.84 &
  84.19 &
  86.40 &
  2.21 &
  0.92 &
  \textbf{0.39} &
  \multicolumn{1}{c}{0.007} \\ \hline
\multicolumn{1}{l}{Qwen3-8b} &
  82.64 &
  84.03 &
  1.67 &
  1.28 &
  80.43 &
  84.03 &
  3.60 &
  0.87 &
  83.05 &
  83.13 &
  \textbf{0.10} &
  0.53 &
  82.31 &
  84.44 &
  2.13 &
  0.89 &
  \textbf{0.41} &
  \multicolumn{1}{c}{\textbf{-0.021}} \\ \hline
\multicolumn{1}{l}{Gemma-3-12b-it} &
  79.77 &
  79.93 &
  0.19 &
  12.52 &
  72.73 &
  83.54 &
  10.81 &
  0.88 &
  79.28 &
  79.28 &
  \textbf{0.00} &
  6.46 &
  74.53 &
  81.82 &
  7.29 &
  0.88 &
  \textbf{0.52} &
  \multicolumn{1}{c}{\textbf{-0.002}} \\ \hline
\multicolumn{1}{l}{Gemma-3-27b-it} &
  81.07 &
  80.67 &
  0.48 &
  2.74 &
  78.38 &
  83.13 &
  4.75 &
  0.89 &
  81.34 &
  81.74 &
  0.48 &
  3.96 &
  77.31 &
  82.96 &
  5.65 &
  0.89 &
  1.44 &
  \multicolumn{1}{c}{0.003} \\ \hline
\multicolumn{1}{l}{Gpt-oss-20b} &
  73.72 &
  72.65 &
  1.29 &
  4.50 &
  71.66 &
  77.97 &
  6.31 &
  0.71 &
  71.95 &
  71.01 &
  \textbf{1.13} &
  1.12 &
  70.52 &
  73.46 &
  2.94 &
  0.69 &
  \textbf{0.25} &
  \multicolumn{1}{c}{0.025} \\ \hline
\multicolumn{1}{l}{Phi-4} &
  82.36 &
  82.88 &
  \textbf{0.62} &
  2.52 &
  78.95 &
  83.70 &
  4.75 &
  0.88 &
  78.43 &
  77.31 &
  1.35 &
  1.94 &
  76.58 &
  80.34 &
  3.76 &
  0.82 &
  \textbf{0.77} &
  \multicolumn{1}{c}{0.060} \\ \hline
\end{tabular}%
}

\label{tab:CSQA_val}
\end{table}
\begin{table}[ht]
\scriptsize

\caption{SCORE comparison between results on \textbf{CommonsenseQA} \cite{talmor-etal-2019-commonsenseqa} with and without few-shot distribution modification.}

\begin{center}
\begin{tabular}{lcccccc}
\cline{2-7}
 &
  \multicolumn{3}{l}{\textbf{CommonsenseQA (prompt modified)}} &
  \multicolumn{3}{l}{\textbf{CommonsenseQA (prompt unmodified)}} \\ \hline
 
\multicolumn{1}{c}{\textbf{Model name}} &
  \multicolumn{1}{c}{\textbf{$SCORE_{S\&L}$}} &
  \multicolumn{1}{c}{\textbf{$SCORE_{M\&D}$}} &
  \textbf{$\Delta_{SCORE}$} &
  \multicolumn{1}{c}{\textbf{$SCORE_{S\&L}$}} &
  \multicolumn{1}{c}{\textbf{$SCORE_{M\&D}$}} &
  \textbf{$\Delta_{SCORE}$} \\ \hline
\multicolumn{1}{l}{Deepseek-r1-0528-qwen3} &
  0.458 &
  0.455 &
  0.002 &
  0.317 &
  0.471 &
  -0.153 \\ \cline{1-1}
\multicolumn{1}{l}{Deepseek-r1-distill-qwen} &
  0.511 &
  0.537 &
  -0.025 &
  0.489 &
  0.557 &
  -0.068 \\ \cline{1-1}
\multicolumn{1}{l}{Llama-3.1-8b-instruct} &
  0.783 &
  0.785 &
  -0.001 &
  0.781 &
  0.780 &
  0.002 \\ \cline{1-1}
\multicolumn{1}{l}{Ministral-3-8b-base} &
  0.612 &
  0.630 &
  -0.018 &
  0.586 &
  0.628 &
  -0.042 \\ \cline{1-1}
\multicolumn{1}{l}{Nvidia-nemotron-3-nano} &
  0.768 &
  0.754 &
  0.015 &
  0.757 &
  0.767 &
  -0.010 \\ \cline{1-1}
\multicolumn{1}{l}{Nemotron-cascade-8b} &
  0.905 &
  0.899 &
  0.005 &
  0.902 &
  0.897 &
  0.005 \\ \cline{1-1}
\multicolumn{1}{l}{Qwen3-14b} &
  0.906 &
  0.909 &
  -0.003 &
  0.887 &
  0.908 &
  -0.021 \\ \cline{1-1}
\multicolumn{1}{l}{Qwen3-30b-a3b-instruct} &
  0.930 &
  0.912 &
  0.018 &
  0.925 &
  0.917 &
  0.007 \\ \cline{1-1}
\multicolumn{1}{l}{Qwen3-8b} &
  0.882 &
  0.888 &
  -0.005 &
  0.873 &
  0.894 &
  -0.021 \\ \cline{1-1}
\multicolumn{1}{l}{Gemma-3-12b-it} &
  0.884 &
  0.881 &
  0.003 &
  0.876 &
  0.878 &
  -0.002 \\ \cline{1-1}
\multicolumn{1}{l}{Gemma-3-27b-it} &
  0.893 &
  0.895 &
  -0.002 &
  0.891 &
  0.888 &
  0.003 \\ \cline{1-1}
\multicolumn{1}{l}{Gpt-oss-20b} &
  0.779 &
  0.693 &
  0.086 &
  0.715 &
  0.690 &
  0.025 \\ \cline{1-1}
\multicolumn{1}{l}{Phi-4} &
  0.877 &
  0.819 &
  0.058 &
  0.876 &
  0.816 &
  0.060 \\ \hline
 
\multicolumn{1}{c}{\textbf{Mean}} &
  \textbf{0.784} &
  \textbf{0.774} &
  \textbf{0.010} &
  \textbf{0.760} &
  \textbf{0.776} &
  \textbf{-0.014} \\ \hline
\end{tabular}%
\end{center}

\label{tab:score_diff}
\end{table}

\section{Cross-Benchmark Agreement - Kendall Tau and Spearman differences}
\label{appendix:tau_and_spearman}
Let $B_i$ and $B_j$ denote two benchmarks, and let $\mathbf{x}_{B_i}^{S\&L}$ and $\mathbf{x}_{B_j}^{S\&L}$ be the vectors of mean model performance, averaged across the original permutation and all moving-answer attack permutations, under the standard Select-and-Letter evaluation protocol (S\&L) on benchmarks $B_i$ and $B_j$, respectively. We define
\begin{equation}
  \rho^{S\&L}(B_i,B_j)
\end{equation}
as the Spearman rank correlation computed between $\mathbf{x}_{B_i}^{S\&L}$ and $\mathbf{x}_{B_j}^{S\&L}$ across all evaluated models. For brevity, we omit the benchmark arguments and refer to this quantity as $\rho^{S\&L}$. 

Analogously, we define $\rho^{M\&D}$ or the proposed Matched-and-Dashed (M\&D) evaluation protocol, as well as their Kendall rank correlation counterparts, $\tau^{S\&L}$ and $\tau^{M\&D}$. 

To quantify changes in cross-benchmark agreement within the evaluation protocol, we consider the difference $\rho^{M\&D} - \rho^{S\&L}$. A positive value indicates stronger agreement between benchmarks under the M\&D protocol, while a negative value indicates higher agreement under S\&L. We define an analogous difference measure for Kendall’s tau, $\tau^{M\&D} - \tau^{S\&L}$.


\end{document}